\documentclass{article}
\usepackage{times}
\usepackage{mathtools}
\usepackage{graphicx}
\usepackage{sidecap}
\usepackage[small]{caption}
\usepackage{subcaption}
\usepackage{amsmath}
\allowdisplaybreaks
\usepackage{amsthm}

\usepackage{nicefrac}
\usepackage{clipboard}

\usepackage{contour}
\usepackage[normalem]{ulem}

\newcommand{\mydashuline}[1]{%
	\setlength{\ULdepth}{1.8pt}%
	\contourlength{0.8pt}%
	\dashuline{\phantom{#1}}%
	\llap{\contour{white}{#1}}%
}

\newcommand{\myuline}[1]{%
  \setlength{\ULdepth}{1.8pt}%
  \contourlength{0.8pt}%
  \uline{\phantom{#1}}%
  \llap{\contour{white}{#1}}%
}
\newcommand{\myuuline}[1]{%
  \setlength{\ULdepth}{3pt}%
  \contourlength{0.8pt}%
  \uuline{\phantom{#1}}%
  \llap{\contour{white}{#1}}%
}

\usepackage{multicol}

\usepackage{setspace}
\AtBeginDocument{%
  \addtolength\abovedisplayskip{-0.25\baselineskip}%
  \addtolength\belowdisplayskip{-0.25\baselineskip}%
  \addtolength\abovedisplayshortskip{-0.25\baselineskip}%
  \addtolength\belowdisplayshortskip{-0.25\baselineskip}%
}

\usepackage{pifont}
\usepackage{xspace}
\newcommand{\circone}{\ding{172}\xspace}
\newcommand{\circtwo}{\ding{173}\xspace}
\newcommand{\circthree}{\ding{174}\xspace}

\DeclareMathOperator{\argmax}{arg\,max}

\usepackage[round]{natbib}

\usepackage{appendix}

\newcommand{\rightcomment}[1]{\(\triangleright\) {\small \it #1}}
\newcommand{\eqcomment}[1]{\addtocounter{equation}{1}\tag*{\rightcomment{#1}\quad(\theequation)}}
\usepackage{suffix}
\WithSuffix\newcommand\eqcomment*[1]{\tag*{\rightcomment{#1}}}

\usepackage{algorithm}
\usepackage[hidelinks]{hyperref}

\usepackage[accepted]{icml2020}
\setcitestyle{yysep={,}}

\renewcommand{\paragraph}[1]{\textbf{#1}}

\usepackage[noend]{algpseudocode}

\renewcommand\algorithmicdo{:}
\renewcommand\algorithmicthen{:}
\algnewcommand{\IfThen}[2]{\State \algorithmicif\ #1\ \algorithmicthen\ #2}
\algnewcommand{\IfThenElse}[3]{\State \algorithmicif\ #1\ \algorithmicthen\ #2\ \algorithmicelse\ #3}
\algrenewcommand{\algorithmiccomment}[1]{\hfill \rightcomment{#1}}
\algnewcommand{\LineComment}[1]{\State \rightcomment{#1}}
\algnewcommand\algorithmicinput{{\bfseries Input:}}
\algnewcommand\INPUT{\item[\algorithmicinput]}
\algnewcommand\algorithmicoutput{{\bfseries Output:}}
\algnewcommand\OUTPUT{\item[\algorithmicoutput]}
\makeatletter

\newcommand{\algmargin}{\the\ALG@thistlm}
\makeatother
\algnewcommand{\Statepar}[1]{\State\parbox[t]{\dimexpr\linewidth-\algmargin}{\strut #1\strut}}
\newcommand{\pluseq}{\mathrel{+\!\!=}}

\usepackage{xpatch}
\xapptocmd\normalsize{%
 \abovedisplayskip=11pt plus 3pt minus 9pt
 \abovedisplayshortskip=0pt plus 3pt
 \belowdisplayskip=11pt plus 3pt minus 9pt
 \belowdisplayshortskip=6.5pt plus 3.5pt minus 3pt
}{}{}

\usepackage{amsmath}
\usepackage{amssymb}
\usepackage{mathrsfs}
\usepackage{mathtools, cuted}

\usepackage{tabularx, booktabs}
\newcolumntype{C}{>{\centering\arraybackslash}X}
\newcolumntype{R}{>{\raggedleft\arraybackslash}X}
\newcolumntype{S}{>{\raggedleft\arraybackslash\hsize=.5\hsize}X}
\usepackage{latexsym}
\usepackage{url}
\usepackage{xspace}
\usepackage{bm,array}
\usepackage{amsfonts}
\usepackage{enumitem}
\usepackage{mathtools}
\usepackage{bm}
\usepackage{esvect}

\usepackage{listings}
\usepackage{parcolumns}
\lstdefinestyle{datalogstyle}{
	basicstyle={\codefont\small},
	xleftmargin={14pt},
	numbers=left,
	frame=l,
	stepnumber=1,
	firstnumber=1,
	numberfirstline=true,
	tabsize=2,
	showtabs=false,
	showspaces=false,
	showstringspaces=false,
	extendedchars=true,
	breaklines=true,
	columns=fullflexible,
	keepspaces=true,
	escapeinside={@}{@},
	firstnumber=last,
	captionpos=b,
	commentstyle=\color{black!65},
	numberstyle=\tiny\color{black!65},
	stringstyle=\color{codepurple},
	breakatwhitespace=false, 
	keepspaces=true,                 
	numbersep=5pt,                  
	showspaces=false,                
	showstringspaces=false,
	showtabs=false,
	aboveskip={0.2\baselineskip},
	belowskip={-0.2\baselineskip},
}
\usepackage[noabbrev,capitalize]{cleveref}

\crefname{equation}{equation}{equations}
\crefname{section}{section}{sections}
\crefname{footnote}{footnote}{footnotes}   
\crefname{lstlsting}{listing}{listings}   
\crefname{lstlsting}{Listing}{Listings}   
\crefname{line}{rule}{rules}
\crefname{section}{\S}{\S\S}
\Crefname{section}{\S}{\S\S}
\crefformat{section}{#2\S#1#3}
\Crefformat{section}{#2\S#1#3}
\crefrangeformat{section}{\S\S#3#1#4--#5#2#6}
\Crefrangeformat{section}{\S\S#3#1#4--#5#2#6}
\crefmultiformat{section}{\S#2#1#3}{ and~\S#2#1#3}{, \S#2#1#3}{ and~\S#2#1#3}
\Crefmultiformat{section}{\S#2#1#3}{ and~\S#2#1#3}{, \S#2#1#3}{ and~\S#2#1#3}
\crefrangemultiformat{section}{\S\S#3#1#4--#5#2#6}{ and~\S\S#3#1#4--#5#2#6}{, \S\S#3#1#4--#5#2#6}{ and~\S\S#3#1#4--#5#2#6}
\Crefrangemultiformat{section}{\S\S#3#1#4--#5#2#6}{ and~\S\S#3#1#4--#5#2#6}{, \S\S#3#1#4--#5#2#6}{ and~\S\S#3#1#4--#5#2#6}

\usepackage{bbm}
\renewcommand{\vec}[1]{{\boldsymbol{\mathbf{#1}}}}

\newcommand{\defn}[1]{\textbf{#1}}
\newcommand{\defeq}{\mathrel{\stackrel{\textnormal{\tiny def}}{=}}}

\newcommand{\Real}{\mathbb{R}}

\newcommand{\Nat}{\mathbb{N}}

\renewcommand{\th}{\textsuperscript{th}\xspace}
\newcommand{\old}{_{\mathrm{old}}} 
\newcommand{\new}{_{\mathrm{new}}} 
\newcommand{\sign}{\mathrm{sign}}
\newcommand{\softplus}{\mathrm{softplus}}
\newcommand{\inv}[1]{#1^{\scriptscriptstyle-\!1}}

\newcommand{\dt}{\mathrm{d}t}

\newcommand{\inten}[2]{\lambda_{{#1}}(#2)}

\usepackage[usenames,dvipsnames,svgnames,table]{xcolor}
\usepackage[disable]{todonotes}

\newcommand{\Fixme}[2][]{\noindent}
\newcommand{\Notewho}[3][]{\noindent}
\newcommand{\Jason}[2][]{\noindent}
\newcommand{\Hongyuan}[2][]{\noindent}
\newcommand{\Guanghui}[2][]{\noindent}
\newcommand{\Minjie}[2][]{\noindent}

\makeatletter
\newcommand*\iftodonotes{\if@todonotes@disabled\expandafter\@secondoftwo\else\expandafter\@firstoftwo\fi}
\newlength{\extramargin}
\usepackage{calc}
\iftodonotes{
	\usepackage[paperwidth=\paperwidth+\extramargin*2,marginparwidth=\marginparwidth+\extramargin,width=\textwidth,height=\textheight]{geometry}
	\addtolength{\marginparwidth}{-7mm}
}{}
\makeatother
\newcommand{\cutforspace}[1]{}

\usepackage{tikz}
\usetikzlibrary{shapes.geometric}

\newcommand{\bluedot}{\begin{tikzpicture} \draw [fill=blue] (0,0) circle (0.10); \end{tikzpicture}\xspace}

\usetikzlibrary{arrows,decorations.markings}
\usetikzlibrary{arrows}

\newcommand{\greentriangle}{\begin{tikzpicture} \path node[regular polygon, regular polygon sides=3, fill=black!30!green, color=black!30!green, draw, thick, scale=0.4, rotate=180] (0,0.06) (triangle) {}; \end{tikzpicture}\xspace}

\newcommand{\aggrraw}[1]{\bigoplus^{#1}}
\newcommand{\aggr}[1]{\mathop{\mbox{$\aggrraw{#1}$}}}

\newcommand{\codefont}{\fontfamily{lmtt}\selectfont}
\newcommand{\data}[1]{\texttt{\codefont#1}}
\newcommand{\code}[1]{\data{#1}}
\newcommand{\dvar}[1]{\textcolor{black}{\data{#1}}}%
\newcommand{\mvar}[1]{\code{\textit{#1}\hspace{-1pt}}}
\newcommand{\set}[1]{{\mathcal{#1}}}
\newcommand{\key}[1]{\textbf{\data{#1}}}

\newcommand{\embed}{\textcolor{brown}{\key{embed}}}
\newcommand{\isevent}{\textcolor{brown}{\key{event}}}
\newcommand{\iseventgroup}{\textcolor{brown}{\key{eventgroup}}}
\newcommand{\boo}[1]{\data{#1}}
\newcommand{\blk}[1]{\textcolor{blue}{\data{#1}}}
\newcommand{\ise}[1]{\textcolor{orange}{\data{#1}}}
\newcommand{\exo}[1]{\mydashuline{\ise{#1}}}
\newcommand{\exoo}[1]{\mydashuline{\boo{#1}}}
\newcommand{\evt}[1]{\myuline{\ise{#1}}}
\newcommand{\evto}[1]{\myuline{\boo{#1}}}
\newcommand{\grp}[1]{\myuuline{\ise{#1}}}

\newcommand{\prm}[1]{\textcolor{purple}{\data{#1}}}

\newcommand{\usec}{\textcolor{purple}{\frenchspacing\data{:}}\xspace}
\newcommand{\useb}{\textcolor{purple}{\frenchspacing\data{:\!:}}\xspace}
\newcommand{\use}{\useb}
\newcommand{\dep}{\textcolor{brown}{\data{:\!-}}\xspace}
\newcommand{\dephighway}{\textcolor{brown}{\data{:\!=}}\xspace} 
\newcommand{\see}{\textcolor{LimeGreen}{\data{<\!-}}\xspace}
\newcommand{\grpdep}{\textcolor{LimeGreen}{\data{<\!<\!-}}\xspace}
\newcommand{\depgrp}{\textcolor{LimeGreen}{\data{<\!-{}-}}\xspace}

\newcommand{\trueval}{\textsf{true}}

\newcommand{\nullval}{\textsf{null}}

\usepackage{stmaryrd}
\newcommand{\sem}[1]{\llbracket #1\rrbracket}
\newcommand{\sema}[2]{\sem{\code{\blk{#1}(#2)}}}
\newcommand{\seme}[2]{\sem{\code{\evt{#1}(#2)}}}
\newcommand{\depval}[2]{\boldsymbol{[}#1\boldsymbol{]}^{\dep}_{#2}}
\newcommand{\seeval}[2]{\boldsymbol{[}#1\boldsymbol{]}^{\see}_{#2}}
\newcommand{\seeupd}[3][]{\boldsymbol{[}#2\boldsymbol{]}^{\Delta#1}_{#3}} 
\newcommand{\seedelta}[2]{\boldsymbol{[}#1\boldsymbol{]}^{\vec{\delta}}_{#2}}
\newcommand{\seecell}[1]{{\setlength{\fboxsep}{0pt}\framebox{$\phantom{[}#1\phantom{]}$}}}

\newcommand{\timescale}{\tau}
\newcommand{\warpfunc}{v}

\newcommand{\dur}{{T}}

\usepackage{verbatim}

\newcommand{\titlenameshort}{Neural Datalog Through Time}
\newcommand{\titlename}{\titlenameshort: \\ Informed Temporal Modeling via Logical Specification}
\icmltitlerunning{\titlenameshort}

\begin{document}

\twocolumn[
\icmltitle{\titlename}
\begin{icmlauthorlist}
	\icmlauthor{Hongyuan Mei}{jhu}
	\icmlauthor{Guanghui Qin}{jhu}
	\icmlauthor{Minjie Xu}{bbg}
	\icmlauthor{Jason Eisner}{jhu}
\end{icmlauthorlist}

\icmlaffiliation{jhu}{Computer Science Dept.\@, Johns Hopkins Univ.}
\icmlaffiliation{bbg}{Bloomberg LP}

\icmlcorrespondingauthor{Hongyuan Mei}{hmei@cs.jhu.edu}

\vskip 0.3in
]

\printAffiliationsAndNotice{}

\begin{abstract}\label{sec:abstract}
Learning how to predict future events from patterns of past events is difficult when the set of possible event types is large.  Training an unrestricted neural model might overfit to spurious patterns.
To exploit domain-specific knowledge of how past events might affect an event's present probability, we propose using a \emph{temporal deductive database} to track structured facts over time.  Rules serve to prove facts from other facts
and from past events.  Each fact has a time-varying state---a vector computed by a neural net whose topology is determined by the fact's \emph{provenance}, including its experience of past events.  The possible event types at any time are given by special facts, whose \emph{probabilities} are neurally modeled alongside their states.
In both synthetic and real-world domains, we show that neural probabilistic models derived from concise Datalog programs improve prediction by encoding
appropriate domain knowledge in their architecture.\looseness=-1

\end{abstract}

\vspace{-1.5\baselineskip}
\section{Introduction}\label{sec:intro}

Temporal sequences are abundant in applied machine learning.  A common task is to predict the future from the past or to impute other missing events. Often this is done by fitting a generative probability model.  For evenly spaced sequences, historically popular generative models have included hidden Markov models and discrete-time linear dynamical systems, with more recent interest in recurrent neural network models such as LSTMs.  For irregularly spaced sequences, a good starting point is the Hawkes process (a self-exciting temporal point process) and its many variants, including neuralized versions based on LSTMs.

Under any of these models, each event $e_i$ updates the state of the system from $\vec{s}_i$ to $\vec{s}_{i+1}$, which then determines the distribution from which the next event $e_{i+1}$ is drawn.%
Alas, when the relationship between events and the system
state is unrestricted---when anything can potentially affect anything---fitting an accurate model is very difficult, particularly in a real-world domain that allows millions of event types including many rare types. 
Thus, one would like to introduce domain-specific structure into the model.

For example, 
one might declare that the probability that Alice travels to Chicago is determined entirely by Alice's state, the states of Alice's coworkers such as Bob, and the state of affairs in Chicago. Given that modeling assumption, parameter estimation can no longer incorrectly overfit this probability using spurious features based on unrelated temporal patterns of (say) wheat sales and soccer goals.

To improve extrapolation, one can reuse this ``Alice travels to Chicago'' model for any person \dvar{A} traveling to any place \dvar{C}.  
Our main contribution is a modeling language 
that can concisely model all these \code{\evto{travel}(\dvar{A},\dvar{C})} probabilities using a few rules over variables \dvar{A}, \dvar{B}, \dvar{C}.  Here \dvar{B} ranges over \dvar{A}'s coworkers, where the  \boo{coworker} relation is also governed by rules and can itself be affected by stochastic events.\looseness=-1

In our paradigm, a domain expert simply writes down the rules of a \defn{temporal deductive database}, which tracks the possible event types and other \emph{boolean} facts over time.  This logic program is then used to automatically construct a deep recurrent neural architecture, whose distributed state consists of vector-space \emph{embeddings} of all present facts.  Its output specifies the distribution of the next event.

What sort of rules?  An \defn{event} has a \emph{structured} description with zero or more participating \defn{entities}.  
When an event happens, pattern-matching against its description triggers \defn{update rules}, which modify the database facts to reflect the new properties and relationships of these entities.  Updates may have a cascading effect if the database contains \defn{deductive rules} that derive further facts from existing ones at any time.  (For example,
\code{\boo{coworker}(\dvar{A},\dvar{B})} is jointly implied by \code{\boo{boss}(\dvar{U},\dvar{A})} and \code{\boo{boss}(\dvar{U},\dvar{B})}).  
In particular, deductive rules can 
state that entities combine into a possible event type whenever they
have the appropriate properties and relationships.  (For example, \code{\evto{travel}(\dvar{A},\dvar{C})} is possible if \dvar{C} is a place and \dvar{A} is a person who is not already at \dvar{C}.)\looseness=-1

Since the database defines possible events and is updated by the event that happens, it already resembles the system state $\vec{s}_i$ of a temporal model.  We enrich this logical state by associating an embedding with each fact currently in the database.  This time-varying vector represents the state of \emph{that fact}; recall that the set of facts may also change over time.  When a fact is added by events or derived from other facts, its embedding is derived from their embeddings in a standard way, using parameters associated with the rules that established the fact.  In this way, the model's rules together with the past events and the initial facts define the topology of a deep recurrent neural architecture, which can be trained via back-propagation through time \cite{williams-zipser-1989}.  For the facts that state that specific event types are possible, the architecture computes not only embeddings but also the probabilities of these event types.

The number of parameters of such a model grows only with the number of rules, not with the much larger number of event types or other facts.  This is analogous to how a probabilistic relational model \citep{getoor-07-srl,richardson-06-markov}
derives a graphical model structure from a database, building random variables from database entities and repeating subgraphs with shared parameters.

Unlike graphical models, ours is a neural-symbolic hybrid.
The system state $\vec{s}_i$ includes both rule-governed discrete elements (the set of facts) and learned continuous elements (the embeddings of those facts).  It can learn a neural probabilistic model of people's movements while relying on a discrete symbolic deductive database to cheaply and accurately record who is where.  A purely neural model such as our neural Hawkes process \cite{mei-17-neuralhawkes} would have to \emph{learn} how to encode every location fact in some very high-dimensional state vector, and retain and update it, with no generalization across people and places.\looseness=-1

In our experiments, we show how to write down some domain-specific
models for irregularly spaced event sequences in continuous time, and
demonstrate that their structure improves their ability to predict held-out data.

\vspace{-1pt}
\section{Our Modeling Language}\label{sec:lang}

We gradually introduce our specification language by developing a fragment of a human activity model.
Similar examples could be developed in many other domains---epidemiology, medicine, education, organizational behavior, consumer behavior, economic supply chains, etc.
Such specifications can be trained and evaluated using our implementation,
which can be found at {\small \url{https://github.com/HMEIatJHU/neural-datalog-through-time}}.

For pedagogical reasons, \cref{sec:lang} will focus on our high-level scheme (see also the animated drawings in our ICML 2020 talk video).
We defer the actual neural formulas until \cref{sec:formula}.  

\vspace{-1pt}
\subsection{Datalog}\label{sec:datalog}

We adapt our notation from Datalog \citep{ceri-1989-datalog}, where one can write \defn{deductive rules} of the form
\begin{align}
\code{\mvar{head} \dep} &\code{ \mvar{condit}$_1$, $\ldots$, \mvar{condit}$_N$.} \label{eqn:dep_rule}
\end{align}
Such a rule states that the head is true provided that the conditions are all true.\footnote{\Cref{app:negbody} discusses an extension to negated conditions.} In a simple case, the head and conditions are \defn{atoms}, i.e., structured terms that represent boolean propositions.  For example,
\begin{lstlisting}[name=human,firstnumber=auto]
@\boo{compatible}(eve,adam) \dep \\ \boo{likes}(eve,apples), \boo{likes}(adam,apples).@
\end{lstlisting}

If $N=0$, the rule simply states that the head is true.  This case is useful to assert basic facts:
\begin{lstlisting}[name=human,firstnumber=auto]
@\boo{likes}(eve,apples).@ @\label{line:eveapples}@
\end{lstlisting}
Notice that in this case, the {\dep} symbol is 
omitted.

A rule that contains \defn{variables} (capitalized identifiers) represents the infinite collection of \defn{ground} rules obtained by
instantiating (grounding) those variables.  For example,
\begin{lstlisting}[name=human,firstnumber=auto]
@\boo{compatible}(\dvar{X},\dvar{Y}) \dep \boo{likes}(\dvar{X},\dvar{U}), \boo{likes}(\dvar{Y},\dvar{U}). \label{line:compatible}@
\end{lstlisting}
says that \emph{any} two entities \dvar{X} and \dvar{Y} are compatible provided that there exists \emph{any} \dvar{U} that they both like.

A Datalog \defn{program} is an unordered set of rules.
The atoms that can be proved from these rules are called \defn{facts}.  Given a program, one would use $\sem{\mvar{h}} \in \{ \trueval, \nullval \}$ to denote the semantic value of atom \mvar{h}, where $\sem{\mvar{h}} = \trueval$ iff \mvar{h} is a fact.

\vspace{-3pt}
\subsection{Neural Datalog}\label{sec:ndl}

In our formalism, a fact has an \defn{embedding} in a vector space, so the semantic value of atom \code{\boo{likes}(eve,apples)} describes more than just \emph{whether} \code{eve} likes \code{apples}.  To indicate this, let us rename and colorize the functors in \cref{line:compatible}:
\begin{lstlisting}[name=human,firstnumber=auto]
@\blk{rel}(\dvar{X},\dvar{Y}) \dep \blk{opinion}(\dvar{X},\dvar{U}), \blk{opinion}(\dvar{Y},\dvar{U}). \label{line:rel}@
\end{lstlisting}
Now $\sema{opinion}{eve,apples}$ is a vector describing \code{eve}'s complex opinion about apples (or $\nullval$ if she has no opinion).
$\sema{rel}{eve,adam}$ is a vector describing \code{eve} and \code{adam}'s relationship (or $\nullval$ if they have none).  

With this extension, $\sem{\mvar{h}} \in \Real^{D_{\mvar{h}}} \cup \{ \nullval \}$, where the embedding dimension $D_{\mvar{h}}$ depends on the atom $\mvar{h}$.
The declaration
\begin{lstlisting}[name=human,firstnumber=auto]
@\code{\dep \embed(\blk{opinion},8).}@
\end{lstlisting}
says that if \mvar{h} has the form \code{\blk{opinion}(...)} then $D_{\mvar{h}} = 8$.\footnote{\label{fn:nocolor}In the absence of such a declaration, $D_{\mvar{h}} = 0$. Then $\sem{\mvar{h}}$ has only two possible values, just as in Datalog; we do not color $\mvar{h}$.}

When an atom is proved via a rule, its embedding is affected by the conditions of that rule, in a way that depends on trainable parameters associated with that rule.  For example, according to \cref{line:rel}, $\sema{rel}{eve,adam}$ is a parametric function of the opinion vectors that \code{eve} and \code{adam} have about various topics \code{\dvar{U}}.  The influences from all their shared topics are pooled together as detailed in \cref{sec:ndl_math} below.

A model might say that {\em each} person has an opinion about {\em each} food, which is a function of the embeddings of the person and the food, using parameters associated with \cref{line:personfood}:  
\begin{lstlisting}[name=human,firstnumber=auto]
@\blk{opinion}(\dvar{X},\dvar{U}) \dep \blk{person}(\dvar{X}), \blk{food}(\dvar{U}). \label{line:personfood}@
\end{lstlisting}
If the foods are simply declared as basic facts, as follows, then each food's embedding is independently specified by the parameters associated with the rule that declares it:
\begin{lstlisting}[name=human,firstnumber=auto]
@\blk{food}(apples).@
@\blk{food}(manna). \progvdots@
\end{lstlisting}

Given all the rules above, whenever \code{\blk{person}(\dvar{X})} and \code{\blk{person}(\dvar{Y})} are facts, it follows that \code{\blk{rel}(\dvar{X},\dvar{Y})} is a fact, and $\sema{rel}{\dvar{X},\dvar{Y}}$ is defined by a multi-layer feed-forward neural network whose topology is given by the proof DAG for \code{\blk{rel}(\dvar{X},\dvar{Y})}.  The network details will be given in \cref{sec:ndl_math}.

Recursive Datalog rules can lead to arbitrarily deep networks that recursively build up a compositional embedding, just as in sequence encoders \citep{elman-90-finding}, tree encoders \citep{socher-etal-2012-semantic,tai-15-improved}, and DAG encoders \cite{goller-kuchler-1996,le-15-forest}---all of which could be implemented in our formalism.  E.g.:
\begin{lstlisting}[name=human,firstnumber=auto]
@\blk{cursed}(cain).@ @\label{line:curse_base}@
@\blk{cursed}(\dvar{Y}) \dep \blk{cursed}(\dvar{X}),  \blk{parent}(\dvar{X},\dvar{Y}).@ @\label{line:curse_rec}@
\end{lstlisting}
In Datalog, this system simply states that all descendants of \code{cain} are cursed.
In neural Datalog, however, a child has a \emph{specific} curse: a vector $\sema{cursed}{\dvar{Y}}$
that is computed from the parent's curse $\sema{cursed}{\dvar{X}}$
in a way that also depends on their relationship, as encoded by the vector $\sema{parent}{\dvar{X},\dvar{Y}}$.  
\Cref{line:curse_rec}'s parameters model how the curse evolves (and hopefully attenuates) as each generation is re-cursed.  Notice that $\sema{cursed}{\dvar{Y}}$ is essentially computed by a recurrent neural network that encodes the sequence of \code{\blk{parent}} edges that connect \code{cain} to \dvar{Y}.\footnote{Assuming that this path is unique.  More generally, \dvar{Y} might descend from \code{cain} by multiple paths.  The computation actually encodes the DAG of \emph{all} paths, by pooling over all of \dvar{Y}'s cursed parents at each step, just as \cref{line:rel} pooled over multiple topics.}

We currently consider it to be a model specification error if any atom $\mvar{h}$ participates in its own proof, leading to a circular definition of $\sem{\mvar{h}}$.  This would happen in \crefrange{line:curse_base}{line:curse_rec} only if \code{\blk{parent}} were bizarrely defined to make some cursed person their own ancestor.  \Cref{app:cyclic} discusses extensions that would define $\sem{\mvar{h}}$ even in these cyclic cases. 

\subsection{Datalog Through Time}\label{sec:dltt}

For temporal modeling, we use atoms such as \code{\evt{help}(\dvar{X},\dvar{Y})} as the structured names for events.  We underline their functors.  As usual, we colorize them if they have vector-space embeddings (see \cref{fn:nocolor}), but as orange rather than blue.

We extend Datalog with \defn{update rules} so that whenever a \code{\evt{help}(\dvar{X},\dvar{Y})} event occurs 
under appropriate conditions, it can add to the database by proving new atoms:
\begin{lstlisting}[name=human,firstnumber=auto]
@\blk{grateful}(\dvar{Y},\dvar{X}) \see \evt{help}(\dvar{X},\dvar{Y}), \blk{person}(\dvar{Y}). \label{line:grateful}@
\end{lstlisting}
An event can also cancel out such additions, which may make atoms false again.\footnote{The atom will remain true if it remains provable by a {\dep} rule,
 or is proved by another {\see} rule at the same time.
}
The \code{!} symbol means ``not'':
\begin{lstlisting}[name=human,firstnumber=auto]
@!\blk{grateful}(\dvar{Y},\dvar{X}) \see \evt{harm}(\dvar{X},\dvar{Y}). \label{line:disgrate}@
\end{lstlisting}

The general form of these \defn{update rules} is
\begin{subequations}\label{eqn:see_rules}
\begin{align}
\hspace{-2mm}\code{\mvar{head} \see} &\code{ \mvar{event}, \mvar{condit}$_1$, $\ldots$, \mvar{condit}$_N$.} \label{eqn:see_launch}\\
\hspace{-2mm}\code{!\mvar{head} \see} &\code{ \mvar{event}, \mvar{condit}$_1$, $\ldots$, \mvar{condit}$_N$.} \label{eqn:see_dock}
\end{align}
\end{subequations}
which state that \mvar{event} makes \mvar{head} true or false, respectively, provided that the conditions are all true. 
An event occurring at time $s$ affects the set of facts at times $t > s$, both directly through $\see$ rules, and also indirectly, since the facts added or removed by $\see$ rules may affect the set of additional facts that can be derived by $\dep$ rules at time $t$.  Our approach can be used for either discrete time ($s,t \in \Nat$) or continuous time ($s,t \in \Real_{\geq 0}$), where the latter supports irregularly spaced events \citep[e.g.,][]{mei-17-neuralhawkes}.\looseness=-1

\subsection{Neural Datalog Through Time}\label{sec:ndtt}\label{sec:exo}

In \cref{sec:ndl}, we derived each fact's embedding from its proof DAG, representing its set of Datalog proofs.  For Datalog through time, we must also consider how to embed facts that were proved by an earlier update.  Furthermore, once an atom is proved, an update rule can prove it again.  This will update its embedding, in keeping with our principle that a fact's embedding is influenced by \emph{all} of its proofs.

As an example, when \dvar{X} helps \dvar{Y} and \code{\blk{grateful}(\dvar{Y},\dvar{X})} first becomes true via \cref{line:grateful}, the new embedding $\sema{grateful}{\dvar{Y},\dvar{X}}$ is computed---using parameters associated with \cref{line:grateful}---from the embeddings of \code{\evt{help}(\dvar{X},\dvar{Y})} and \code{\blk{person}(\dvar{Y})}.  Those embeddings model the nature of the help and the state of person \dvar{Y}.  (This was the main reason for \cref{line:grateful} to include \code{\blk{person}(\dvar{Y})} as a condition.)
Each time \dvar{X} helps \dvar{Y} again, $\sema{grateful}{\dvar{Y},\dvar{X}}$ is further updated by \cref{line:grateful}, so this gratitude vector records the \emph{history} of help.  The updates are LSTM-like (see \cref{sec:ndtt_math} for details).\looseness=-1

In general, an atom's semantics can now vary over time and so should be denoted as $\sem{\mvar{h}}(t)$: the \defn{state} of atom $\mvar{h}$ at time $t$, which is part of the overall database state.   A \dep rule as in \cref{eqn:dep_rule} says that $\sem{\mvar{head}}(t)$ depends parametrically on $\{\sem{\mvar{condit}_i}(t): 1 \leq i \leq N\}$.  A {\see} rule as in \cref{eqn:see_launch} says that if \mvar{event} occurred at time $s < t$ and no events updating \mvar{head} occurred on the time interval $(s,t)$, then $\sem{\mvar{head}}(t)$ depends parametrically on its previous value%
\footnote{More precisely, it depends on the LSTM cells that contributed to that previous value, as we will see in \cref{sec:ndtt_math}.} 
$\sem{\mvar{head}}(s)$ along with $\sem{\mvar{event}}(s)$, $\{\sem{\mvar{condit}_i}(s): 1 \leq i \leq N\}$, and the elapsed time $t-s$.  We will detail the parametric formulas in \cref{sec:ndtt_math}.

Thus, $\sem{\mvar{head}}(t)$ depends via {\dep} rules on \mvar{head}'s \emph{provenance} in the database at time $t$, and depends via {\see} rules on its \emph{experience} of events at strictly earlier times.\footnote{See \cref{sec:ndtt_math} for the precise interaction of {\dep} and {\see} rules.}  This yields a neural architecture similar to a stacked LSTM: the {\dep} rules make the neural network deep at a single time step, while the {\see} rules make it temporally recurrent across time steps.  The network's irregular topology is defined by the $\dep$ and $\see$ rules plus the events that have occurred.

\subsection{Probabilistic Modeling of Event Sequences}\label{sec:eventfacts}

Because events can \defn{occur}, atoms that represent event types are special.  They can be declared as follows:

\begin{lstlisting}[name=human,firstnumber=auto]
@\dep \isevent(\evt{help}, 8). @ @\label{line:help_event}@
\end{lstlisting}

Because the declaration is \code{\isevent} rather than \code{\embed}, 
at times when \code{\evt{help}(\dvar{X},\dvar{Y})} is a fact, it will have a positive probability along with its embedding $\seme{help}{\dvar{X},\dvar{Y}} \in \Real^8$.  This is what the underlined functor really indicates.

At times $s$ when \code{\evt{help}(\dvar{X},\dvar{Y})} is not a fact, the semantic value $\seme{help}{\dvar{X},\dvar{Y}}(s)$ will be \nullval, and it will have neither an embedding nor a probability.  At these times, it is simply not a possible event; its probability is effectively 0.

Thus, the model must include rules that establish the set of possible events as facts.  For example, the rule
\begin{lstlisting}[name=human,firstnumber=auto]
@\code{\evt{help}(\dvar{X},\dvar{Y}) \dep \blk{rel}(\dvar{X},\dvar{Y}).} \label{line:help}@
\end{lstlisting}
says if \dvar{X} and \dvar{Y} have a relationship, then \code{\evt{help}(\dvar{X},\dvar{Y})} is true, meaning that events of the type \code{\evt{help}(\dvar{X},\dvar{Y})} have positive probability (i.e., \dvar{X} can help \dvar{Y}).
The embedding and probability are computed deterministically from $\sema{rel}{\dvar{X},\dvar{Y}}$ using parameters associated with \cref{line:help}, as detailed in \cref{sec:prob}.

Now a neural-Datalog-through-time program specifies a probabilistic model over event sequences.  Each stochastic event can update some database facts or their embeddings, as well as the probability distribution over possible next events.  As \cref{sec:intro} outlined, each \emph{stochastic draw} from the next-event distribution results in a \emph{deterministic update} to that distribution---just as in a recurrent neural network language model \cite{mikolov-10-rnnlm,sundermeyer-12-lstm}.\looseness=-1

Our approach also allows the possibility of \defn{exogenous} events that are not generated by the model, but are given externally.  Our probabilistic model is then \emph{conditioned} on these exogenous events. The model itself might have probability 0 of generating these event types at those times.  Indeed, if an event type is to occur \emph{only} exogenously, then the model should not predict any probability for it, so it should not be declared using \code{\isevent}.  We use a dashed underline for undeclared events since they have no probability.  

For example, we might wish to use rules of the form \code{\mvar{head} \see \exo{earthquake}(\dvar{C}),$\ldots$} to model how an earthquake in city \dvar{C} tends to affect subsequent events, even if we do not care to model the \emph{probabilities} of earthquakes.  The \emph{embeddings} of possible earthquake events can still be determined by parametric rules, e.g., \code{\exo{earthquake}(\dvar{C}) \dep \blk{city}(\dvar{C})}, if we request them by declaring \code{\embed(\exo{earthquake},5)}.

\vspace{-4pt}
\subsection{Continuing the Example}\label{sec:continue}

In our example, the following rules are also plausible.  They say that when \dvar{X} helps \dvar{Y}, this event updates the states of the helper \dvar{X} and the helpee \dvar{Y} and also the state of their relationship:
\begin{lstlisting}[name=human,firstnumber=auto]
@\blk{person}(\dvar{X}) \see \evt{help}(\dvar{X},\dvar{Y}).@ @\label{line:helper}@
@\blk{person}(\dvar{Y}) \see \evt{help}(\dvar{X},\dvar{Y})@ @\label{line:helpee}@
@\blk{rel}(\dvar{X},\dvar{Y}) \ \see \evt{help}(\dvar{X},\dvar{Y}).@ @\label{line:relhelp}@ 
\end{lstlisting}
To enrich the model further, we could add (e.g.) \code{\blk{rel}(\dvar{X},\dvar{Y})} as a condition to these rules.  Then the update when \dvar{X} helps \dvar{Y} depends quantitatively on the state of their relationship.

There may be many other kinds of events observed in a human activity dataset, such as \code{\evt{sleep}(\dvar{X})}, \code{\evt{eat}(\dvar{X})}, \code{\evt{email}(\dvar{X},\dvar{Y})}, \code{\evt{invite}(\dvar{X},\dvar{Y})}, \code{\evt{hire}(\dvar{X},\dvar{Y})}, etc.  These can be treated similarly to \code{\evt{help}(\dvar{X},\dvar{Y})}.

Our modeling architecture is intended to limit dependencies to those that are explicitly specified, just as in graphical models.  However, the resulting independence assumptions may be too strong.  To allow unanticipated influences back into the model, it can be useful to include a low-dimensional global state, which is updated by all events:
\begin{lstlisting}[name=human,firstnumber=auto]
@\blk{world} \see \evt{help}(\dvar{X},\dvar{Y}). \progvdots@
\end{lstlisting}
\blk{world} records a ``public history'' in its state, and it can be a condition for any rule.  E.g., we can replace \cref{line:help} with
\begin{lstlisting}[name=human,firstnumber=auto]
@\evt{help}(\dvar{X},\dvar{Y}) \dep \blk{rel}(\dvar{X},\dvar{Y}), \blk{world}.@
\end{lstlisting}
so that \code{eve}'s probability of helping \code{adam} might be affected by the history of other individuals' interactions.

Eventually \code{eve} and \code{adam} may die, which means that they are no longer available to help or be helped:
\begin{lstlisting}[name=human,firstnumber=auto]
@\evt{die}(X) \dep \blk{person}(\dvar{X}).@
\end{lstlisting}

If we want \code{\blk{person}(eve)} to then become false, the model cannot place that atom in the database with a {\dep} rule like 
\begin{lstlisting}[name=human,firstnumber=auto]
@\blk{person}(eve).@
\end{lstlisting}
which would ensure that \code{\blk{person}(eve)} can \emph{always} be proved.  
Instead, we use a {\see} rule that initially adds \code{\blk{person}(eve)} to the database via a special event, \code{\exoo{init}}, that always occurs exogenously at time $t=0$:
\begin{lstlisting}[name=human,firstnumber=auto]
@\blk{person}(eve) \see \exoo{init}.@
\end{lstlisting}
With this treatment, the following rule can remove \code{\blk{person}(eve)} again when she dies:
\begin{lstlisting}[name=human,firstnumber=auto]
@!\blk{person}(\dvar{X}) \see \evt{die}(\dvar{X}).@ @\label{line:die}
\end{lstlisting}
The reader may enjoy extending this model to handle possessions, movement, tribal membership/organization, etc.

\subsection{Finiteness}\label{sec:finite}

Under our formalism, any given model allows only a finite set of possible events.  This is because a Datalog program's facts are constructed by using functors mentioned in the program, with arguments mentioned in the program,\footnote{A rule such as \code{\blk{likes}(adam,\dvar{Y}) \dep \blk{likes}(adam,eve)} might be able to prove that \code{adam} likes everyone, including infinitely many unmentioned entities.  To preserve finiteness, such rules are illegal in Datalog.  A Datalog rule must be \defn{range-restricted}: any variable in the head must also appear in the body.} and nesting is disallowed.  Thus, the set of facts is finite (though perhaps much larger than the length of the program).\looseness=-1  

It is this property that will ensure in \cref{sec:prob} that our probability model---which sums over all possible events---is well-defined.  Yet this is also a limitation.  In some domains, a model should not really place any {\em a priori} bound on the number of event types, since an infinite sequence may contain infinitely many distinct types---the number of types represented in the length-$n$ prefix grows unboundedly with $n$.  Even our running example should really support the addition of new entities: the event \code{\evt{procreate}(eve,adam)} should result in a fact such as \code{\evt{person}(cain)}, where \code{cain} is a newly allocated entity. Similarly, new species are allocated in the course of drawing a sequence from Fisher's (\citeyear{fisher-corbet-williams-1943}) species-sampling model or from a Chinese restaurant process; new words are allocated as a document is drawn from an infinite-vocabulary language model; and new real numbers are constantly encountered in a sequence of sensor readings.  In these domains, no model can \emph{prespecify} all the entities that can appear in a dataset.  \Cref{app:infinite} discusses potential extensions to handle these cases.\looseness=-1

\section{Formulas Associated With Rules}\label{sec:formula}

\subsection{Neural Datalog}\label{sec:ndl_math}

\newcommand{\rth}{$r$\th}

Recall from \cref{sec:datalog} that if $\mvar{h}$ is a fact,
it is provable by at least one $\dep$ rule in at least one way.  
For neural Datalog (\cref{sec:ndl}), we then choose to define the embedding $\sem{\mvar{h}}\neq\nullval$ as
\begin{align}\label{eqn:term_emb}%
\sem{\mvar{h}} \;&\defeq\; \tanh \big( \sum_r \depval{\mvar{h}}{r} \big) \; \in (-1,1)^{D_{\mvar{h}}}
\end{align}
where $\depval{\mvar{h}}{r}$ represents the \defn{contribution}
of the \rth rule of the Datalog program.  For example, 
$\sema{opinion}{eve,apples}$ receives non-zero contributions from \emph{both} \cref{line:eveapples} and \cref{line:personfood}.\footnote{Recall that we renamed \code{likes} in \cref{line:eveapples} to \code{opinion}.}  For a given $\dvar{Y}$, $\sema{cursed}{\dvar{Y}}$ may receive a non-zero contribution from \cref{line:curse_base}, \cref{line:curse_rec}, or neither, according to whether \dvar{Y} is \code{cain} himself, a descendant of \code{cain}, or neither.

The contribution $\depval{\mvar{h}}{r}$ has been pooled over all the ways (if any) that the \rth rule proves $\mvar{h}$.  For example, for any entity \dvar{Y},  $\depval{\code{\blk{cursed}(\dvar{Y})}}{\ref{line:curse_rec}}$ needs to compute the \emph{aggregate} effect of the curses that \dvar{Y} inherits through \emph{all} of \dvar{Y}'s cursed parents \dvar{X} in \cref{line:curse_rec}. Similarly, $\depval{\code{\blk{rel}(\dvar{X},\dvar{Y})}}{\ref{line:rel}}$ computes the aggregate effect on the relationship from \emph{all} of \dvar{X} and \dvar{Y}'s shared interests \dvar{U}  in \cref{line:rel}.
Recall from \cref{sec:datalog} that a rule with variables represents a collection of ground rules obtained by instantiating those variables.  We define its contribution by\looseness=-1
\begin{align}\label{eqn:rule_emb}
\depval{\mvar{h}}{r} &\defeq \aggr{\beta_r}_{\mvar{g}_1, \ldots, \mvar{g}_N} 
\vec{W}_r \underbrace{[1; \sem{\mvar{g}_1}; \ldots; \sem{\mvar{g}_N}]}_{\clap{\text{\scriptsize concatenation of column vectors}}} \;\in \Real^{D_{\mvar{h}}}
\end{align}
where for the summation, we allow \code{\mvar{h} \dep \mvar{g}$_1$, $\ldots$, \mvar{g}$_N$} to range over all instantiations of the \rth rule such that the head equals $\mvar{h}$ and $\mvar{g}_1, \ldots, \mvar{g}_N$ are all facts.  There are only finitely many such instantiations (see \cref{sec:finite}).  $\vec{W}_r$ is a conformable parameter matrix associated with the \rth rule.  
(\Cref{app:param_share} offers extensions that allow more control over how parameters are shared among and within rules.)

The pooling operator $\aggr{\beta}$ that we used above is defined to aggregate a set of vectors $\{ \vec{x}_1, \ldots, \vec{x}_M \}$:
\begin{align}\label{eqn:aggr}
	\aggr{\beta}_{m} \vec{x}_m 
	\;&\defeq\; \inv{\warpfunc}(\sum_m \warpfunc(\vec{x}_m) )  
\end{align}%
Remarks: For any definition of function $\warpfunc$ with inverse $\inv{\warpfunc}$, $\aggr{\beta}$ has a unique identity element, $\inv{\warpfunc}(\vec{0})$, which is also the result of pooling no vectors ($M\!=\!0$).  
Pooling a single vector ($M\!=\!1$) returns that vector---so when rule $r$ proves \mvar{h} in only one way, the contribution of the $\sem{\mvar{g}_i}$ to $\sem{\mvar{h}}$ does not have to involve an ``extra'' nonlinear pooling step in \cref{eqn:rule_emb}, but only the nonlinear $\tanh$ in \cref{eqn:term_emb}.

Given $\beta \neq 0$, we take $\warpfunc$ to be the differentiable function
\begin{subequations}\label[equation]{eqn:squashes}
	\begin{align}
		\warpfunc(\vec{x}) 
		   \;&\defeq\; \sign(\vec{x})\,|\vec{x}|^{\beta} \label{eqn:squash} \\
		\inv{\warpfunc}(\vec{y}) 
		   &=\; \sign(\vec{y})\,|\vec{y}|^{1/\beta} \label{eqn:unsquash}
	\end{align}
\end{subequations}
where all operations are applied elementwise.  Now the result of aggregating no vectors is $\vec{0}$, so rules that achieve no proofs of $\mvar{h}$ contribute nothing to \cref{eqn:term_emb}.
If $\beta = 1$, then $\warpfunc = \mathrm{identity}$ and $\aggr{\beta}$ is just summation. 
As $\beta \rightarrow \infty$, $\aggr{\beta}$ emphasizes more extreme values, approaching a signed variant of max-pooling that chooses (elementwise) the argument with the largest absolute value. 
As a generalization, one could replace the scalar $\beta$ with a vector $\vec{\beta}$, so that different dimensions are pooled differently.
Pooling is scale-invariant: $\aggrraw{\beta}_m \alpha \vec{x}_m = \alpha \aggrraw{\beta}_m \vec{x}_m$ for $\alpha \in \Real$.

For each rule $r$, we learn a scalar $\beta_r$,\footnote{It can be parameterized as $\beta=\exp b > 0$ (ensuring that aggregating positive numbers exceeds their max), or as $\beta=1+b^2 \geq 1$ (ensuring that the aggregate of positive numbers also does not exceed their sum).  Our present experiments do the latter.}  
 and use $\aggr{\beta_r}$ in \eqref{eqn:rule_emb}.

\subsection{Probabilities and Intensities}\label{sec:prob}\label{sec:event_math}

When a fact \mvar{h}
has been declared by {\isevent} to represent an event type, we need it to have not only an embedding but also a positive probability.  We extend our setup by appending an extra row to the matrix $\vec{W}_r$ in \eqref{eqn:rule_emb}, leading to an extra element in the column vectors $\depval{\mvar{h}}{r}$.  We then pass only the first $D_{\mvar{h}}$ elements of $\sum_r \depval{\mvar{h}}{r}$ through $\tanh$, obtaining the same $\sem{\mvar{h}}$ as \cref{eqn:term_emb} gave before.  We pass the one remaining element through an $\exp$ function to obtain $\lambda_{\mvar{h}} > 0$.

Recall that for neural Datalog through time (\cref{sec:ndtt}), all these quantities, including $\lambda_{\mvar{h}}$, vary with the time $t$.  \Copy{discretetimenorm}{To model a discrete-time event sequence, define the \defn{probability} of an event of type $\mvar{h}$ at time step $t$ to be proportional to $\lambda_{\mvar{e}}(t)$, normalizing over all event types that are possible then.}  This imitates the softmax distributions in other neural sequence models \citep{mikolov-10-rnnlm,sundermeyer-12-lstm}.

When time is continuous, as in our experiments (\cref{sec:exp}), we need instantaneous probabilities.  We take $\lambda_{\mvar{h}}(t)$ to be the (Poisson) \defn{intensity} of $\mvar{h}$ at time $t$: that is, it models the limit as $dt \rightarrow 0^+$ of the expected \emph{rate} of $\mvar{h}$ on the interval $[t,t+dt)$ (i.e., the expected number of occurrences of $\mvar{h}$ divided by $dt$).  This follows the setup of the neural Hawkes process \cite{mei-17-neuralhawkes}.  Also following that paper, we replace $\exp(x) > 0$ in the above definition of $\lambda_{\mvar{h}}$ with the function $\softplus_\timescale(x) = \timescale \log (1+\exp(x/\timescale)) > 0$.  We learn a separate temporal scale parameter $\timescale$ for each functor and use the one associated with the functor of $\mvar{h}$.

In both discrete and continuous time, the exact model likelihood (\cref{sec:train}) will involve a summation (at each time $t$) over the finite set of event types (\cref{sec:finite}) that are possible at time $t$.\looseness=-1

\Cref{app:simult} offers an extension to simultaneous events.

\subsection{Updates Through Time}\label{sec:ndtt_math}\label{sec:zerodim}

We now add an LSTM-like component so that each atom will track the sequence of events that it has ``seen''---that is, the sequence of events that updated it via $\see$ rules (\cref{sec:dltt}).  Recall that an LSTM is constructed from \defn{memory cells} that can be increased or decreased as successive inputs arrive.  

Every atom $\mvar{h}$ has a \defn{cell block} $\seecell{\mvar{h}} \in \Real^{D_{\mvar{h}}} \cup \{\nullval\}$.  When $\seecell{\mvar{h}} \neq \nullval$, we augment $\mvar{h}$'s embedding formula \eqref{eqn:term_emb} to\footnote{\label{fn:eventprob}Recall from \cref{sec:prob} that if $\mvar{h}$ is an event, we extend $\sem{\mvar{h}}$ with an extra dimension to carry the probability.  For \cref{eqn:term_emb_cell} to work, we must likewise extend $\seecell{\mvar{h}}$ with an extra cell (when $\seecell{\mvar{h}} \neq\nullval$).}
\begin{align}\label{eqn:term_emb_cell}
\sem{\mvar{h}} \;&\defeq\; \tanh \big(\, \seecell{\mvar{h}} + \sum_r \depval{\mvar{h}}{r} \big) \; \in (-1,1)^{D_{\mvar{h}}}
\end{align}
Properly speaking, $\sem{\mvar{h}}$, $\seecell{\mvar{h}}$, and $\depval{\mvar{h}}{r}$ are all functions of $t$.

At times when $\seecell{\mvar{h}}=\nullval$, we like to say that $\mvar{h}$ is \defn{docked}.  Every atom $\mvar{h}$ is docked initially (at $t=0$), but may be \defn{launched} through an update of type \eqref{eqn:see_launch}, which ensures that $\seecell{\mvar{h}} \neq \nullval$ and thus $\sem{\mvar{h}} \neq \nullval$ by \eqref{eqn:term_emb_cell}.  $\mvar{h}$ is subsequently \defn{adrift} (and remains a fact)
until it is docked again through an update of type \eqref{eqn:see_dock},
which sets $\seecell{\mvar{h}}=\nullval$.

How is $\seecell{\mvar{h}}$ updated by an event (or events\footnote{\label{fn:simultevents}If exogeneous events are used (\cref{sec:exo}), then the instantiations in \eqref{eqn:rule_preupdate} could include multiple events $\mvar{e}$ that occurred at time $s$.%
}) occurring at time $s$?  Suppose the \rth rule is an update rule of type \eqref{eqn:see_launch}.  Consider its instantiations  \code{\mvar{h} \see \mvar{e}, \mvar{g}$_1$,$\ldots$,\mvar{g}$_N$} (if any) with head $\mvar{h}$, such that $\mvar{e}$ occurred at time $s$ and $\mvar{g}_1, \ldots, \mvar{g}_N$ are all facts at time $s$.  For the $m$\th instantiation,
define
\begin{align}\label{eqn:rule_preupdate}
\seeval{\mvar{h}}{rm} & \defeq \vec{W}_r \underbrace{[1; \sem{\mvar{e}}; \sem{\mvar{g}_1}; \ldots; \sem{\mvar{g}_N}]}_{\clap{\text{\scriptsize concatenation of column vectors}}} 
\end{align}
where all embeddings are evaluated at time $s$, and $\vec{W}_r$ is again a conformable matrix associated with the \rth rule.  We now explain how to convert $\seeval{\mvar{h}}{rm}$ to an \defn{update vector} $\seeupd{\mvar{h}}{rm}$, and how all update vectors combine to modify $\seecell{\mvar{h}}$.\looseness=-1

\paragraph{Discrete-time setting.} Here we treat the update vectors $\seeupd{\mvar{h}}{rm}$ as increments to $\seecell{\mvar{h}}$.  To update $\seecell{\mvar{h}}$ from time $s$ to time $t=s+1$, we pool these increments within and across rules (much as in \eqref{eqn:term_emb}--\eqref{eqn:rule_emb}) and increment by the result:
\begin{align}\label{eqn:cell_update}
\seecell{\mvar{h}} \;&\pluseq\; \sum_r \aggr{\beta_r}_m \seeupd{\mvar{h}}{rm}
\end{align}
We skip the update \eqref{eqn:cell_update} if $\mvar{h}$ has no update vectors.  If we apply \eqref{eqn:cell_update}, we first set $\seecell{\mvar{h}}$ to $\vec{0}$ if it is $\nullval$ at time $s$, or has just been set to $\nullval$ at time $s$ by a \eqref{eqn:see_dock} rule (docking).

How is $\seeupd{\mvar{h}}{rm}$ obtained? In an ordinary LSTM \cite{hochreiter-97-lstm},
a cell block $\seecell{\mvar{h}}$ is updated by
\begin{align}
\seecell{\mvar{h}}\new &= \vec{f}\cdot \seecell{\mvar{h}}\old + \vec{i}\cdot (2\vec{z}-1) \label{eqn:lstm_c_update}
\intertext{corresponding to an increment}
\seecell{\mvar{h}} \;&\pluseq\; (\vec{f}-1) \cdot \seecell{\mvar{h}} + \vec{i}\cdot (2\vec{z}-1) \label{eqn:lstm_c_increment}
\end{align}
where the forget gates $\vec{f}$, input gates $\vec{i}$, and inputs $\vec{z}$ are all in $(0,1)^{D_\mvar{h}}$.  Thus, we define $\seeupd{\mvar{h}}{rm}$ as the right side of \eqref{eqn:lstm_c_increment} when $(\vec{f};\, \vec{i};\, \vec{z}) \defeq \sigma(\seeval{\mvar{h}}{rm})$, with $\seeval{\mvar{h}}{rm} \in \Real^{3D_{\mvar{h}}}$ from \eqref{eqn:rule_preupdate}.

A small difference from a standard LSTM is that our updated cell values $\seecell{\mvar{h}}$ are transformed into equally many output values $\sem{\mvar{h}}$ via \cref{eqn:term_emb_cell}, instead of through   $\tanh$ and output gates.  A more important difference is that in a standard LSTM, the model's state is a single large cell block.  The state update when new input arrives depends on the entire current state.  Our innovation is that the update to $\seecell{\mvar{h}}$ (a \emph{portion} of the model state) depends on only a relevant \emph{portion} of the current state, namely $[\sem{\mvar{e}}; \sem{\mvar{g}_1}; \ldots; \sem{\mvar{g}_N}]$.  If there are many choices of this portion, \eqref{eqn:cell_update} pools their effects across instantiations and sums them across rules.

\paragraph{Continuous-time setting.}  Here we use the continuous-time LSTM as defined by \citet{mei-17-neuralhawkes}, in which cells \defn{drift} between updates to record the passage of time.  Each cell drifts according to some parametric function.  We will update a cell's parameters just at times when a \emph{relevant} event happens.  A fact's embedding $\sem{\mvar{h}}(t)$ at time $t$ is still given by \eqref{eqn:term_emb_cell}, but $\seecell{\mvar{h}}(t)$ in that equation is given by $\seecell{\mvar{h}}$'s parametric functions as most recently updated (at some earlier time $s < t$). \Cref{app:continuous_cells} reviews the simple family of parametric functions used in the continuous-time LSTM, and specifies how we update the parameters using a collection of update vectors $\seeupd{\mvar{h}}{rm}$ obtained from the $\seeval{\mvar{h}}{rm}$.

\paragraph{Remark.} It is common for event atoms $\mvar{e}$ to have $D_{\mvar{e}}=0$.  Then they still have  time-varying probabilities (\cref{sec:prob})---often via $\dep$ rules whose conditions have time-varying embeddings---but have no embeddings.  Even so, different events will result in different updates.  This is thanks to Datalog's pattern matching: the event's atom $\mvar{e}$ controls which update rules $\code{\mvar{head} \see \mvar{event}, \mvar{condits\ldots}}$ it triggers, and with what head and condition atoms (since variables in $\mvar{event}$ are \emph{reused} elsewhere in the rule).  The update to the head atom then depends on the parameters of the selected rules and the current embeddings of their condition atoms.\looseness=-1

\vspace{-4pt}
\section{Training and Inference}\label{sec:event}\label{sec:train}\label{sec:mbr_brief}

Suppose we observe that the events on time interval $[0, \dur]$ are
$\mvar{e}_1, \ldots, \mvar{e}_I$ at respective times $t_1 < \cdots < t_I$.
In the \emph{continuous-time} setting, the log-likelihood of the parameters is
\vspace{-12pt}
\begin{align}\label{eqn:loglikcont}
\ell \;&\defeq\; \sum_{i=1}^I \log \inten{\mvar{e}_i}{t_i} - \int_{t=0}^{\dur} \inten{}{t} \,dt
\end{align} 
where $\inten{}{t} \defeq \sum_{\mvar{e}\in \set{E}(t)} \inten{\mvar{e}}{t}$ and $\set{E}(t)$ is the set of event types that are possible at time $t$.  We can estimate the parameters by locally maximizing $\ell$ using any stochastic gradient method.  Details are given in \cref{app:loglik_details}, including 
Monte Carlo approximations to the integral.  In the \emph{discrete-time} setting,\footnote{Here each time $t$ has exactly one event (possibly just a \evto{none} event), as the event probabilities sum to 1.  So $I=T$ and $t_i=i$.}
 the integral is replaced by $\sum_{t=1}^\dur \log \inten{}{t}$.

Given the learned parameters, we may wish to make a minimum Bayes risk prediction about the next event given the past history. 
A recipe can be found in \cref{app:mbr}. 

\vspace{-4pt}
\section{Related Work}\label{sec:related}

Past work \cite{sato-1995,ailog-2010,richardson-06-markov,PROBLOG-2007,PPDL-2017} has used logic programs to help define probabilistic relational models \cite{getoor-07-srl}.  These models do not make use of vector-space embeddings or neural networks.  Nor do they usually have a temporal component. However, some other (directed) graphical model formalisms do allow the model architecture to be affected by data generated at earlier steps \cite{minka-08-gates,van-18-introduction}.\looseness=-1

Our ``neural Datalog through time'' framework uses a deductive database augmented with update rules to define and dynamically reconfigure the architecture of a neural generative model.
Conditional neural net structure has been used for natural language---e.g., conditioning a neural architecture on a given syntax tree or string \citep{andreas-16-compose,lin-2019-fst}.  Also relevant are neural architectures that use external read-write memory to achieve coherent sequential generation, i.e., their decisions are conditioned on a possibly symbolic record of data generated from the model at earlier steps  \citep{graves-14-turing,graves-16-hybrid,weston-15-memory,sukhbaatar-15-end,kumar-16-ask,kiddon-16-global,dyer-2016-rnng,lample-19-large,xiao-19-grammatical}.
We generalize some such approaches by providing a logic-based specification language.

Many papers have presented domain-specific sequential neural architectures \citep{natarajan-08-logical,van-14-learning,shelton-14-markov,meek-14-toward,bhattacharjya-18-pgem,wang-19-factor}.
The models closest to ours are \defn{Know-Evolve} \citep{trivedi-17-know} and \defn{DyRep} \citep{trivedi-19-dyrep}, which exploit explicit domain knowledge about how structured events depend on and modify the neural states of their participants.
DyRep also conditions event probabilities on a temporal graph encoding binary relations among a fixed set of entities.
In \cref{sec:exp}, we will demonstrate that fairly simple programs in our framework can substantially 
outperform these strong competitors by leveraging even richer types of knowledge, e.g.:
\circone
Complex $n$-ary relations among entities that are constructed by join, disjunction, and recursion (\cref{sec:datalog}) and have derived embeddings (\cref{sec:ndl}). 
\circtwo
Updates to the set of possible events (\cref{sec:eventfacts}).
\circthree 
Embeddings of entities and relations that reflect selected past events (\cref{sec:ndtt,sec:continue}).

\vspace{-4pt}
\section{Experiments}\label{sec:exp}

In several continuous-time
domains, we exhibit informed models specified using neural Datalog through time (NDTT).  We evaluate these models on their held-out log-likelihood, and on their success at predicting the time and type of the next event.  We compare with the unrestricted neural Hawkes process (NHP) and with Know-Evolve (KE) and DyRep.  Experimental details are given in \cref{app:exp_details}.

We implemented our NDTT framework using PyTorch \citep{paszke-17-pytorch} and pyDatalog \citep{pydatalog}.  We then used it to implement our individual models---and to reimplement all three baselines, after discussion with their authors, to ensure a controlled comparison.  
Our code and datasets are available at the URL given in \cref{sec:lang}.

\subsection{Synthetic Superposition Domain}\label{sec:superposition}\label{sec:synthetic}

\begin{figure}
	\centering
	\begin{subfigure}[t]{0.48\linewidth}
		\includegraphics[width=1.00\linewidth]{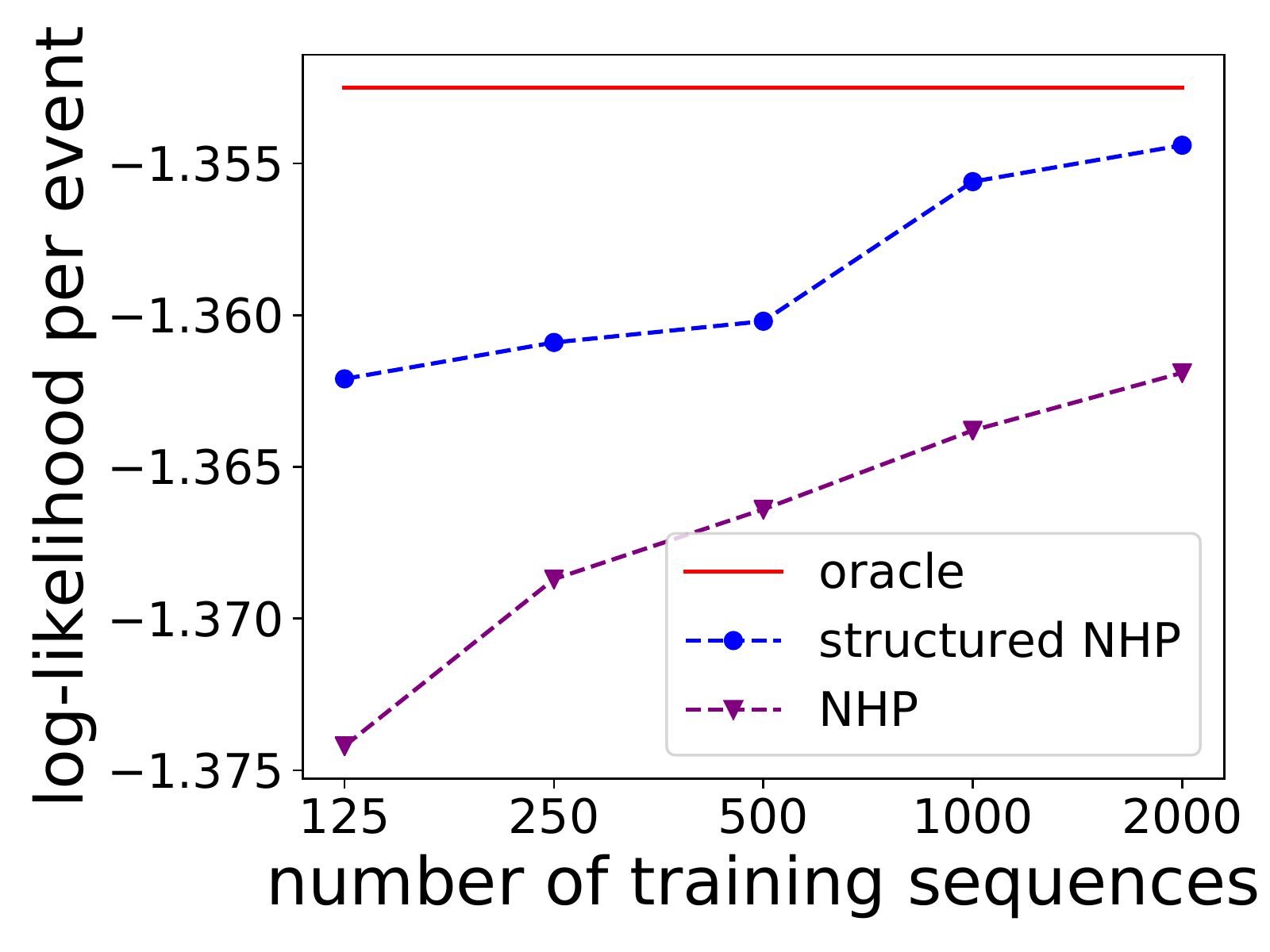}
		\vspace{-16pt}
		\caption{$M=8$}\label{fig:m8}
	\end{subfigure}
	~
	\begin{subfigure}[t]{0.48\linewidth}
		\includegraphics[width=1.00\linewidth]{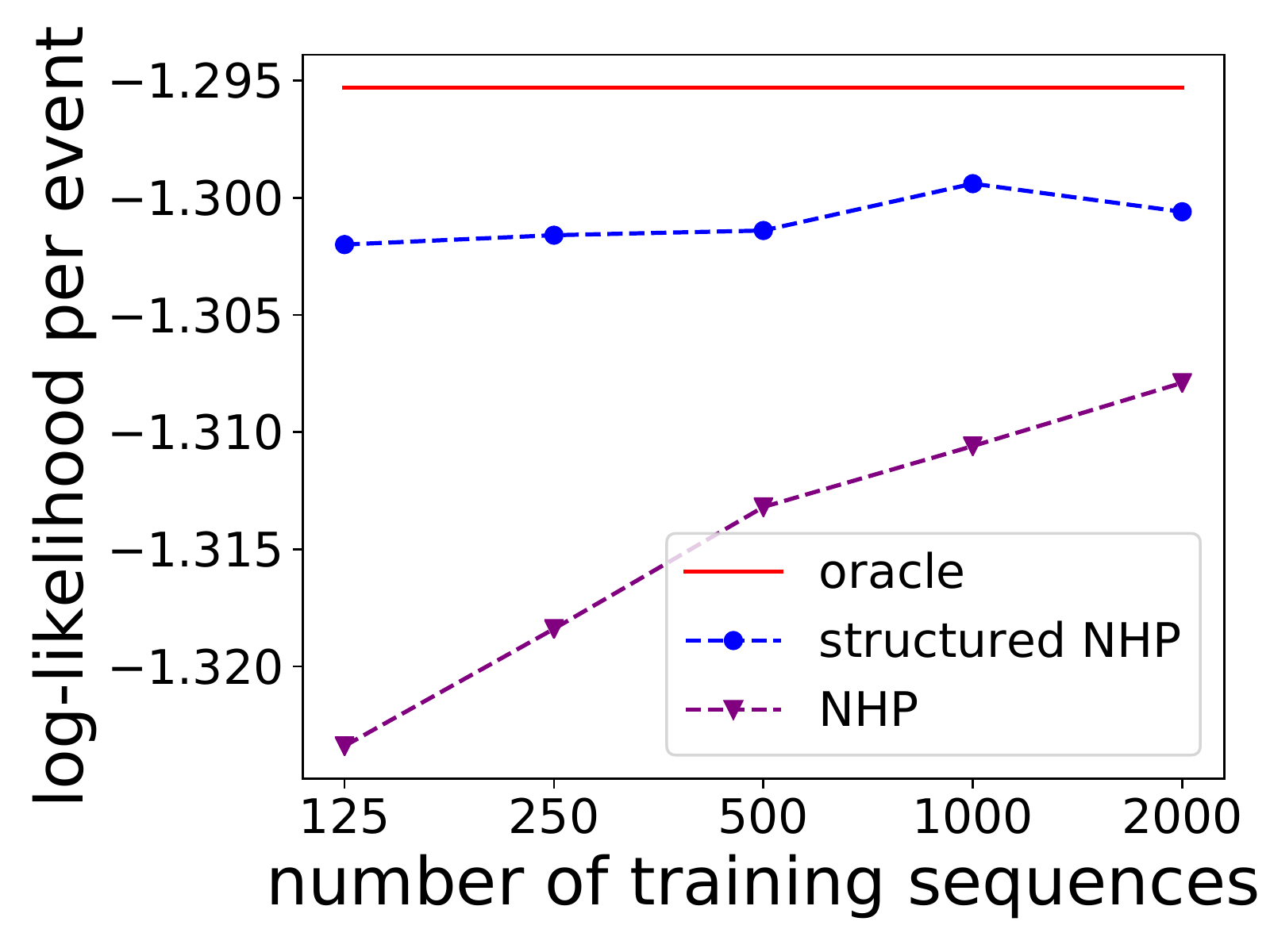}
		\vspace{-16pt}
		\caption{$M=16$}\label{fig:m16}
	\end{subfigure}
	\vspace{-8pt}
	\caption{Learning curves of structured model \bluedot and NHP \greentriangle, on sequences drawn from the structured model. 
		The former is significantly better at each training size ($p < 0.01$, paired perm.\@ test).
	}
	\label{fig:learncurve_nhp}
        \vspace{-1pt}
\end{figure}

The activities of strangers rarely influence each other, even if they are all observed within a single sequence. 
We synthesized a domain where each sequence is a superposition of data drawn from $M$ different processes that do not interact with one another at all. Each process generates events of $N$ types, so there are $M N$ total event types \code{\evt{e}(M,N)}.

\begin{minipage}[t]{0.49\linewidth}
\begin{lstlisting}[name=nhp,firstnumber=auto]
@\code{\boo{is\_process}(1).}@ @\label{line:process_fact}\progvdots@
@\code{\boo{is\_process}($M$).}@
\end{lstlisting}
\end{minipage}
\hfill
\begin{minipage}[t]{0.49\linewidth}
\begin{lstlisting}[name=nhp,firstnumber=auto]
@\code{\boo{is\_type}(1).}\progvdots@
@\code{\boo{is\_type}($N$).}@
\end{lstlisting}
\end{minipage}

The baseline model is a neural Hawkes process (NHP).
It assigns to each event type a separate embedding\footnote{The list of facts like \cref{line:emb_fact_1,line:emb_fact_2} can be replaced by a single rule if we use ``parameter names'' as explained in \cref{app:param_share}. }
\begin{lstlisting}[name=nhp,firstnumber=auto]
@\code{\dep }\code{ {\embed}(\blk{is\_event}, 8).} @
@\code{\blk{is\_event}(1,1) {\dep} \boo{is\_process}(1), \boo{is\_type}(1).} @ @\label{line:emb_fact_1}@
@\code{\blk{is\_event}(1,2) {\dep} \boo{is\_process}(1), \boo{is\_type}(2).} \progvdots@ @\label{line:emb_fact_2}@
\end{lstlisting}
This unrestricted model allows all event types to influence one another by depending on and affecting a \code{\blk{world}} state:
\begin{lstlisting}[name=nhp,firstnumber=auto]
@\code{\dep }\code{ {\isevent}(\evto{e}, 0).} @ @\label{line:kdim0}@
@\code{\dep }\code{ {\embed}(\blk{world}, 8).}@ @\label{line:world_start}@
@\code{\evto{e}(M,N)}\code{ \dep }\code{ \blk{world}, \boo{is\_process}(M), \boo{is\_type}(N).}@ @\label{line:dep_nhp}@
@\code{\blk{world}}\code{ \see }\code{ \exoo{init}.}@
@\code{\blk{world}}\code{ \see }\code{ \evto{e}(M,N), \blk{is\_event}(M,N), \blk{world}.}@ @\label{line:see_nhp}@ @\label{line:world_end}@
\end{lstlisting}
Note that \code{\evto{e}(M,N)} in \cref{line:see_nhp} has no embedding, since any such embedding would vary along with the probability.  As explained in \cref{sec:zerodim}, \cref{line:see_nhp} instead uses \code{\evto{e}(M,N)} to draw in the embedding of \code{\blk{is\_event}(M,N)}, which does not depend on \blk{world} so is static, as called for by the standard NHP.

\begin{figure*}[!ht]
	\begin{center}
		\begin{minipage}[t]{0.75\linewidth}
            \vspace{0pt}
			\begin{subfigure}[]{0.99\linewidth}
				\begin{center}
					\includegraphics[width=0.31\linewidth]{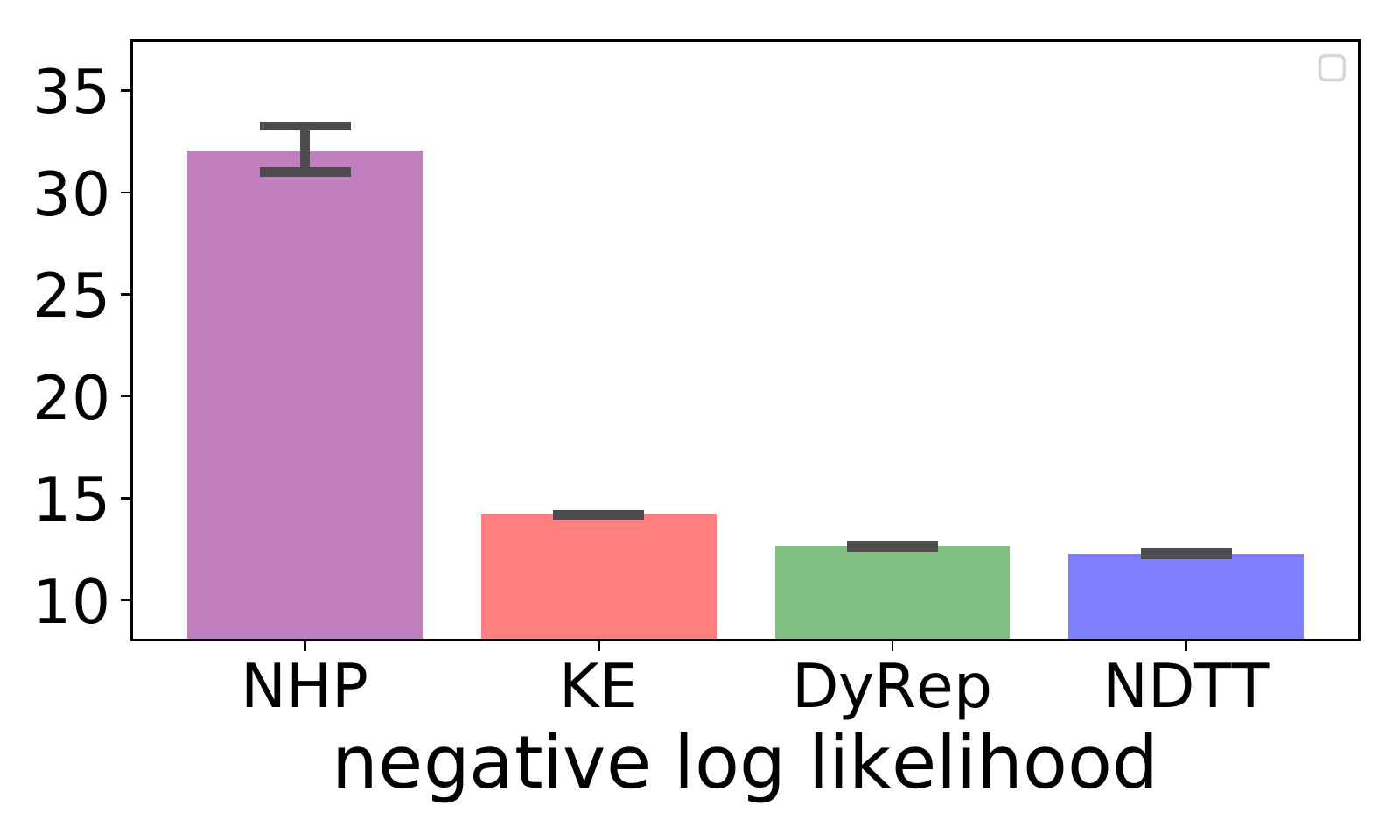}
					~
					\includegraphics[width=0.31\linewidth]{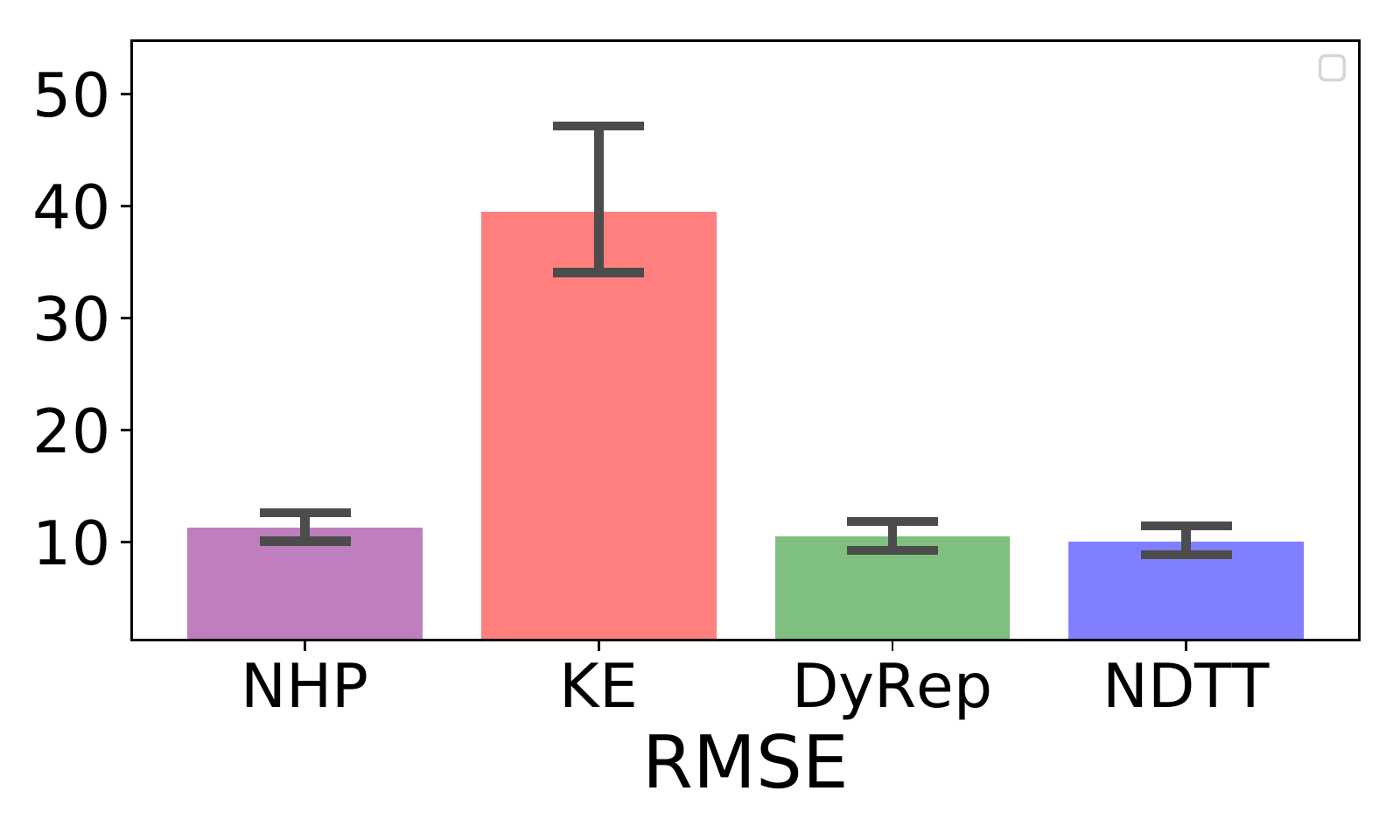}
					~
					\includegraphics[width=0.31\linewidth]{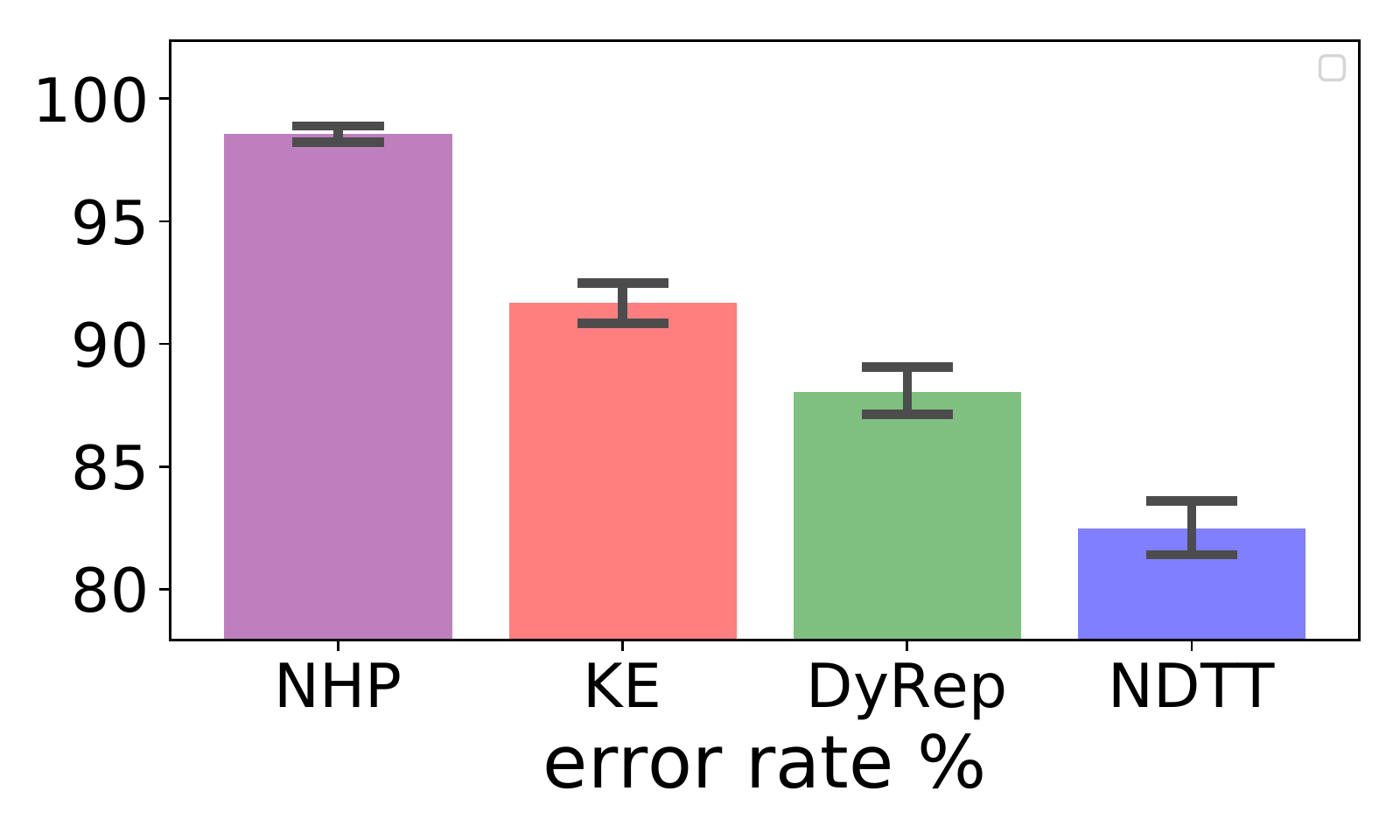}
					\vspace{-8pt}
					\caption{IPTV Dataset}\label{fig:pred_iptv}
				\end{center}
			\end{subfigure}

			\vspace{-2pt}

			\begin{subfigure}[]{0.99\linewidth}
				\begin{center}
					\includegraphics[width=0.31\linewidth]{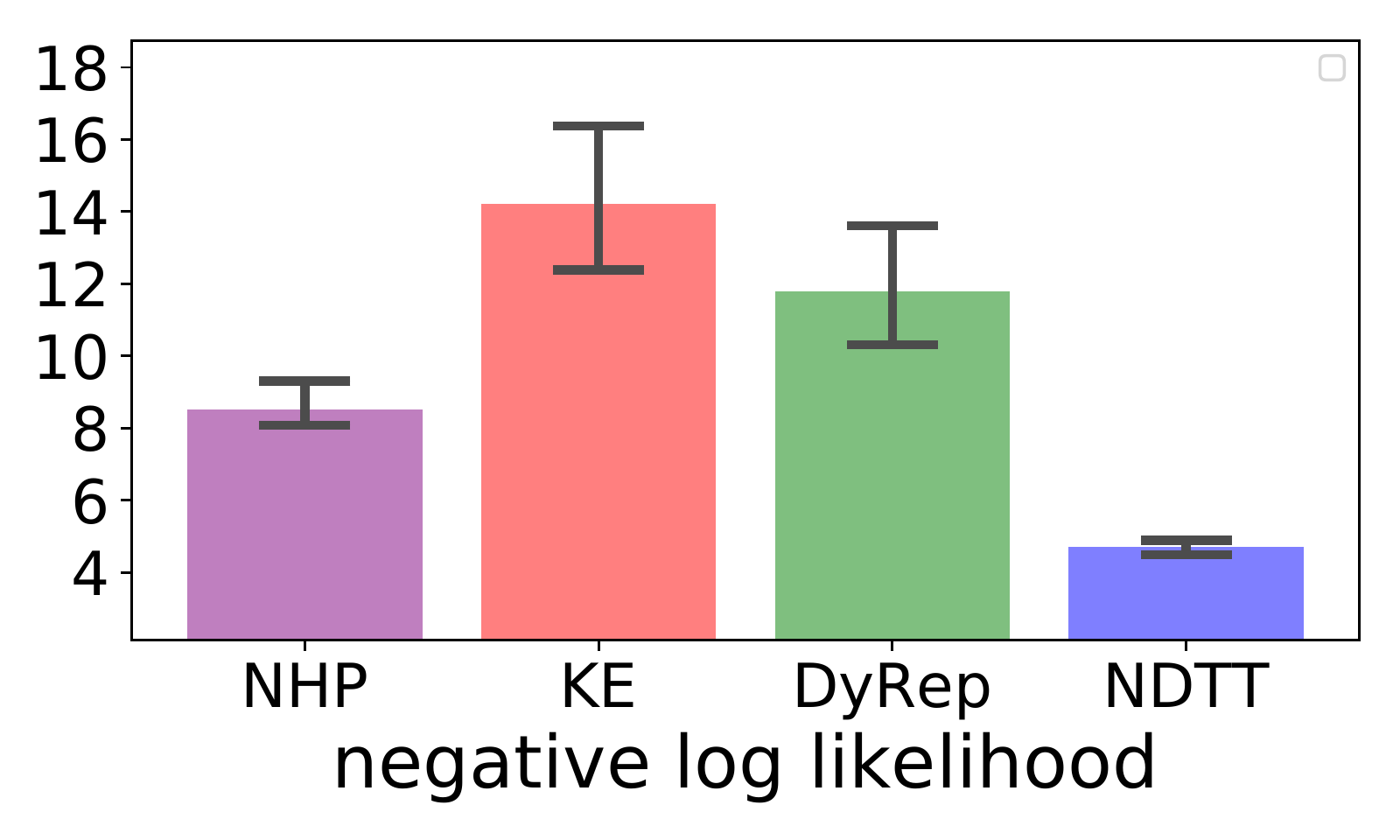}
					~
					\includegraphics[width=0.31\linewidth]{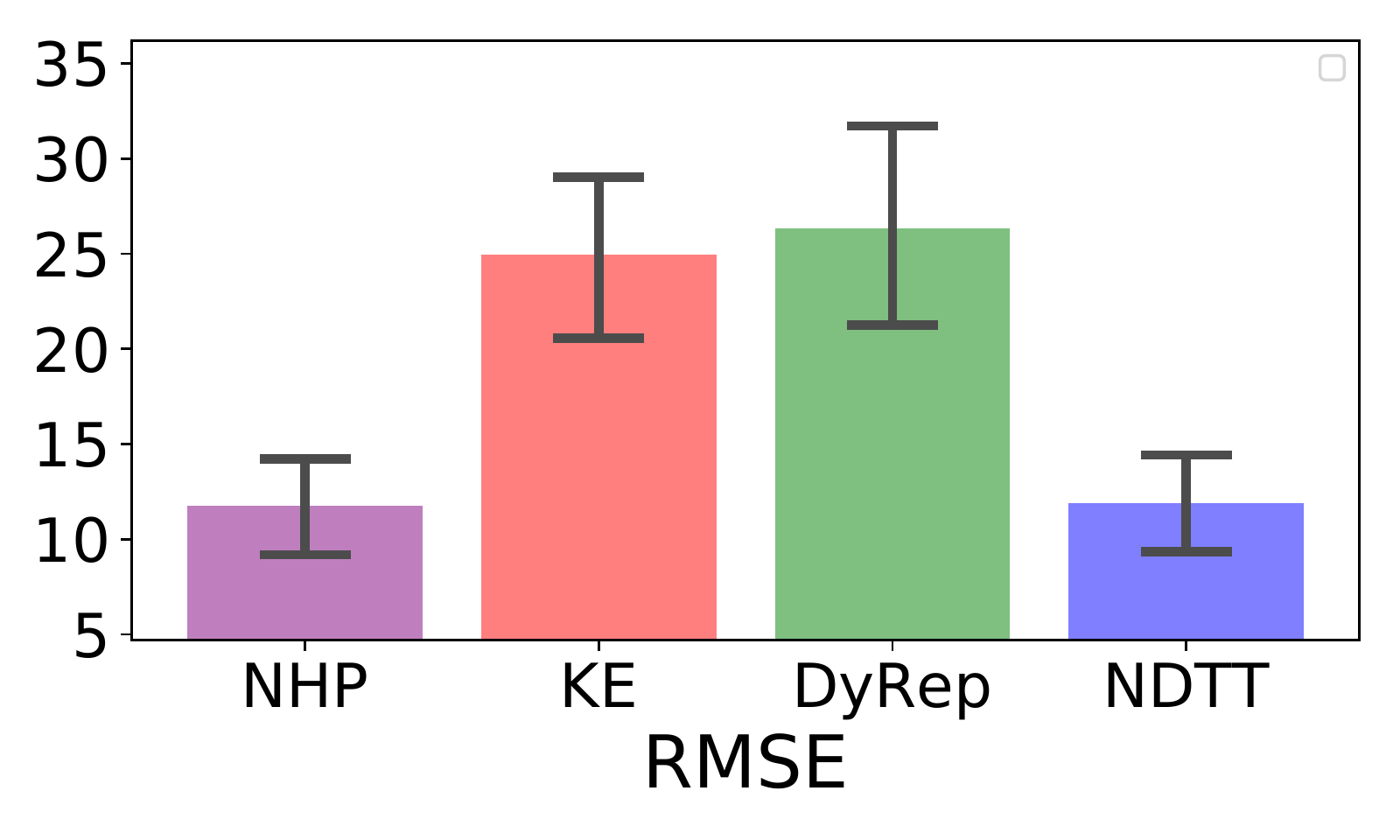}
					~
					\includegraphics[width=0.31\linewidth]{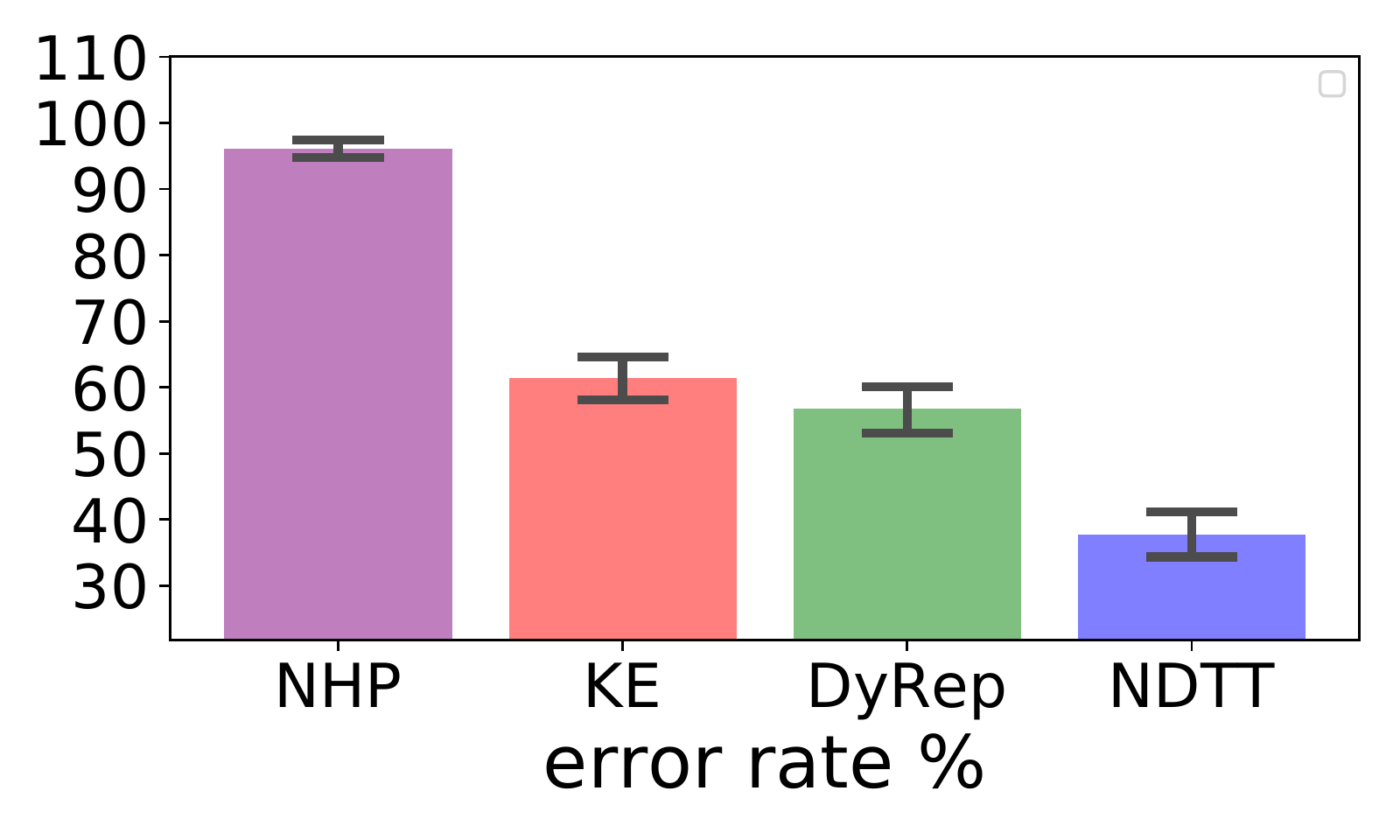}
					\vspace{-8pt}
					\caption{RoboCup Dataset}\label{fig:pred_robocup}
				\end{center}
			\end{subfigure}
		\end{minipage}
		\hfill
		\begin{minipage}[t]{.24\linewidth}
            \vspace{-5pt}
			\caption{Evaluation results with 95\% bootstrap confidence intervals on the real-world datasets of our Datalog program vs.\ the neural Hawkes process (NHP), KnowEvolve (KE) and DyRep. The RMSE is the root of mean squared error for predicted time.  Error rate \% denotes the fraction of incorrect predictions of the watched TV program (in IPTV) or the specific player (in RoboCup), given the event time.}
			\label{fig:pred_results}
		\end{minipage}
	\vspace{-14pt}
	\end{center}
\end{figure*}

To obtain a \emph{structured} NHP that recognizes that events from different processes cannot influence each other, 
we replace \code{\blk{world}} with multiple \code{\blk{local}} states: each \code{\evto{e}(M,N)} only interacts with \code{\blk{local}(M)}.
Replace \crefrange{line:world_start}{line:world_end} with
\begin{lstlisting}[name=nhp,firstnumber=auto]
@\code{\dep }\code{ {\embed}(\blk{local}, 8).}@
@\code{\evto{e}(M,N) \dep \blk{local}(M), \boo{is\_type}(N).}@
@\code{\blk{local}(M) \see \exoo{init}, \boo{is\_process}(M).}@
@\code{\blk{local}(M) \see \evto{e}(M,N), \blk{is\_event}(M,N), \blk{local}(M).}@
\end{lstlisting}

For various small $N$ and $M$ values (see \cref{app:synthetic_details}), we randomly set the parameters of the structured NHP model and draw training and test sequences from this distribution.
We then generated learning curves by training the correclty structured model versus the standard NHP on increasingly long prefixes of the training set, and evaluating them on held-out data. 
\Cref{fig:learncurve_nhp} shows that although NHP gradually improves its performance as more training sequences become available, 
the structured model unsurprisingly learns faster, e.g., only 1/16 as much training data to achieve a higher likelihood.  In short, it helps to use domain knowledge of which events come from which processes.

\begin{figure*}[!ht]
	\begin{center}
		\begin{minipage}[]{0.76\linewidth}
			\includegraphics[width=0.31\linewidth]{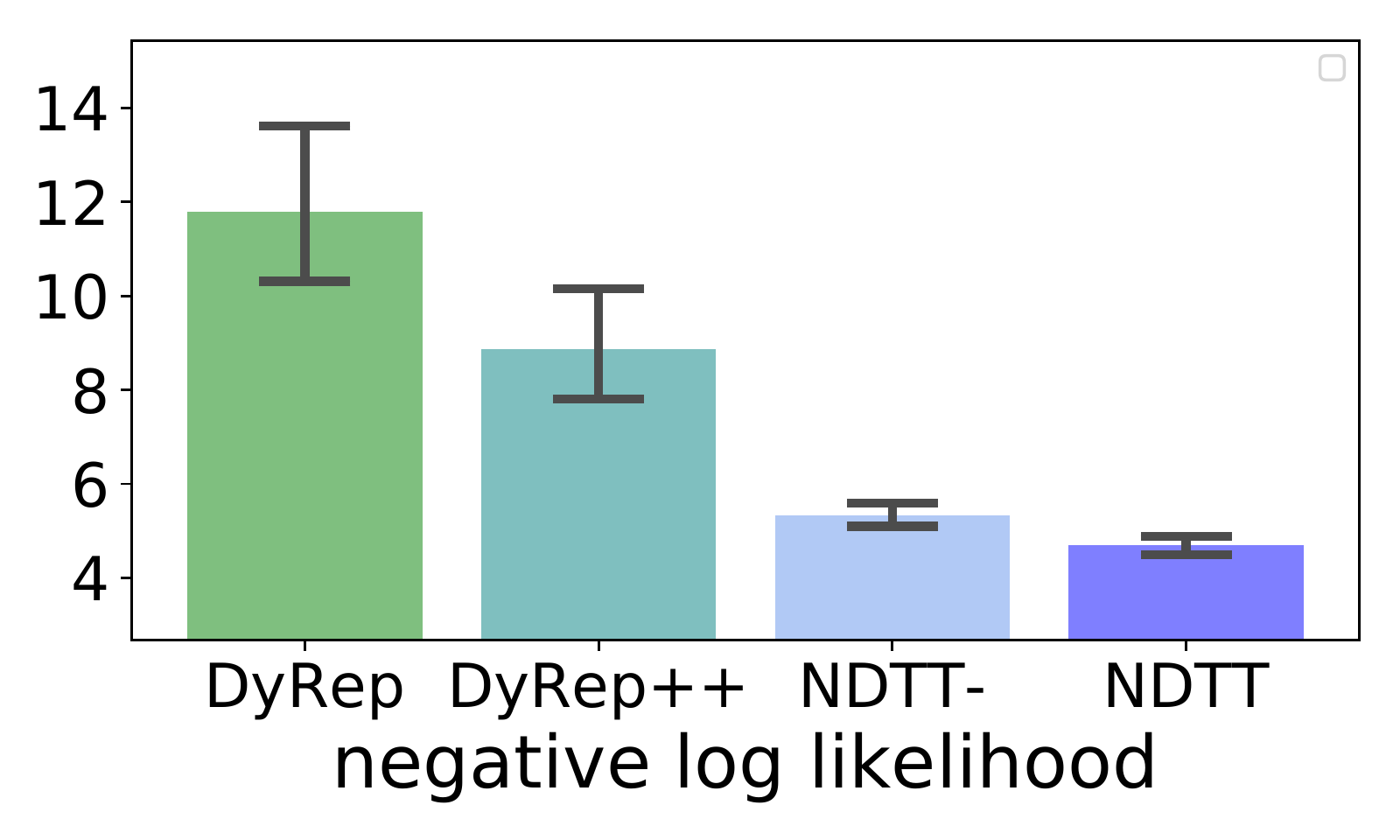}
			~
			\includegraphics[width=0.31\linewidth]{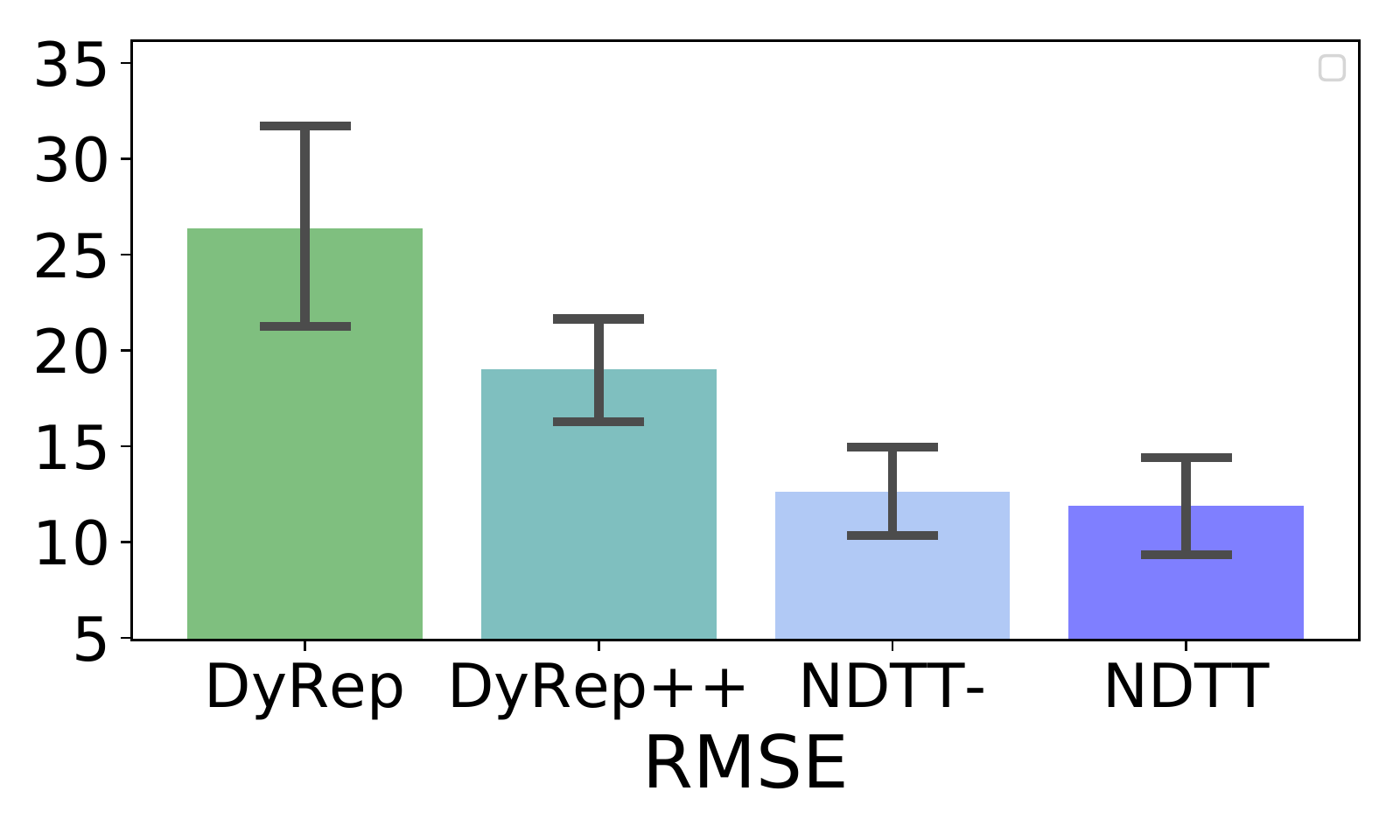}
			~
			\includegraphics[width=0.31\linewidth]{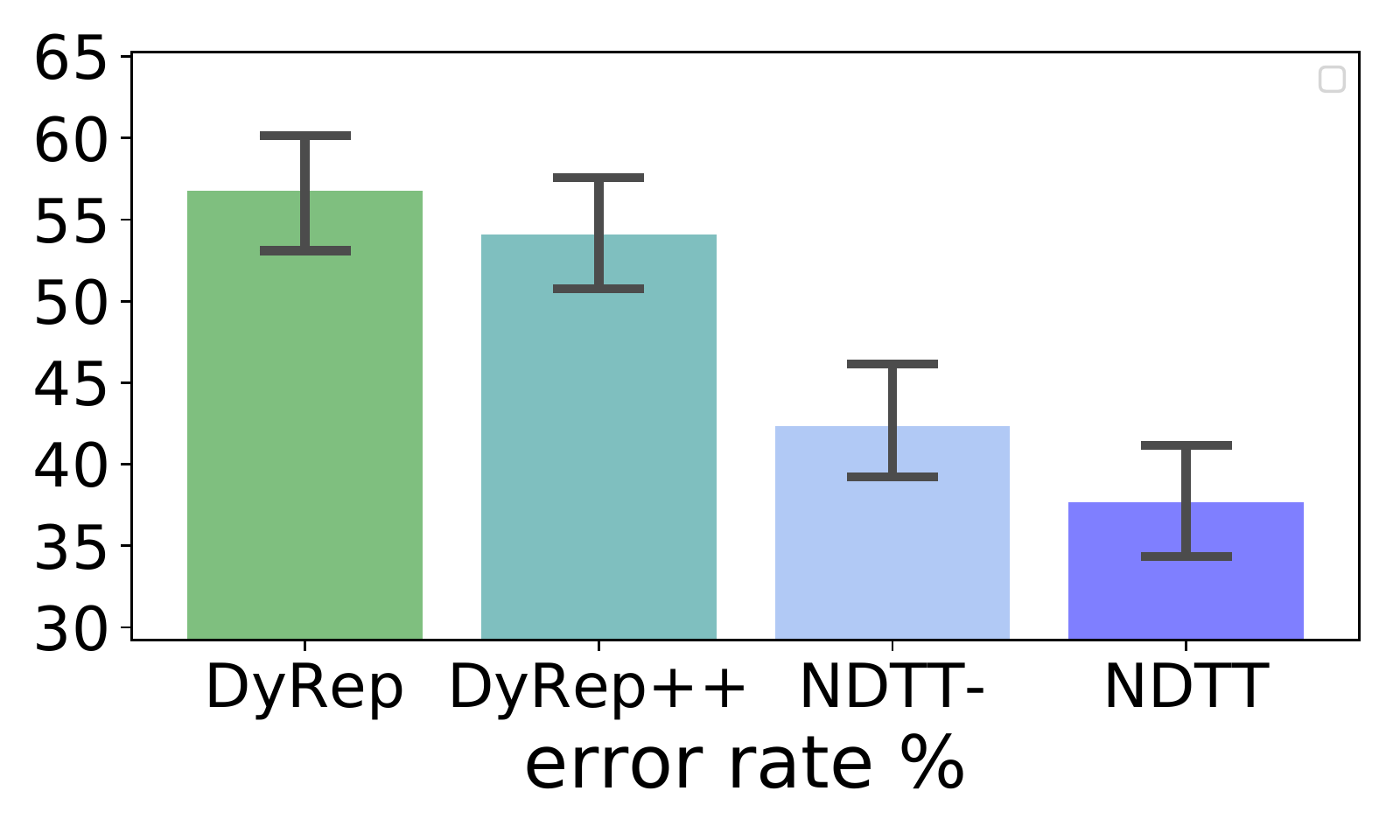}
			\vspace{-6pt}
		\end{minipage}
		\hfil
		\begin{minipage}[]{0.23\linewidth}
			\vspace{-14pt}
			\caption{Ablation study in the RoboCup domain. ``DyRep++'' has the same {\see} rules as our structured model and ``NDTT$-$'' uses 0-dimensional \code{\blk{team}} embeddings.}
			\label{fig:pred_robocup_ablation}	
		\end{minipage}
		\vspace{-16pt}
	\end{center}
\end{figure*}

\subsection{Real-World Domains: IPTV and RoboCup}\label{sec:iptv}\label{sec:robocup}

\paragraph{IPTV Domain \textnormal{\citep{xu-18-online}}.}
This dataset contains records of 1000 users watching 49 TV programs over the first 11 months of 2012.  Each event has the form \code{\evt{watch}(\dvar{U},\dvar{P})}.  Given each prefix of the test event sequence, we attempted to predict the next test event's time $t$, and to predict its program \code{\dvar{P}} given its actual time $t$ and user \code{\dvar{U}}.\looseness=-1

We exploit two types of structural knowledge in this domain. First, each program \code{\dvar{P}} has (exactly) 5 out of 22 genre tags such as \code{action}, \code{comedy}, \code{romance}, etc.  We encode these as known static facts \code{\boo{has\_tag}(\dvar{P},\dvar{T})}.  We allow each tag's embedding $\sema{tag}{\dvar{T}}$ to not only influence the embedding of its programs (\cref{line:tag_dep}) but also track which users have recently watched programs with that tag (\cref{line:tag_see}):
\begin{lstlisting}[name=iptv,firstnumber=auto]
@\code{\blk{program}(\dvar{P}) \dep \boo{has\_tag}(\dvar{P},\dvar{T}), \blk{tag}(\dvar{T}).}@ @\label{line:tag_dep}@
@\code{\blk{tag}(\dvar{T}) \see \evt{watch}(\dvar{U},\dvar{P}), \boo{has\_tag}(\dvar{P},\dvar{T}).}@ @\label{line:tag_see}@
\end{lstlisting}
As a result, a program's embedding $\sema{program}{\dvar{P}}$ changes over time as its tags shift in meaning.

Second, there is a dynamic hard constraint that a program
cannot be watched until it is released, since only then
is it added to the database:
\begin{lstlisting}[name=iptv,firstnumber=auto]
@\code{\blk{program}(\dvar{P}) \see \exoo{release}(\dvar{P}).}@
@\code{\evt{watch}(\dvar{U},\dvar{P}) \dep \blk{user}(\dvar{U}), \blk{program}(\dvar{P}).}@ 
\end{lstlisting}
Here \code{\exoo{release}(\dvar{P})} is an exogenous event with no embedding.
More details can be found in \cref{app:iptv_details}, including full NDTT programs that specify the architectures used by the KE and DyRep papers and by our model.

\paragraph{RoboCup Domain \textnormal{\citep{chen-08-robocup}}.}
This dataset logs actions of soccer players such as \code{\evt{kick}(\dvar{P})} and \code{\evt{pass}(\dvar{P},\dvar{Q})} during RoboCup Finals 2001--2004.  There are 528 event types in total.  For each history, we made minimum Bayes risk predictions of the next event's time, and of that event's participant(s) given its time and action type.

Database facts change frequently in this domain.  The ball is transferred between robot players at a high rate:
\begin{lstlisting}[name=robocup,firstnumber=auto]
@\code{!\boo{has\_ball}(\dvar{P}) \see \evt{pass}(\dvar{P},\dvar{Q}).} \text{\% ball passed from \code{\dvar{P}}} @
@\code{\boo{has\_ball}(\dvar{Q}) \ \see \evt{pass}(\dvar{P},\dvar{Q}).} \text{\% ball passed to \code{\dvar{Q}}} @
\end{lstlisting}
which leads to highly dynamic constraints on the possible events (since only the ball possessor can \evt{kick} or \evt{pass}): 
\begin{lstlisting}[name=robocup,firstnumber=auto]
@\code{\evt{pass}(P,Q) \dep \boo{has\_ball}(P), \boo{teammate}(P,Q), \ldots} @
\end{lstlisting}
This example also illustrates how relations between players affect events: the ball can only be \code{\evt{pass}}ed to a \code{\boo{teammate}}.  Similarly, only an \code{\boo{opponent}} may \code{\evt{steal}} the ball:  
\begin{lstlisting}[name=robocup,firstnumber=auto]
@\code{\evt{steal}(Q,P) \dep \boo{has\_ball}(P), \boo{opponent}(P,Q), \ldots} @
\end{lstlisting}

We allow each event to update the states of involved players as both KE and DyRep do. 
We further allow the event observers such as the entire \code{\blk{team}} to be affected as well:
\begin{lstlisting}[name=robocup,firstnumber=auto]
@\code{\blk{team}(T) \see \evt{pass}(P,Q), \boo{in\_team}(P,T), \ldots .}@
\end{lstlisting}
so all players can be aware of this event by consulting their \code{\blk{team}} states.
More details can be found in \cref{app:robocup_details}, including our full Datalog programs.
The hard logical constraints on possible events are not found in past models.

\paragraph{Results and Analysis.} After training, we used minimum Bayes risk (\cref{sec:mbr_brief}) to predict events in test data (details in \cref{app:mbr}). \Cref{fig:pred_results} shows that our NDTT model enjoys consistently lower error than strong competitors, across datasets and prediction tasks.

NHP performs poorly in general since it doesn't consider any knowledge. 
KE handles relational information, but doesn't accommodate dynamic facts such as \code{\boo{released}(game\_of\_thrones)} and \code{\boo{has\_ball}(a8)} that reconfigure model architectures on the fly. 

In the IPTV domain, DyRep handles dynamic facts (e.g., newly released programs) and thus substantially outperforms KE.
Our NDTT model's moderate further improvement results from its richer {\dep} and {\see} rules related to \code{\blk{tag}}s. 

In the RoboCup domain, our reimplementation of DyRep allows deletion
of facts (player losing ball possession), whereas the original DyRep
only allowed addition of facts.  Even with this improvement, it performs much worse than our full NDTT model.  To understand why,
we carried out further ablation studies, finding that NDTT benefits from its hybridization of logic and neural networks.

\paragraph{Ablation Study I: Taking Away Logic.}\label{sec:logic_away}
In the RoboCup domain, we investigated how the model performance degrades if we remove each kind of rule from the NDTT model.
We obtained ``NDTT\texttt{-}'' by dropping the \code{\blk{team}} states, and ``DyRep++'' by not tracking the ball possessor. 
The latter is still an enhancement to DyRep because it adds useful {\see} rules: the first ``+'' stands for the {\see} rules in which some conditions are not neighbors of the head, and the second ``+'' stands for the {\see} rules that update event observers.%

As \cref{fig:pred_robocup_ablation} shows, both
ablated models outperform DyRep but underperform our full NDTT model.
DyRep++ is interestingly close to NDTT on the participant prediction, implying that its neural states learn to track who possesses the ball---though such knowledge is not tracked in the logical database---thanks to rich {\see} rules that see past events.

\paragraph{Ablation Study II: Taking Away Neural Networks.}
We also investigated how the performance of our structured model would change if we reduce the dimension of all embeddings to zero.  The model still knows logically which events are possible, but events of the same type are now more interchangeable.  The performance turns out to degrade greatly, indicating that the neural networks had been learning representations that are actually helpful for prediction. 
See \cref{app:neural_away_details} for discussion and experiments.

\vspace{-4pt}
\section{Conclusion}\label{sec:conclusion}

We showed how to specify a neural-symbolic probabilistic model simply by writing down the rules of a deductive database.  ``Neural Datalog'' makes it simple to define a large set of structured objects (``facts'') and equip them with embeddings and probabilities, using pattern-matching rules to explicitly specify which objects depend on one another.

To handle temporal data, we proposed an extended notation to support \emph{temporal} deductive databases.  ``Neural Datalog through time'' allows the facts, embeddings, and probabilities to change over time, both by gradual drift and in response to discrete events.  We demonstrated the effectiveness of our framework by generatively modeling irregularly spaced event sequences in real-world domains.

\section*{Acknowledgments}
We are grateful to Bloomberg L.P. for enabling this work through a Ph.D.\ Fellowship Award to the first author, and to the National Science Foundation for supporting the other JHU authors under Grant No.\@ 1718846.
We thank Karan Uppal, Songyun Duan and Yujie Zha from Bloomberg L.P.\ for helpful comments and support to apply the framework to Bloomberg's real-world data.  
We thank the anonymous ICLR reviewers for helpful comments on an earlier version of this paper, Hongteng Xu for such comments and also for additional data, and Rakshit Trivedi for insightful discussion about Know-Evolve and DyRep. 
Moreover, we thank NVIDIA Corporation for kindly donating two Titan X Pascal GPUs, and the state of Maryland for the Maryland Advanced Research Computing Center.

\bibliographystyle{icml2020_url}
\bibliography{structure-nhp-clean}
\clearpage
\appendix
\appendixpage

\section{Extensions to the Formalism}\label{app:extensions}

In this appendix, we consider possible extensions to our formalism.  These illuminate interesting issues, and the extensions are compatible with our overall approach to modeling.  Some of these extensions are already supported in our implementation at {\small \url{https://github.com/HMEIatJHU/neural-datalog-through-time}}, and more of them may be supported in future versions.

\subsection{Cyclicity}\label{app:cyclic}

Our embedding definitions in \cref{sec:ndl,sec:ndl_math} assumed that the proof graph was acyclic.  However, it is possible in general Datalog programs for a fact to participate in some of its own proofs.  

For example, the following classical Datalog program finds the nodes in a directed graph that are reachable from the node \code{start}:
\begin{lstlisting}[name=reachability,firstnumber=auto]
@\blk{reachable}(start).@ 
@\blk{reachable}(\dvar{V}) \dep \blk{reachable}(\dvar{U}), \blk{edge}(\dvar{U},\dvar{V}).@ 
\end{lstlisting}
In neural Datalog, the embedding of each fact of the form \code{\blk{reachable}(\dvar{V})} depends on all paths from \code{start} to \dvar{V}.  However, if $\dvar{V}$ appears on a cycle in the directed graph defined by the \blk{edge} facts, then there will be infinitely many such paths, and our definition of $\sema{reachable}{\dvar{V}}$ would then be circular.

\paragraph{Restricting to acyclic proofs.}  One could define embeddings and probabilities in a cyclic proof graph by considering only the acyclic proofs of each atom \mvar{h}.  This is expensive in the worst case, because it can exponentially increase the number of embeddings and probabilities that need to be computed.  Specifically, if $S$ is a (finite) set of atoms, let $\sem{\mvar{h}/S}$ denote the embedding constructed from acyclic proofs of $\mvar{h}$ that do not use any of the atoms in the finite set $S$.  We define $\sem{\mvar{h}/S}$ to be $\nullval$ if $\mvar{h} \in S$, and otherwise to be defined similarly to $\sem{\mvar{h}}$ but where \cref{eqn:rule_emb,eqn:rule_preupdate} are modified to replace each $\sem{\mvar{g}_i}$ with $\sem{\mvar{g}_i/(S\cup\{\mvar{h}\})}$.\footnote{For increased efficiency, one can simplify $S\cup\{\mvar{h}\}$ here to eliminate atoms that can be shown by static analysis or depth-first search not to appear in any proof of $\mvar{g}_i$.  This allows more reuse of previously computed $\sem{\cdot}$ terms and can sometimes prevent exponential blowup.  In particular, if it can be shown that all proofs of $\mvar{h}$ are acyclic, then $\sem{\mvar{h}/S}$ can always be simplified to $\sem{\mvar{h}/\emptyset}$ and the computation of $\sem{\mvar{h}/\emptyset}$ is isomorphic to the ordinary computation of $\sem{\mvar{h}}$; the algorithm then reduces to the ordinary algorithm from the main paper.}  As usual, these formulas skip pooling over instantiations where any $\sem{\cdot}$ values in the body are $\nullval$. The recursive definition terminates because $S$ grows at each recursive step but its size is bounded above (\cref{sec:finite}).  

In particular, this scheme defines $\sem{\mvar{h}/\emptyset}$, the \defn{acyclic embedding} of $\mvar{h}$,  which we consider to be an output of the neural Datalog program.  Similarly, in neural Datalog through time, the probability of an event \mvar{e} is derived from $\lambda_{\mvar{e}/\emptyset}$, which is computed in the usual way  (\cref{sec:prob}) as an extra dimension of the acyclic embedding $\sem{\mvar{e}/\emptyset}$.  

\paragraph{Forward propagation.}  This is a more practical approach, used by \citet{hamilton-17-gnn} to embed the vertices of a graph.  This method recomputes all embeddings in parallel, and repeats this for some number of iterations.  In our case, for a given time $t$, each $\sem{\mvar{h}}$ is initialized to $\vec{0}$, and at each iteration it is recomputed via the formulas of \cref{sec:ndl_math,sec:ndtt_math}, using the $\sem{\mvar{g}_i}$ values from the previous iteration (also at time $t$) and the cell block $\seecell{\mvar{h}}$ (determined by events at times $s < t$).

We suggest the following variant that takes the graph structure into account.  At time $t$, construct the (finite) Datalog proof graph, whose nodes are the facts at time $t$.  Visit its strongly connected components in topologically sorted order.  Within each strongly connected component $C$, initialize the embeddings to $\vec{0}$ and then recompute them in parallel for $|C|$ iterations.  If the graph is acyclic, so that each component $C$ consists of a single vertex, then the algorithm reduces to an efficient and exact implementation of \cref{sec:ndl_math,sec:ndtt_math}.  In the general case, visiting the components in topologically sorted order means that we wait to work on component $C$ until its strictly upstream nodes have ``converged,'' so that the limited iterations on $C$ make use of the best available embeddings of the upstream nodes.  By choosing $|C|$ iterations for component $C$, we ensure that all nodes in $C$ have a chance to communicate: information has the opportunity to flow end-to-end through all cyclic or acyclic paths of length $< |C|$, and this is enough to include all acyclic paths within $C$.  Note that the embeddings computed by this algorithm (or by the simpler method of \citet{hamilton-17-gnn}) are well-defined: they depend only on the graph structure, not on any arbitrary ordering of the computations.

\subsection{Negation in Conditions}\label{app:negbody}

A simple extension to our formalism would allow negation in the body of a rule (i.e., the part of the rule to the right of {\dep} or {\see}).  In rules of the form \eqref{eqn:dep_rule} or \eqref{eqn:see_rules}, each of the conditions $\mvar{condit}_i$ could optionally be preceded by the negation symbol \code{!}.  In general, a rule only applies when the ordinary conditions are true and the negated conditions are false.  The concatenation of column vectors in \cref{eqn:rule_emb,eqn:rule_preupdate} omits $\sem{\mvar{g}_i}$ if $\mvar{condit}_i$ is negated, since then $\mvar{g}_i$ is not a fact and does not have a vector (rather, $\sem{\mvar{g}_i}=\nullval$).

Many dialects of Datalog permit programs with negation.  If we allow cycles (\cref{app:cyclic}), we would impose the usual restriction that negation may not appear on cycles, i.e., programs may use only \defn{stratified negation}.  This restriction ensures that the set of facts is well-defined, by excluding rules like \code{paradox \dep{} !paradox.}

\paragraph{Example.} Extending our example of \cref{sec:lang}, we might say that a person can eventually grow up into an adult and acquire a gender.  Whether person \dvar{X} grows up into (say) a woman, and the time at which this happens, depends on the probability or intensity (\cref{sec:prob}) of the \code{\evt{growup}(\dvar{X},female)} event.  We use negation to say that a \evt{growup} event can happen only once to a person---after that, all \evt{growup} events for that person become false atoms (have probability 0).
\begin{lstlisting}[name=human,firstnumber=auto]v
@\code{\blk{adult}(\dvar{X},\dvar{G}) \see \evt{growup}(\dvar{X},\dvar{G}).}@
@\code{\blk{adult}(\dvar{X}) \dep \blk{adult}(\dvar{X},\dvar{G}).}@
@\code{\evt{growup}(\dvar{X},\dvar{G}) \dep \blk{person}(\dvar{X}), \blk{gender}(\dvar{G}), !\blk{adult}(\dvar{X}).} @
@\code{\blk{gender}(female).}@
@\code{\blk{gender}(male).}@
@\code{\blk{gender}(nonbinary).} \progvdots@
\end{lstlisting}

As a result, an adult has exactly one gender, chosen stochastically.  Female and male adults who know each other can procreate:
\begin{lstlisting}[name=human,firstnumber=auto]
@\code{\evt{procreate}(\dvar{X},\dvar{Y}) \dep \blk{rel}(\dvar{X},\dvar{Y}),\\ \blk{adult}(\dvar{X},female), \blk{adult}(\dvar{Y},male).}@
\end{lstlisting}

\subsection{Highway Connections}\label{app:highway}

As convenient ``syntactic sugar,'' we introduce a variant \dephighway of the \dep connector.  The extra horizontal line introduces extra \defn{highway connections} that skip a level in the neural network.  A fact's embedding can now be directly affected by its grandparents in the proof DAG, not just its parents.  This does not change the set of facts that are proved.

Highway connections of roughly this sort have been argued to help neural network training by providing shorter, more direct paths for backpropagation \cite{srivastava2015highway}.  They also increase the number of parameters in the model.

We use an example to show how they are specified in neural Datalog.  Consider the following \dephighway rules.  The first rule replaces \cref{line:rel} from \cref{sec:lang} with a \dephighway version.  The second rule is added to make the example more interesting.  It uses a high-dimensional \blk{teacher} embedding that represents the academic relationship between \dvar{X} and \dvar{Y} (which is presumably updated by every academic interaction between them).
\begin{lstlisting}[name=human,firstnumber=auto]
@\blk{rel}(\dvar{X},\dvar{Y}) \dephighway \blk{opinion}(\dvar{X},\dvar{U}), \blk{opinion}(\dvar{Y},\dvar{U}). \label{line:relhighwayone}@
@\blk{rel}(\dvar{X},\dvar{Y}) \dephighway \blk{teacher}(\dvar{X},\dvar{Y}). \label{line:relhighwaytwo}@
\end{lstlisting}

The embeddings of $\blk{rel}$ facts are computed as before.  However, the \dephighway rules in the definition of $\blk{rel}$ affect the interpretation of the other \dep, \dephighway, and \see rules in the program whose body contains $\blk{rel}$.  A simple example of such a rule is \cref{line:help} from \cref{sec:eventfacts}:
\begin{lstlisting}[name=human,firstnumber=auto]
@\code{\evt{help}(\dvar{X},\dvar{Y}) \dep \blk{rel}(\dvar{X},\dvar{Y}).} \label{line:help-repeat}@
\end{lstlisting}
The following rules are now \emph{automatically added to the program}:
\begin{lstlisting}[name=human,firstnumber=auto]
@\code{\evt{help}(\dvar{X},\dvar{Y}) \dep \blk{opinion}(\dvar{X},\dvar{U}), \blk{opinion}(\dvar{Y},\dvar{U}).} \label{line:helphighwayone}@ 
@\code{\evt{help}(\dvar{X},\dvar{Y}) \dep \blk{teacher}(\dvar{X},\dvar{Y}).} \label{line:helphighwaytwo}@
\end{lstlisting}

As a result, an embedding such as $\seme{help}{eve,adam}$ is defined using \emph{not only} $\sema{rel}{eve,adam}$, but \emph{also} the embeddings of any lower-level facts that proved \code{\blk{rel}(eve,adam)} via the \dephighway \cref{line:relhighwayone,line:relhighwaytwo}.

In the simple case where \code{\evt{rel}(eve,adam)} has only one proof, this scheme is equivalent to augmenting $\seme{rel}{eve,adam}$ by concatenating it with the embeddings of its parent or parents.  This higher-dimensional version of $\seme{rel}{eve,adam}$ now participates as usual in the computation of other embeddings such as $\seme{help}{eve,adam}$.  However, notice that the dimensionality of the augmented $\seme{rel}{eve,adam}$ will differ according to whether \code{\evt{rel}(eve,adam)} was proved via \cref{line:relhighwayone} or \cref{line:relhighwaytwo}.  Therefore, different parameters must be used for the additional dimensions, associated with \cref{line:helphighwayone} or \cref{line:helphighwaytwo} respectively.

More generally, notice that $\seme{help}{eve,adam}$ will \emph{sum} over the contributions from the two \cref{line:helphighwayone,line:helphighwaytwo} (via \cref{eqn:term_emb} or \cref{eqn:term_emb_cell}).  The former contribution may itself involve \emph{pooling} (via \cref{eqn:rule_emb}) over all topics \dvar{U} about which \code{eve} and \code{adam} both have opinions.  This pooling is performed separately from the pooling over \dvar{U} used in \cref{line:relhighwayone}: in particular, it may use a different $\beta$ parameter.

Of course, the definition of \blk{rel} may also include \emph{non-highway} rules such as
\begin{lstlisting}[name=human,firstnumber=auto]
@\code{\blk{rel}(\dvar{X},\dvar{Y}) \dep \blk{married}(\dvar{X},\dvar{U}).}@
@\code{\blk{rel}(\dvar{X},\dvar{Y}) \see \evt{hire}(\dvar{X},\dvar{Y}).}@
\end{lstlisting}
Since \cref{line:help-repeat} is still in the program, however, proving \code{\blk{rel}(eve,adam)} remains sufficient to prove the possible event \code{\evt{help}(eve,adam)} even when \code{\blk{rel}(eve,adam)} is proved by non-highway rules.

Longer highways can be created by chaining multiple \dephighway rules
together.  For example, if we replace \cref{line:help-repeat} with
a \dephighway version,
\begin{lstlisting}[name=human,firstnumber=auto]
@\code{\evt{help}(\dvar{X},\dvar{Y}) \dephighway \blk{rel}(\dvar{X},\dvar{Y}).} \label{line:help-highway}@
\end{lstlisting}
then \crefrange{line:helphighwayone}{line:helphighwaytwo} will also use \dephighway.  Hence, any rule whose body uses \evt{help} will
automatically acquire versions that mention \blk{rel}, \blk{opinion}, and \blk{teacher} (by repeating the bodies of \crefrange{line:help-repeat}{line:helphighwaytwo} respectively).

There are several subtleties in our highway program transformation:

The additional \crefrange{line:helphighwayone}{line:helphighwaytwo} were constructed by expanding (``inlining'') the call to \code{\blk{rel}} within the body of \cref{line:help-repeat}.  In logic programming, the inlining transformation is known as \defn{unfolding}.  In general it may involve unification, as well as variable renaming to avoid capture.

When we unfold a rule condition, the original condition is usually deleted from the new (unfolded) version of the rule, since it is now redundant.  However, the event that triggers an update rule cannot be deleted in this way.  Consider
the update \cref{line:grateful} from \cref{sec:dltt}:
\begin{lstlisting}[name=human,firstnumber=auto]
@\blk{grateful}(\dvar{Y},\dvar{X}) \see \evt{help}(\dvar{X},\dvar{Y}), \blk{person}(\dvar{Y}). \label{line:grateful-repeat}@
\end{lstlisting}
Suppose \code{\evt{help}(X,Y)} is defined using the highway  \cref{line:help-highway}.  The rule that we automatically add cannot be
\begin{lstlisting}[name=human,firstnumber=auto]
@\blk{grateful}(\dvar{Y},\dvar{X}) \see \blk{rel}(\dvar{X},\dvar{Y}), \blk{person}(\dvar{Y}). \label{line:gratefulhighway-bad}@
\end{lstlisting}
as one might expect, because \code{\blk{rel}(X,Y)} is not even an event that
can be used in this position.  Instead, we must ensure that the event is still triggered by the original event:
\begin{lstlisting}[name=human,firstnumber=auto]
@\blk{grateful}(\dvar{Y},\dvar{X}) \see \evt{help}(\dvar{X},\dvar{Y}) : 0, \blk{rel}(\dvar{X},\dvar{Y}), \blk{person}(\dvar{Y}) : 0. \label{line:gratefulhighway}@
\end{lstlisting}
As explained in \cref{app:param_share} below, the \code{:~0} notation says that although the highway \cref{line:gratefulhighway} is \emph{triggered} by the event \code{\evt{help}(\dvar{X},\dvar{Y})}, it ignores the event's \emph{embedding}.  After all, the event's embedding is still considered by the original \cref{line:grateful-repeat} and does not need to be considered again.  The contributions of these two rules will be summed by \cref{eqn:term_emb} or \cref{eqn:term_emb_cell} before $\tanh$ is applied.

The above example also illustrates the handling of rule conditions that are \emph{not} unfolded, such as \blk{person}.  The unfolded rule (e.g., \cref{line:gratefulhighway}) marks these conditions with \code{:~0} as well, to say that while they are still boolean conditions on the update, their embeddings should also be ignored.  Again, their embeddings are considered in the original \cref{line:grateful-repeat}, so they do not need to be considered again.

Finally, notice that a rule body may contain multiple events and/or conditions that are defined using highway rules.  How do we expand
\begin{lstlisting}[name=multhighway,firstnumber=auto]
@\blk{world} \see \evt{e}, \blk{f}, \blk{g}.@
\end{lstlisting}
given the following highway definitions?
\begin{lstlisting}[name=multhighway,firstnumber=auto]
@\evt{e} \dephighway \blk{e1}.@
@\evt{e} \dephighway \blk{e2}.@
@\blk{g} \dephighway \blk{g1}.@
@\blk{g} \dephighway \blk{g2}.@
\end{lstlisting}
The general answer is that we unfold each of the body elements in parallel, to allow highway connections from that element.  In this case we add 4 new rules:
\begin{lstlisting}[name=multhighway,firstnumber=auto]
@\blk{world} \see \evt{e} :~0, \blk{e1}, \blk{f} :~0, \blk{g} :~0.@
@\blk{world} \see \evt{e} :~0, \blk{e2}, \blk{f} :~0, \blk{g} :~0.@
@\blk{world} \see \evt{e} :~0, \ \ \ \ \blk{f} :~0, \blk{g1}.@
@\blk{world} \see \evt{e} :~0, \ \ \ \ \blk{f} :~0, \blk{g2}.@
\end{lstlisting}

\subsection{Infinite Domains}\label{app:infinite}

\Cref{sec:finite} explained that under our current formalism, any given model only allows a finite set of atoms.  Thus, it is not possible for new persons to be born.  

One way to accommodate that might be to relax Datalog's restriction on nesting.\footnote{To be safe, we should allow \emph{only} the $\see$ rules (which are novel in our formalism) to derive new facts with greater nesting depth than the facts that appear in the body of the rule.  This means that the nesting depth of the database may increase over time, by a finite amount each time an event happens.  If we allowed that in traditional $\dep$ rules, for example \code{\blk{\blk{peano}(s(\dvar{X})) \dep \blk{peano}(\dvar{X})}}, then we could get an infinite set of facts at an \emph{single} time.  But then computation at that time might not terminate, and our $\aggr{\beta}$ operators might have to aggregate over infinite sets (see \cref{sec:finite}).}  This allows us to build up an infinite set of atoms from a finite set of initial entities:
\begin{lstlisting}[name=human,firstnumber=auto]
@\code{\blk{birth}(\dvar{X},\dvar{Y},child(\dvar{X},\dvar{Y})) \see \evt{procreate}(\dvar{X},\dvar{Y}).}@
\end{lstlisting}
Thus, each new person would be named by a tree giving their ancestry, e.g., \code{child(eve,adam)} or \code{child(awan,child(eve,adam))}.  But while this method may be useful in other settings, it unfortunately does not allow \code{eve} and \code{adam} to have \emph{multiple} children.  

Instead, we suggest a different extension, which allows events to create new \emph{anonymous} entities (rather than nested terms):
\begin{lstlisting}[name=human,firstnumber=auto]
@\code{\blk{birth}(\dvar{X},\dvar{Y},*) \see \evt{procreate}(\dvar{X},\dvar{Y}).}@
\end{lstlisting}
The special symbol \code{*} denotes a new entity that is created during the update, in this case representing the child being born.  Thus, the event \code{\evt{procreate}(eve,adam)} will launch the fact \code{\blk{birth}(eve,adam,cain)}, where \code{cain} is some internal name that the system assigns to the new entity.  In the usual way when launching a fact, the cell block $\seecell{\code{\blk{birth}(eve,adam,cain)}}$ is updated from an initial value of $\vec{0}$ by \cref{eqn:lstm_c_update} in a way that depends on $\seme{procreate}{eve,adam}$.  

From the new fact \code{\blk{birth}(eve,adam,cain)}, 
additional rules derive further facts, stating that \code{cain} is a person and has two parents:\footnote{Somewhat awkwardly, under our design, \cref{line:die} is not enough to remove \code{\blk{person}(cain)} from the database, since that fact was established by a \dep rule.  We actually have to write a rule canceling \code{cain}'s birth: \code{!\blk{birth}(\dvar{X},\dvar{Y},\dvar{Z}) \see \evt{die}(\dvar{Z}).}  Notice that this rule will remove not only \code{\blk{person}(cain)} but also \code{\blk{parent}(eve,cain)} and \code{\blk{parent}(adam,cain)}.  Even then, the entity \code{cain} may still be referenced in the database as a \blk{parent} of his own children, until they die as well.}
\begin{lstlisting}[name=human,firstnumber=auto]
@\code{\blk{person}(\dvar{Z}) \dep \evt{birth}(\dvar{X},\dvar{Y},\dvar{Z}).}@
@\code{\blk{parent}(\dvar{X},\dvar{Z}) \dep \evt{birth}(\dvar{X},\dvar{Y},\dvar{Z}).}@
@\code{\blk{parent}(\dvar{Y},\dvar{Z}) \dep \evt{birth}(\dvar{X},\dvar{Y},\dvar{Z}).}@
\end{lstlisting}
Notice that the embedding $\sema{person}{cain}$ initially depends on the state of his parents and their relationship at the time of his procreation.  This is because it depends on $\seme{birth}{eve,adam,cain}$ which depends through its cell block on $\seme{procreate}{eve,adam}$, as noted above.  $\sema{person}{cain}$ may be subsequently updated over time by events such as \code{\evt{help}(eve,cain)}, which affect its cell block.

As another example, here is a description of a sequence of orders in a restaurant: 
\begin{lstlisting}[name=crp,firstnumber=auto]
@\code{\dep \embed(\evt{dish}, 5).}@
@\code{\dep \isevent(\evto{order}, 0).}@
@\code{\evto{order}(\dvar{X}) \dep \blk{dish}(\dvar{X}).}@ @\label{line:olddish}@
@\code{\evto{order}(*)}.@ @\label{line:newdish}@
@\code{\blk{dish}(\dvar{X}) \see \evto{order}(\dvar{X}).}@ @\label{line:dishembed}@
\end{lstlisting}
This program says that the possible orders consist of any existing dish or a new dish.  When used in the discrete-time setting, this model is similar to the Chinese restaurant process (CRP) \citep{aldous_exchangeability_1985}.  Just as in the CRP, 
\begin{itemize}
\item The relative probability of ordering a new dish at time $s \in \Nat$ is a (learned) constant (because \cref{line:newdish} has no conditions).
\item The relative probability of each possible $\evto{order}(\dvar{X})$ event, where $\dvar{X}$ is an existing dish, depends on the embedding of $\blk{dish}(\dvar{X})$ (\cref{line:olddish}).  That embedding reflects only the number of times $\dvar{X}$ has been ordered previously (\cref{line:dishembed}), though its (learned) dependence on that number does not have to be linear as in the CRP.  
\end{itemize}
Interestingly, in the continous-time case---or if we added a rule \code{\blk{dish}(\dvar{X}) \see \exoo{tick}} that causes an update at every discrete time step (see \cref{app:tick} below)---the relative probability of the $\evto{order}(\dvar{X})$ event would also be affected by the time intervals between previous orders of $\dvar{X}$.  It is also easy to modify this program to get variant processes in which the relative probability of \dvar{X} is also affected by previous orders of dishes $\dvar{Y} \neq \dvar{X}$ \citep[cf.][]{blei_correlated_2006} or by the exogenous events at the present time and at times when $\dvar{X}$ was ordered previously \citep[cf.][]{blei2009distance}.

\Cref{app:simult} below discusses how an event may trigger an unbounded number of dependent events that provide details about it.  This could be used in conjunction with the \code{*} feature to create a whole tree of facts that describe a new anonymous entity.

\subsection{Uses of Exogenous Events}\label{app:tick}

The extension to allow exogeneous events was already discussed in the main paper (\cref{sec:exo}).  Here we mention two specific uses in the discrete-time case.

It is useful in the discrete-time case to provide an exogenous \exoo{tick} event at \emph{every} $s \in \Nat$.  (Note that this results in a second event at every time step; see \cref{fn:simultevents}.)  Any cell blocks that are updated by the exogenous \exoo{tick} events will be updated even at time steps $s$ between the modeled events that affect those cell blocks.  For example, one can write a rule such as \code{\blk{person}(\dvar{X}) \see \exoo{tick}, \blk{person}(\dvar{X}), \blk{world}.}\@ so that persons continue to evolve even when nothing is happening to them.  This is similar to the way that in the continous-time case, cell blocks with $\delta\neq 0$ will drift via \cref{eqn:cell_update} during the intervals between the modeled events that affect those cell blocks.\footnote{\label{fn:conttick}In fact, tick events can \emph{also} be used in the continuous case, if desired \cite{mei-17-neuralhawkes}.  Then the drifting cells not only drift, but also undergo periodic learned updates that may depend on other facts (as specified by the \exoo{tick} update rules).}

Another good use of exogenous events in discrete time is to build a conditional probability model such as a word sequence tagger.  At every time step $s$, a word occurs as an exogenous event, at the same time that the model generates an tag event that supplies a tag for the word at the previous time step.  These two events at time $s$ \emph{together} update the state of the model to determine the distribution over the next tag at time $t=s+1$.  Notice that the influences of the word and the tag on the update vector are summed (by the $\sum_r$ in \cref{eqn:cell_update}). This architecture is similar to a left-to-right LSTM tagger \citep[cf.][]{ling-15-finding,tran-etal-2016-unsupervised}.\looseness=-1

\subsection{Modeling Multiple Simultaneous Events}\label{app:simult}

\Cref{sec:prob} explained how to model a discrete-time event sequence:
\begin{quote}
\Paste{discretetimenorm}
\end{quote}
In such a sequence, \emph{exactly} one event is generated at each time $t$.  To change this to ``\emph{at most} one event,'' an additional event type \evto{none} can be used to encode ``nothing occurred.''

Our continuous-time models are also appropriate for data in which \emph{at most} one event occurs at each time $t$, since almost surely, there are no times $t$ with multiple events.  Recall from \cref{sec:prob} that in this setting, the expected number of occurrences of $\mvar{e}$ on the interval $[t,t+dt)$, divided by $dt$, approaches $\lambda_{\mvar{e}}(t)$ as $dt \rightarrow 0^+$.  Thus, given a time $t$ at which one event occurs, the expected total number of \emph{other} events on $[t,t+dt)$ approaches 0 as $dt \rightarrow 0^+$. 

However, there exist datasets in which multiple events do occur at time $t$---even multiple copies of the same event.  By extending our formalism with a notion of \defn{dependent events}, we can model such datasets generatively.  The idea is that an event $\mvar{e}$ at time $t$ can stochastically generate dependent events that also occur at time $t$.

(When multiple events occur at time $t$, our model already specifies
how to handle the \see rule updates that result from these events.
Specifically, multiple events that simultaneously update the same head
are pooled within and across rules by \cref{eqn:cell_update}.)

To model the events that depend on \mvar{e}, we introduce the notion of an \defn{event group}, which represents a group of competing events at a particular instant.  Groups do not persist over time; they appear momentarily in response to particular events.  If event \mvar{e} at time $t$ \defn{triggers} group \mvar{g} and \mvar{g} is non-empty at time $t$, then exactly one event $\mvar{e}'$ in $\mvar{g}$ (perhaps \evto{none}) will stochastically occur at time $t$ as well.

Under some programs, it will be possible for multiple copies---that
is, \defn{tokens}---of the \emph{same} event type to occur at the same time.
For precision, we use $\mvar{e}$ below for a particular event token at
a particular time, using $\bar{\mvar{e}}$ to denote the Datalog atom
that names its event type.  Similarly, we use $\mvar{g}$ for a
particular token of a triggered group, using $\bar{\mvar{g}}$ to
denote the Datalog atom that names the type of group.  We write
$\sem{\mvar{e}}$ and $\sem{\mvar{g}}$ for the token embeddings: this
allows different tokens of the same type to have different embeddings
at time $t$, depending on how they arose.

We allow new program lines of the following forms:\footnote{Mnemonically, note that the ``doubled'' side of the symbol \grpdep or \depgrp is next to the group, since the group usually contains multiple events.  This is also why group names are double-underlined in the examples below.}
\begin{subequations}\label{eqn:grp_rules}
\begin{align}
\MoveEqLeft[4.5]{\code{\dep \iseventgroup(\mvar{functor}, \mvar{dimension})}.} \label{eqn:declgrp} \\
\code{\mvar{group} \grpdep} &\code{ \mvar{event}, \mvar{condit}$_1$, $\ldots$, \mvar{condit}$_N$.} \label{eqn:grpdep}\\
\code{\mvar{event}$\!$ \depgrp} &\code{ \mvar{group}, \mvar{condit}$_1$, $\ldots$, \mvar{condit}$_N$.} \label{eqn:depgrp}
\end{align}
\end{subequations}

An {\iseventgroup} declaration of the form \eqref{eqn:declgrp} is used
to declare that atoms with a particular functor refer to event groups,
similar to an {\isevent} declaration.  We will display such functors
with a double underline.

A rule of the form \eqref{eqn:grpdep} is used to trigger a group of
possible dependent events.  If \mvar{e} is an event token at time $t$,
then it triggers a token \mvar{g} of group type $\bar{\mvar{g}}$ at
time $t$, for each $\bar{\mvar{g}}$ and each rule $r$ having at least
one instantiation of the form \mbox{\code{$\bar{\mvar{g}}$ \grpdep
    $\bar{\mvar{e}}$, \mvar{c}$_1$, $\ldots$, \mvar{c}$_N$}} for which
the \mvar{c}$_i$ are all facts at time $t$.  The embedding of this
group token \mvar{g} pools over all such instantiations of rule $r$ (as in \cref{eqn:rule_emb}):
\begin{align}\label{eqn:grpdep_emb}
\sem{\mvar{g}} &\defeq \aggr{\beta_r}_{\mvar{c}_1, \ldots, \mvar{c}_N} 
\vec{W}_r \underbrace{[1; \sem{\mvar{e}}; \sem{\mvar{c}_1}; \ldots; \sem{\mvar{c}_N}]}_{\clap{\text{\scriptsize concatenation of column vectors}}} \;\in \Real^{D_{\mvar{g}}}
\end{align}
where all embeddings are evaluated at time $t$.

Rules of the form \eqref{eqn:depgrp} are used to specify the possible
events in a group.  Very similarly to the above, if the group \mvar{g}
is triggered at time $t$, then it contains a token $\mvar{e}'$ of
event type $\bar{\mvar{e}}'$, for each $\bar{\mvar{e}}'$ and each rule
$r$ having at least one instantiation of the form
\code{$\bar{\mvar{e}}'$ \depgrp $\bar{\mvar{g}}$, \mvar{c}$_1$,
  $\ldots$, \mvar{c}$_N$} for which the \mvar{c}$_i$ are all facts at
time $t$.  The embedding of this event token $\mvar{e}'$ pools over all such
instantiations of rule $r$:
\begin{align}\label{eqn:depgrp_emb}
\sem{\mvar{e}'} &\defeq \aggr{\beta_r}_{\mvar{c}_1, \ldots, \mvar{c}_N} 
\vec{W}_r \underbrace{[1; \sem{\mvar{g}}; \sem{\mvar{c}_1}; \ldots; \sem{\mvar{c}_N}]}_{\clap{\text{\scriptsize concatenation of column vectors}}} \;\in \Real^{D_{\mvar{g}}}
\end{align}
where all embeddings are evaluated at time $t$.

Since each $\mvar{e}'$ in group $\mvar{g}$ is an event, we compute not only an embedding $\sem{\mvar{e}'}$ but also an unnormalized probability $\lambda_{\mvar{e}'}$, computed just as in \cref{sec:prob} (using $\exp$ rather than $\softplus$).  Exactly one of the finitely many event tokens in $\mvar{g}$ will occur at time $t$, with event type $\mvar{e}'$ being chosen from $\mvar{g}$ with probability proportional to $\lambda_{\mvar{e}'}$.

\paragraph{Training.} In fully supervised training of this model, the dependencies are fully observed.  For each dependent event token $\mvar{e}'$ that occurs at time $t$, the training set specifies what it depends on---that it is a dependent event, which group \mvar{g} it was chosen from, and which rule $r$ established that $\mvar{e}'$ was an element of \mvar{g}.  Furthermore, the training set must specify for \mvar{g} which event \mvar{e} triggered it and via which rule $r$.  However, if these dependencies are not fully observed, then it is still possible to take the training objective to be the incomplete-data likelihood, which involves computing the total probability of the bag of events at each time $t$ by summing over all possible choices of the dependencies.

\paragraph{Marked events.}  To see the applicability of our formalism, consider a marked point process (such as the marked Hawkes process).  This is a traditional type of event sequence model in which each event occurrence also generates a stochastic \defn{mark} from some distribution.  The mark contains details about the event.  For example, each occurrence of \code{\evt{eat\_meal}(eve)} might generate a mark that specifies the food eaten and the location of the meal.

Why are marked point processes used in practice?  An alternative would be to refine the atoms that describe events so that they contain the additional details.  This leads to fine-grained event types such as \code{\evt{eat\_meal}(eve,apple,tree\_of\_knowledge)}.  However, that approach means that computing $\inten{}{t} \defeq \sum_{\mvar{e}\in \set{E}(t)} \inten{\mvar{e}}{t}$ during training (\cref{sec:train}) or sampling (\cref{app:synthetic_details}) involves summing over a large set of fine-grained events, which is computationally expensive.  Using marks makes it possible to generate a coarse-grained event first, modeling its probability without yet considering the different ways to refine it.  The event's details are considered only once the event has been chosen.  This is simply the usual computational efficiency argument for locally normalized generative models.  

Our formalism can treat an event's mark as a dependent event, using
the neural architecture above to model the mark probability
$p(\mvar{e}' \mid \mvar{e})$ as proportional to $\lambda_{\mvar{e}'}$.
The set of possible marks for an event is defined by rules of the form
\eqref{eqn:grp_rules} and may vary by event type and vary by time.

\paragraph{Multiply marked events.}  Our approach also makes it easy for an event to independently generate multiple marks, which describe different attributes of an event.  For example, each meal at time $t$ may select a dependent location,
\begin{lstlisting}[name=restaurant,firstnumber=auto]
@\code{\dep \iseventgroup(\grp{restaurants}, 5).}@ 
@\code{\dep \isevent(\evto{eat\_at}, 0).}@
@\code{\grp{restaurants} \grpdep \evt{eat\_meal}(\dvar{X}).}@
@\code{\evto{eat\_at}(\dvar{Y}) \depgrp \grp{restaurants}, \blk{is\_restaurant}(\dvar{Y}).}@
@\code{\evto{eat\_at}(home) \depgrp \grp{restaurants}.}@
\end{lstlisting}
which associates some dependent restaurant \dvar{Y} (or \code{home}) with the meal.\footnote{Notice that the choice of event $\evto{eat\_at}(\dvar{Y})$ depends on the person \dvar{X} who is eating the meal, through the embedding of this token of $\sem{\grp{restaurants}}$, which depends on $\sem{\evt{eat\_meal(\dvar{X})}}$.}
At the same time, the meal may select a \emph{set} of foods to eat, where each food \dvar{U}\footnote{Notice that the unnormalized probability of including \dvar{U} in \dvar{X}'s meal depends on \dvar{X}'s opinion of \dvar{U}.} is in competition with \evto{none}\footnote{The annotation \code{\usec \prm{0}} in the last line (explained in \cref{app:param_share} below) is included as a matter of good practice.  In keeping with the usual practice in binary logistic regression, it simplifies the computation of the normalized probabilities, without loss of generality, by ensuring that the unnormalized probability of \evto{none} is constant rather than depending on \dvar{U}.}
to indicate that it may or may not be chosen:
\begin{lstlisting}[name=restaurant,firstnumber=auto]
@\code{\dep \iseventgroup(\grp{optdish}, 7).}@
@\code{\dep \isevent(\evto{eat\_dish}, 0).}@
@\code{\dep \isevent(\evto{none}, 0).}@
@\code{\grp{optdish}(\dvar{U}) \grpdep \evt{eat\_meal}(\dvar{X}), \\ \blk{food}(\dvar{U}), \blk{opinion}(\dvar{X},\dvar{U}).}@
@\code{\evto{eat\_dish}(\dvar{U}) \depgrp \grp{optdish}(\dvar{U}).}@
@\code{\evto{none} \depgrp \grp{optdish}(\dvar{U}) \usec \prm{0}.}@
\end{lstlisting}

\paragraph{Recursive marks.} Dependent events can recursively trigger dependent events of their own, leading to a tree of event tokens at time $t$.  This makes it possible to model the top-down generation of tree-structured metadata, such as a syntactically well-formed sentence that describes the event \cite{zhang-etal-2016-top}.  Observing such sentences in training data would then provide evidence of the underlying embeddings of the events.  For example, to generate derivation trees from a context-free grammar, encode each nonterminal symbol as an event group, whose events are the production rules that can expand that nonterminal. In general, the probability of a production rule depends on the sequence of production rules at its ancestors, as determined by a recurrent neural net.

A special case of a tree is a sequence: in the meal example, each dish could be made to generate the next dish until the sequence terminates by generating \evto{none}.  The resulting architecture precisely mimics the architecture of an RNN language model \cite{mikolov-10-rnnlm}.

\paragraph{Multiple agents.}
A final application of our model is in a discrete-time setting where there are multiple agents, which naturally leads to multiple simultaneous events.  For example, at each time step $t$, every person stochastically chooses an action to perform (possibly \evto{none}).  This can be accomplished by allowing the \exoo{tick} event (\cref{app:tick}) to trigger one group for each person:
\begin{lstlisting}[name=discretehuman,firstnumber=auto]
@\code{\dep \iseventgroup(\grp{actions}, 7).}@
@\code{\grp{actions(X)} \grpdep \evt{tick}, \grp{person}(\dvar{X}).}@
@\code{\evt{help}(\dvar{X},\dvar{Y}) \depgrp \grp{actions}(\dvar{X}), \blk{rel}(\dvar{X},\dvar{Y}).}\progvdots@
\end{lstlisting}
This is a group-wise version of \cref{line:help} in the main paper.  

A similar structure can be used to produce a ``node classification'' model in which each node in a graph stochastically generates a label at each time step, based on the node's current embedding \cite{hamilton-17-representation,xu-20-inductive}.  The event group for a node contains its possible labels.  The graph structure may change over time thanks to exogeneous or endogenous events.  

\paragraph{Example.} For concreteness, below is a fully generative model of a dynamic colored directed graph, using several of the extensions described in this appendix.  The model can be used in either a discrete-time or continuous-time setting.  

The graph's nodes and edges have embeddings, as do the legal colors for nodes:
\begin{lstlisting}[name=graph,firstnumber=auto]
@\code{\dep \embed(\blk{node}, 8).}@
@\code{\dep \embed(\blk{edge}, 4).}@
@\code{\dep \embed(\blk{color}, 3).}@
\end{lstlisting}

In this version, edges are stochastically added and removed over time, one at a time.  Any two unconnected nodes determine through their embeddings the probability of adding an edge between them, as well as the initial embedding of this edge.  The edge's embedding may drift over time,\footnote{In the continuous-time setting, the drift is learned.  In the discrete-time setting, we must explicitly specify drift as explained in \cref{app:tick}, via a rule such as \code{\blk{edge}(\dvar{U},\dvar{V}) <- \exoo{tick}}.} and at any time determines the edge's probability of deletion.
\begin{lstlisting}[name=graph,firstnumber=auto]
@\code{\dep \isevent(\evt{add\_edge}, 8).}@
@\code{\dep \isevent(\evto{del\_edge}, 0).}@
@\code{\evt{add\_edge}(\dvar{U},\dvar{V}) \dep \blk{node}(\dvar{U}), \blk{node}(\dvar{V}), !\blk{edge}(\dvar{U},\dvar{V}).}@
@\code{\evto{del\_edge}(\dvar{U},\dvar{V}) \dep \blk{edge}(\dvar{U},\dvar{V}).}@
@\code{\blk{edge}(\dvar{U},\dvar{V}) \see \evt{add\_edge}(\dvar{U},\dvar{V}).}@
@\code{!\blk{edge}(\dvar{U},\dvar{V}) \see \evto{del\_edge}(\dvar{U},\dvar{V}).}@
\end{lstlisting}

Adding \code{\blk{edge}(\dvar{U},\dvar{V})} to the graph causes two dependent events that simultaneously and stochastically relabel both \dvar{U} and \dvar{V} with new colors.  This requires triggering two event groups (unless \dvar{U}=\dvar{V}).  A node's new color \dvar{C} depends stochastically  on the embeddings of the node and its neighbors, as well as the embeddings of the colors:
\begin{lstlisting}[name=graph,firstnumber=auto]
@\code{\dep \iseventgroup(\grp{labels}, 8).}@
@\code{\dep \isevent(\evt{label}, 8).}@
@\code{\grp{labels}(\dvar{U}) \grpdep \evt{add\_edge}(\dvar{U},\dvar{V}).}@
@\code{\grp{labels}(\dvar{V}) \grpdep \evt{add\_edge}(\dvar{U},\dvar{V}).}@
@\code{\evt{label}(\dvar{X},\dvar{C}) \depgrp \grp{labels}(\dvar{X}), \blk{color}(\dvar{C}), \blk{node}(\dvar{X}), \blk{edge}(\dvar{X},\dvar{Y}), \blk{node}(\dvar{Y}).}@ 
\end{lstlisting}

Finally, here is how a relabeling event does its work.  The \boo{has\_color} atoms that are updated here are simply facts that record the current coloring, with no embedding.  However, the rules below ensure that a \emph{node}'s embedding records its \emph{history} of colors (and that it has only one color at a time):
\begin{lstlisting}[name=graph,firstnumber=auto]
@\code{!\boo{has\_color}(\dvar{U},\dvar{D}) \see \evt{label}(\dvar{U},\dvar{C}), \blk{color}(\dvar{D}).}@
@\code{\boo{has\_color}(\dvar{U},\dvar{C}) \see \evt{label}(\dvar{U},\dvar{C}).}@
@\code{\blk{node}(\dvar{U}) \see \boo{has\_color}(\dvar{U},\dvar{C}), \blk{color}(\dvar{C}).}@ 
\end{lstlisting}

The initial graph at time $t=0$ can be written down by enumeration:
\begin{lstlisting}[name=graph,firstnumber=auto]
@\code{\blk{color}(red).}@
@\code{\blk{color}(green).}@
@\code{\blk{color}(blue).}@
@\code{\boo{has\_color}(0,red).}@
@\code{\boo{has\_color}(1,blue)}@
@\code{\boo{has\_color}(2,red).}@
@\code{\blk{node}(\dvar{U}) \dep \boo{has\_color}(\dvar{U},\dvar{C}).}@
@\code{\blk{edge}(0,1) \see \exoo{init}.}@
\end{lstlisting}

\paragraph{Inheritance.} As a convenience, we allow an event group to be used anywhere that an event can be used---at the start of the body of a rule of type \eqref{eqn:see_launch}, \eqref{eqn:see_dock}, or \eqref{eqn:grpdep}.  Such a rule applies at times when the group is triggered (just as a rule that mentions an event, instead of a group, would apply at times when that event occurred).  

This provides a kind of inheritance mechanism for events:
\begin{lstlisting}[name=human,firstnumber=auto]
@\code{\dep \iseventgroup(\grp{act}, 5).}@
@\code{\grp{act}(\dvar{X}) \grpdep \evt{sleep}(\dvar{X}).}@
@\code{\grp{act}(\dvar{X}) \grpdep \evt{help}(\dvar{X},\dvar{Y}), \blk{person}(\dvar{Y}).}\progvdots@
@\code{\blk{person(\dvar{Y})} \see \grp{act}(\dvar{X}), \blk{parent}(\dvar{X},\dvar{Y}), \blk{person(\dvar{Y})}.}@ @\label{line:act_start}@
@\code{\blk{animal(\dvar{Y})} \see \grp{act}(\dvar{X}), \blk{own}(\dvar{X},\dvar{Y}), \blk{animal(\dvar{Y})}.}@ @\label{line:act_end}@
\end{lstlisting}
This means that whenever \dvar{X} takes any action---\evt{sleep}, \evt{help}, etc.---\crefrange{line:act_start}{line:act_end} will update the embeddings of \dvar{X}'s children and pets.  

Adopting the terminology of object-oriented programming, \code{\grp{act}(eve)} functions as a class of events (i.e., event type), whose subclasses include \code{\evt{help}(eve,adam)} and many others.  In this view, each particular instance (i.e., event token) of the subclass \code{\evt{help}(eve,adam)} has a method that returns its embedding in $\mathbb{R}^{D_{\code{help}}}$.  But \crefrange{line:act_start}{line:act_end} instead view this \code{\evt{help}(eve,adam)} event as an instance of the superclass \code{\grp{act}(eve)}, and hence call a method of that superclass to obtain the embedding of the group token \code{\grp{act}(eve)} in $\mathbb{R}^{D_{\code{act}}} = \mathbb{R}^5$, as defined via \cref{eqn:grpdep_emb}.

In the above example, the event group is actually empty, as there are no rules of type \eqref{eqn:depgrp} that populate it with dependent events.  Thus, no  dependent events occur as a result of the group being triggered.  The empty event group is simply used as a class.  One could, however, add rules such as
\begin{lstlisting}[name=human,firstnumber=auto]
@\code{\evt{act\_at}(\dvar{L}) \depgrp \grp{act}(\dvar{X}), \blk{location}(\dvar{L})}.@
\end{lstlisting}
which marks each action (of any type) with a location.

\section{Parameter Sharing Details}\label{app:param_share}

Throughout \cref{sec:formula}, the parameters $\vec{W}$ and $\beta$ are indexed by the rule number $r$.  (They appear in \cref{eqn:rule_emb,eqn:rule_preupdate}.)  Thus, the number of parameters grows with the number of rules in our formalism. 
However, we also allow further flexibility to \emph{name} these parameters with atoms, so that they can be shared among and within rules. 

This is achieved by explicitly naming the parameters to be used by a rule:
\begin{align*}
\code{\mvar{head} \usec \mvar{beta}}&\code{ \dep}\\
&\code{ \usec \mvar{bias\_vector}} \\
&\code{ \mvar{condit}$_1$ \usec \mvar{matrix}$_1$, } \\
&\code{ \qquad\vdots } \\
&\code{ \mvar{condit}$_N$ \usec \mvar{matrix}$_N$. }
\end{align*}
Now $\beta_r$ in \cref{eqn:rule_emb} is replaced by a scalar parameter named by the atom \mvar{beta}.
Similarly, the affine transformation matrix $\vec{W}_r$ in \cref{eqn:rule_emb} is replaced by a parameter matrix that is constructed by \emph{horizontally} concatenating the column vector and matrices named by the atoms $\mvar{bias\_vector}$, $\mvar{matrix}_1, \ldots, \mvar{matrix}_N$ respectively.

To be precise, \mvar{matrix}$_i$ will have $D_{\mvar{head}}$ rows and $D_{\mvar{condit}_i}$ columns.  The computation \eqref{eqn:rule_emb} can be viewed as multiplying this matrix by the vector embedding of the atom that instiatiates \mvar{condit}$_i$, yielding a vector in $\Real^{D_{\mvar{head}}}$.  It then sums these vectors for $i=1,\ldots,N$ as well as the bias vector (also in $\Real^{D_{\mvar{head}}}$), obtaining a vector in $\Real^{D_{\mvar{head}}}$ that it provides to the pooling operator.

These parameter annotations with the \code{\usec} symbol are optional (and were not used in the main paper).  If any of them is not specified, it is set automatically to be rule- and position-specific: in the \rth rule, \mvar{beta} defaults to \code{params($r$,beta)}, \mvar{bias\_vector} defaults to \code{params($r$,bias)}, and \mvar{matrix}$_i$ defaults to \code{params($r$,$i$)}.

As shorthand, we also allow the form 
\begin{align*}
\code{\mvar{head} \usec \mvar{beta}}&\code{ \dep}\\
&\code{ \mvar{condit}$_1$, \mvar{condit}$_N$ \useb \mvar{full\_matrix}. } 
\end{align*}
where \mvar{full\_matrix} directly names the concatenation of matrices that replaces $\vec{W}_r$.  

The parameter-naming mechanism lets us share parameters across rules by reusing their names.
For example, blessings and curses might be inherited using the same parameters:
\begin{lstlisting}[name=human,firstnumber=auto]
@\code{\blk{cursed}(\dvar{Y}) \dep \blk{cursed}(\dvar{X}),$\,$\blk{parent}(\dvar{X},\dvar{Y}) \useb \prm{inherit}.}@ @\label{line:curse_rec_pname}@
@\code{\blk{blessed}(\dvar{Y}) \!\dep\!\! \blk{blessed}(\dvar{X}),$\,$\blk{parent}(\dvar{X},\dvar{Y}) \!\useb\!\makebox[4ex][l]{\prm{inherit}.}}@ @\label{line:bless_rec_pname}@
\end{lstlisting}

Conversely, to do \emph{less} sharing of parameters, the parameter names may mention variables that appear in the head or body of the rule.  In this case, different instantiations of the rule may invoke different parameters.  (\mvar{beta} is only allowed to contain variables that appear in the head, because each way of instantiating the head needs a single $\beta$ to aggregate over all the compatible instantations of its body.)

For example, we can modify \cref{line:curse_rec_pname,line:bless_rec_pname} into
\begin{lstlisting}[name=human,firstnumber=auto]
@\code{\blk{cursed}(\dvar{Y}) \usec \prm{descendant(Y)} \dep \\\blk{cursed}(\dvar{X}), \blk{parent}(\dvar{X},\dvar{Y}) \useb \prm{inherit(X,Y)}.}@
@\code{\blk{blessed}(\dvar{Y}) \usec \prm{descendant(Y)} \dep \\\blk{blessed}(\dvar{X}), \blk{parent}(\dvar{X},\dvar{Y}) \useb \prm{inherit(X,Y)}.}@
\end{lstlisting}
Now each \dvar{X}, \dvar{Y} pair has its own $\vec{W}$ matrix (shared by curses and blessings), and similarly, each \dvar{Y} has its own $\beta$ scalar. 
This example has too many parameters to be practical, but serves to illustrate the point.

If \dvar{X} or \dvar{Y} is an entity created by the \code{*} mechanism (\cref{app:infinite}), then the name will be constructed using a literal \code{*}, so that all newly created entities use the same parameters.  This ensures that the number of parameters is finite even if the number of entities is unbounded.  As a result, parameters can be trained by maximum likelihood and reused every time a sequence is sampled, even though different sequences may have different numbers of entities.  Although novel entities share parameters, facts that differ only in their novel entities may nonetheless come to have different embeddings if they are created or updated in different circumstances.

The special parameter name \prm{0} says to use a zero matrix:
\begin{lstlisting}[name=human,firstnumber=auto]
@\code{\blk{cursed}(\dvar{Y}) \usec \prm{descendant} \dep \\\usec \prm{inherit\_bias}, \\\blk{cursed}(\dvar{X}) \usec \prm{inherit}, \\\blk{parent}(\dvar{X},\dvar{Y}) \usec \prm{0}.}@ 
\end{lstlisting}
In this example, the condition \code{\blk{parent}(\dvar{X},\dvar{Y})} must still be non-\code{\nullval} for the rule to apply, but we ignore its embedding. 

The same mechanism can be used to name the parameters of {\see} rules.  In this case, \mvar{event} at the start of the body can also be annotated, as \code{\mvar{event} \usec \mvar{matrix}$_0$}.  The horizontal concatenation of named matrices now includes the matrix named by \mbox{\mvar{matrix}$_0$}, and is used to replace $\vec{W}_r$ in \cref{eqn:rule_preupdate}.  

For a {\see} rule, it might sometimes be desirable to allow finer-grained control over how the rule affects the drift of a cell block over time (see \cref{eqn:driftblock} in \cref{app:continuous_cells} below).  For example, forcing $\vec{\underline{f}} = \vec{1}$ and $\vec{\underline{i}} = \vec{0}$ in \cref{eqn:fizfizd} ensures via \cref{eqn:see_update_initial} that when the rule updates $\mvar{h}$, it will not introduce a discontinuity in the $\seecell{\mvar{h}}(t)$ function, although it might change the function's asymptotic value and decay rate.  (This might be useful for the \exoo{tick} rules mentioned in \cref{fn:conttick}, for example.)  Similarly, forcing $\vec{\bar{f}} = \vec{1}$ and $\vec{\bar{i}} = \vec{0}$ in \cref{eqn:fizfizd} ensures via \cref{eqn:see_update_asymp} that the rule does not change the asymptotic value of the $\seecell{\mvar{h}}(t)$ function.  These effects can be accomplished by declaring that certain values are $\pm\infty$ in the first column of $\vec{W}_r$ in \cref{eqn:rule_preupdate} (as this column holds bias terms).  We have not yet designed a syntax for such declarations.

We can also name the softplus temporal scale parameter $\timescale$ in \cref{sec:event_math}. 
For example, we can rewrite \cref{line:help_event} of \cref{sec:ndtt} as
\begin{lstlisting}[name=human,firstnumber=auto]
@\code{\dep \isevent(\evt{help}, 8) \usec \prm{intervene}.}@
\end{lstlisting}
and allow \code{\evt{harm}} to share $\timescale$ with \code{\evt{help}}: 
\begin{lstlisting}[name=human,firstnumber=auto]
@\code{\dep \isevent(\evt{harm}, 8) \usec \prm{intervene}.}@
\end{lstlisting}

\section{Updating Drift Functions in the Continuous-Time LSTM}\label{app:continuous_cells}

Here we give the details regarding continuous-time LSTMs, which were omitted from \cref{sec:ndtt_math} due to space limitations.  We follow the design of \citet{mei-17-neuralhawkes}, in which each cell changes endogenously between updates, or ``drifts,'' according to an exponential decay curve:
\begin{align}\label{eqn:drift}
c(t) \;&\defeq\; \bar{c} + (\underline{c} - \bar{c}) \exp(-\delta(t-s)) \text{\ \ \ where $t > s$}
\end{align}
This curve is parameterized by $(s,\underline{c},\bar{c},\delta)$, where
\begin{itemize}
\item $s$ is a starting time---specifically, the time when the parameters were last updated
\item $\underline{c}$ is the starting cell value, i.e., $c(s) = \underline{c}$
\item $\bar{c}$ is the asymptotic cell value, i.e., $\lim_{t\to\infty} c(t) = \bar{c}$
\item $\delta > 0$ is the rate of decay toward the asymptote; notice that the derivative $c'(t) = \delta \cdot (\bar{c} - \underline{c})$
\end{itemize}

In the present paper, we similarly need to define the trajectory through $\Real^{D_{\mvar{h}}}$ of the cell block $\seecell{\mvar{h}}$  associated with fact $\mvar{h}$.  That is, we need to be able to compute $\seecell{\mvar{h}}(t) \in \Real^{D_{\mvar{h}}}$ for any $t$.    Since $\seecell{\mvar{h}}$ is not a single cell but rather a block of $D_{\mvar{h}}$ cells, it actually needs to store not 4 parameters as above, but rather $1 + 3 D_{\mvar{h}}$ parameters.  Specifically, it stores $s \in \Real$, which is the time that the block's parameters were last updated: this is shared by all cells in the block.  It also stores vectors that we refer to as $\seecell{\mvar{h}}^{\vec{\underline{c}}}, \seecell{\mvar{h}}^{\vec{\bar{c}}}, \seecell{\mvar{h}}^{\vec{\delta}} \in \Real^{D_{\mvar{h}}}$.  Now analogously to \cref{eqn:drift}, we define the trajectory of the cell block elementwise:
\begin{align}\label{eqn:driftblock}
\seecell{\mvar{h}}(t) \;&\defeq\; \seecell{\mvar{h}}^{\vec{\bar{c}}} + (\seecell{\mvar{h}}^{\vec{\underline{c}}} - \seecell{\mvar{h}}^{\vec{\bar{c}}}) \exp(-\seecell{\mvar{h}}^{\vec{\delta}}\cdot(t-s)),
\end{align}
for all $t > s$ (up to and including the time of the next event that results in updating the block's parameters).

We now describe exactly \emph{how} the block's parameters are updated when an event occurs at time $s$.  Recall that for the discrete-time case, for each $(r,m)$, we obtained $\seeval{\mvar{h}}{rm} \in \Real^{3D_{\mvar{h}}}$ by evaluating  \eqref{eqn:rule_preupdate} at time $s$.  We then set $(\vec{f};\, \vec{i};\, \vec{z}) \defeq \sigma(\seeval{\mvar{h}}{rm})$.  In the continuous-time case, we evaluate \eqref{eqn:rule_preupdate} at time $s$ to obtain $\seeval{\mvar{h}}{rm} \in \Real^{7D_{\mvar{h}}}$ (so $\vec{W}_r$ needs to have more rows), and accordingly obtain 7 vectors in $(0,1)^{D_{\mvar{h}}}$,
\begin{align}\label{eqn:fizfizd}
(\vec{\underline{f}};\, \vec{\underline{i}};\, \vec{\underline{z}};\, \vec{\bar{f}};\, \vec{\bar{i}};\, \vec{\bar{z}};\, \vec{d}) \;&\defeq\; \sigma(\seeval{\mvar{h}}{rm})  
\end{align}
which we use similarly to \cref{eqn:lstm_c_increment}
to define update vectors for the current cell values (time $s$) and the asymptotic cell values (time $\infty$), respectively
\begin{alignat}{2}
\seeupd[\vec{\underline{c}}]{\mvar{h}}{rm} \;&\defeq\;  
   (\vec{\underline{f}}-1) \cdot \seecell{\mvar{h}}(s) &\mbox{}+ \vec{\underline{i}}\cdot (2\vec{\underline{z}}-1) \label{eqn:see_update_initial}\\
\seeupd[\vec{\bar{c}}]{\mvar{h}}{rm} \;&\defeq\; 
  (\vec{\bar{f}}-1) \cdot \seecell{\mvar{h}}^{\vec{\bar{c}}} &\mbox{}+ \vec{\bar{i}}\cdot (2\vec{\bar{z}}-1) \label{eqn:see_update_asymp}
\intertext{as well as a vector of proposed decay rates:\footnotemark}
\seedelta{\mvar{h}}{rm} \;&\defeq\; \mathrlap{\softplus_1(\inv{\sigma}(\vec{d})) \;\;\in \Real_{>0}^{D_{\mvar{h}}}} \label{eqn:see_update_decay}
\end{alignat}
\footnotetext{\Cref{eqn:see_update_decay} simply replaces the $\sigma$ that produced $\vec{d}$ with $\softplus_1$ (defined in \cref{sec:prob}), since there is no reason to force decay rates into $(0,1)$.}%
We then pool the update vectors from different $(r,m)$ and apply this pooled update, much as we did for the discrete-time cell values in \crefrange{eqn:cell_update}{eqn:lstm_c_increment}:
\begin{alignat}{2}
\seecell{\mvar{h}}^{\vec{\underline{c}}} \;&\defeq\; \seecell{\mvar{h}}(s) &\mbox{}+ \sum_r \aggr{\beta_r}_m \seeupd[\vec{\underline{c}}]{\mvar{h}}{rm} \label{eqn:pooled_update_initial} \\
\seecell{\mvar{h}}^{\vec{\bar{c}}} \;&\defeq\; \seecell{\mvar{h}}^{\vec{\bar{c}}} &\mbox{}+ \sum_r \aggr{\beta_r}_m \seeupd[\vec{\bar{c}}]{\mvar{h}}{rm} \label{eqn:pooled_update_asymp}
\end{alignat}
The special cases mentioned just below the update \eqref{eqn:cell_update} are also followed for the updates \eqref{eqn:pooled_update_initial}--\eqref{eqn:pooled_update_asymp}.

The final task is to pool the decay rates to obtain $\seecell{\mvar{h}}^{\vec{\delta}}$.  It is less obvious how to do this in a natural way.  Our basic idea is that for the $i$\th cell, we should obtain the decay rate $(\seecell{\mvar{h}}^{\vec{\delta}})_i$ by a weighted harmonic mean of the decay rates $(\seedelta{\mvar{h}}{rm})_i$ that were proposed by different $(r,m)$ pairs.  A given $(r,m)$ pair should get a high weight in this harmonic mean to the extent that it contributed large updates $(\seeupd[\vec{\underline{c}}]{\mvar{h}}{rm})_i$ or $(\seeupd[\vec{\bar{c}}]{\mvar{h}}{rm})_i$.

Why harmonic mean?  Observe that the exponential decay curve \eqref{eqn:drift} has a \defn{half-life} of $\frac{\ln 2}{\delta}$.  In other words, at any moment $t$, it will take time $\frac{\ln 2}{\delta}$ for the curve to travel halfway from its current value $c(t)$ to $\bar{c}$.  (This amount of time is independent of $t$.)  Thus, saying that the decay rate is a weighted harmonic mean of proposed decay rates is equivalent to saying that the half-life is a weighted arithmetic mean of proposed half-lives,\footnote{It is also equivalent to saying that the (\nicefrac{2}{3})-life is a weighted arithmetic mean of proposed (\nicefrac{2}{3})-lives, since \cref{eqn:drift} has a (\nicefrac{2}{3})-life of $\frac{\ln 3}{\delta}$.  In other words, there is nothing special about the fraction $\nicefrac{1}{2}$.  \emph{Any} choice of fraction would motivate using the harmonic mean.}
which seems like a reasonable pooling principle.

Thus, operating in parallel over all cells $i$ by performing the following vector operations elementwise, we choose
\begin{align}\label{eqn:harmonic_pool_delta}
\seecell{\mvar{h}}^{\vec{\delta}} \;&\defeq\; \left( \frac{\sum_r \sum_m \vec{w}_{rm} \cdot (\seedelta{\mvar{h}}{rm})^{-1}}{\sum_r \sum_m \vec{w}_{rm} \phantom{\mbox{}\cdot (\seedelta{\mvar{h}}{rm})^{-1}}}\right)^{-1}
\end{align}
We define the vector of unnormalized non-negative weights $\vec{w}_{rm}$ from the updated $\seecell{\mvar{h}}^{\vec{\underline{c}}}$ and $\seecell{\mvar{h}}^{\vec{\bar{c}}}$ values by
\begin{align}\label{eqn:decay_weights}
\vec{w}_{rm} \;&\defeq\;       \left( \aggr{\beta_r}_{m'} \left| \seeupd[\vec{\underline{c}}]{\mvar{h}}{rm'} \right| \right) 
                  \cdot \frac{ \left| \seeupd[\vec{\underline{c}}]{\mvar{h}}{rm} \right|^{\beta_r} }{\sum_{m'} \left| \seeupd[\vec{\underline{c}}]{\mvar{h}}{rm'} \right|^{\beta_r} } \nonumber \\[3pt]
            & \quad + \left( \aggr{\beta_r}_{m'} \left| \seeupd[\vec{\bar{c}}]{\mvar{h}}{rm'} \right| \right) 
                  \cdot \frac{ \left| \seeupd[\vec{\bar{c}}]{\mvar{h}}{rm} \right|^{\beta_r} }{\sum_{m'} \left| \seeupd[\vec{\bar{c}}]{\mvar{h}}{rm'} \right|^{\beta_r} } \nonumber \\[3pt]
            & \quad + \left| \seecell{\mvar{h}}^{\vec{\bar{c}}} - \seecell{\mvar{h}}^{\vec{\underline{c}}} \right| 
\end{align}

The following remarks should be read elementwise, i.e., consider a particular cell $i$, and read each vector $\vec{x}$ as referring to the scalar $(\vec{x})_i$. 

The weights defined in \cref{eqn:decay_weights} are valid weights to use for the weighted harmonic mean \eqref{eqn:harmonic_pool_delta}:
\begin{itemize}
\item $\vec{w}_{rm} \geq 0$, because of the use of absolute value.
\item $\vec{w}_{rm} > 0$ strictly unless $\seecell{\mvar{h}}^{\vec{\bar{c}}} = \seecell{\mvar{h}}^{\vec{\underline{c}}}$.  Thus, the decay rate $\seecell{\mvar{h}}^{\vec{\delta}}$ as defined by \cref{eqn:harmonic_pool_delta} can only be undefined (that is, $\frac{0}{0}$) if $\seecell{\mvar{h}}^{\vec{\bar{c}}} = \seecell{\mvar{h}}^{\vec{\underline{c}}}$, in which case that decay rate is irrelevant anyway.
\end{itemize}

The way to understand the first line of \cref{eqn:decay_weights} is as a heuristic assessment of how much the cell's curve \eqref{eqn:drift} was affected by $(r,m)$ via $\seeupd[\vec{\underline{c}}]{\mvar{h}}{rm}$'s effect on $\seecell{\mvar{h}}^{\vec{\underline{c}}}$.  First of all, $\left( \aggr{\beta_r}_{m'}\limits \left| \seeupd[\vec{\underline{c}}]{\mvar{h}}{rm'} \right|\right)$ is the pooled magnitude of \emph{all} of the \rth rule's attempts to affect $\seecell{\mvar{h}}^{\vec{\underline{c}}}$.  Using the absolute value ensures that even if large-magnitude attempts of opposing sign canceled each other out in \cref{eqn:pooled_update_initial}, they are still counted here as large attempts, and thus give the \rth rule a stronger total voice in determining the decay rate $\seecell{\mvar{h}}^{\vec{\delta}}$.  This pooled magnitude for the \rth rule is then partitioned among the attempts $(r,m)$.  In particular, the fraction in the first line denotes the portion of the \rth rule's pooled effect on $\seecell{\mvar{h}}^{\vec{\underline{c}}}$ that should be heuristically attributed to $(r,m)$ specifically, given the way that \cref{eqn:pooled_update_initial} pooled over all $m$ (recall that this invokes \cref{eqn:squash}).

Thus, the first line of \cref{eqn:decay_weights} considers the effect of $(r,m)$ on $\vec{\underline{c}}$.  The second line adds its effect on $\vec{\bar{c}}$.  The third line effectively acts as smoothing so that we do not pay undue attention to the size ratio among different updates if these updates are tiny.
  In particular, if all of the updates $\seeupd[\vec{\underline{c}}]{\mvar{h}}{rm}$ and $\seeupd[\vec{\bar{c}}]{\mvar{h}}{rm}$ are small compared to the total height of the curve, namely $\left| \seecell{\mvar{h}}^{\vec{\bar{c}}} - \seecell{\mvar{h}}^{\vec{\underline{c}}} \right|$, then the third line will dominate the definition of the weights $\vec{w}_{rm}$, making them close to uniform.  The third line is also what prevents inappropriate division by 0 (see the second bullet point above).

\section{Likelihood Computation Details}\label{app:loglik_details}

In this section we discuss the log-likelihood formulas in \cref{sec:train}.

For the discrete-time setting, the formula simply follows from the fact that the log-probability of event $\mvar{e}$ at time $t$ was defined to be $\log \left( \inten{\mvar{e}}{t} / \inten{}{t} \right)$.

The log-likelihood formula \eqref{eqn:loglikcont} for the continuous-time case has been derived and discussed in previous work \citep{hawkes-71,liniger-09-hawkes,mei-17-neuralhawkes}.
Intuitively, during parameter training, each $\log \inten{\mvar{e}_i}{t_i}$ is increased to explain why $\code{\mvar{e}}_i$ happened at time $t_i$ while $\int_{t=0}^{\dur} \inten{}{t} \dt$ is decreased to explain why no event of any possible type $\mvar{e} \in \set{E}(t)$ ever happened at other times. 
Note that there is no $\log$ under the integral in \cref{eqn:loglikcont}, in contrast to the discrete-time setting.

As discussed in \cref{sec:event}, the integral term in \cref{eqn:loglikcont} is computed using the Monte Carlo approximation detailed by Algorithm 1 of \citet{mei-17-neuralhawkes}, which samples times $t$.

However, at each sampled time $t$, that method still requires a summation over all events to obtain $\inten{}{t}$.  This summation can be expensive when there are many event types.  Thus, we estimate the sum using a simple downsampling trick, as follows.  At any time $t$ that is sampled to compute the integral, let $\set{E}(t)$ be the set of \emph{possible} event types under the database at time $t$. We construct a bag $\set{E}'(t)$ by uniformly sampling event types from $\set{E}(t)$ with replacement, and estimate
\begin{align*}
	\inten{}{t} \approx \frac{|\set{E}|}{|\set{E}'|} \sum_{\mvar{e} \in \set{E}'} \inten{\mvar{e}}{t}
\end{align*}
This estimator is unbiased yet remains much less expensive to compute especially when $|\set{E}'| \ll |\set{E}|$. 
In our experiments, we took $|\set{E}'|=10$ and still found empirically that the variance of the log-likelihood estimate (computed by running multiple times) was rather small.

Another computational expense stems from the fact that we have to make Datalog queries after every event to figure out the proof DAG of each provable Datalog atom. 
Queries can be slow, so rather than repeatedly making a given query, we just memoize the result the first time and look it up when it is needed again \cite{swift_warren_2012}.
However, as events are allowed to change the database, results of some queries may also change, and thus the memos for those queries become incorrect (stale).  To avoid errors, we currently flush the memo table every time the database is changed.  This obviously reduces the usefulness of the memos.  An implementation improvement for future work is to use more flexible strategies that create memos and update them incrementally through change propagation \citep{acar-08-self,hammer-12,filardo-12-flexible}.

\section{How to Predict Events}\label{app:mbr}\label{app:particle_smoothing}

\Cref{fig:pred_results,fig:pred_ablation_noneural} include a task-based evaluation where we try to predict the \emph{time} and \emph{type} of the next event.  More precisely, for each event in each held-out sequence, we attempt to predict its time given only the preceding events, as well as its type given both its true time and the preceding events.

These figures evaluate the time prediction with average L$_2$ loss (yielding a root-mean-squared error, or \defn{RMSE}) and evaluate the argument prediction with average 0-1 loss (yielding an \defn{error rate}).

To carry out the predictions, we follow \citet{mei-17-neuralhawkes} and use the minimum Bayes risk (MBR) principle to predict the time and type with lowest expected loss.  To predict the $i$\th event:
\begin{itemize}
\item Its time $t_{i}$ has density $p_{i}(t) = \inten{}{t} \exp ( -\int_{t_{i-1}}^{t} \inten{}{t'} \dt' )$.  We choose $\int_{t_{i-1}}^{\infty} t p_{i}(t) \dt$ as the time prediction because it has the lowest expected L$_2$ loss. The integral can be estimated using i.i.d.\ samples of $t_{i}$ drawn from $p_{i}(t)$ as detailed in \citet{mei-17-neuralhawkes} and \citet{mei-19-smoothing}.
\item Since we are given the next event time $t_{i}$ when predicting the type $\mvar{e}_i$,\footnote{\citet{mei-17-neuralhawkes} also give the MBR prediction rule for predicting $\mvar{e}_i$ \emph{without} knowledge of its time $t_i$.} the most likely type is simply $\argmax_{\mvar{e} \in \set{E}(t_i)} \inten{\mvar{e}}{t_{i}}$.  
\end{itemize}

Notice that our approach will never predict an impossible event type. 
For example, $\code{\evt{help}(eve,adam)}$ won't be in $\set{E}(t_i)$ and thus will have \emph{zero} probability if $\sema{rel}{eve,adam}(t_i) = \nullval$ (maybe because \code{eve} stops having \blk{opinion}s on anything that \code{adam} does anymore).

In some circumstances, one might also like to predict the most likely type out of a \emph{restricted} set $\set{E}'(t_i) \subsetneq \set{E}(t_i)$.  This allows one to answer questions like ``If we know that some event \code{\evt{help}(eve,Y)} happened at time $t_i$,
then which person \code{Y} did \code{eve} \evt{help}, given all past events?''  The answer will simply be $\argmax_{\mvar{e} \in \set{E}'(t_i)} \inten{\mvar{e}}{t_{i}}$.

As another extension, \citet{mei-19-smoothing} show how to predict missing events in a neural Hawkes process conditioned on partial observations of both past and future events.  They used a particle smoothing technique that had previously been used for discrete-time neural sequence models \cite{lin-eisner-2018-naacl}.  This technique could also be extended to neural Datalog through time (NDTT):
\begin{itemize}
\item In \defn{particle filtering}, each particle specifies a hypothesized complete history of past events (both observed and missing).  In our setting, this provides enough information to determine the set of possible events $\set{E}(t)$ at time $t$, along with their embeddings and intensities.  
\item \defn{Neural particle smoothing} is an extension where the guess of the next event is also conditioned on the sequence of future events (observed only), using a learned neural encoding of that sequence.  In our setting, it is not clear what embeddings to use for the future events, as we do not in general have static embeddings for our event types, and their dynamic embeddings cannot yet be computed at time $t$.  We would want to learn a compositional encoding of future events that at least respects their structured descriptions (e.g., $\code{\evt{help}(eve,adam)}$), and possibly also draws on the NDTT program and its parameters in some way.  We leave this design to future work.
\end{itemize}

\section{Experimental Details}\label{app:exp_details}

\subsection{Dataset Statistics}\label{app:data_stats}

\Cref{tab:data_stats} shows statistics about each dataset that we use in this paper (\cref{sec:exp}).
\begin{table*}[t]
	\begin{center}
		\begin{small}
			\begin{sc}
				\begin{tabularx}{1.00\textwidth}{l *{1}{S}*{3}{R}*{3}{S}}
					\toprule
					Dataset & \multicolumn{1}{r}{$|\set{K}|$} & \multicolumn{3}{c}{\# of Event Tokens} & \multicolumn{3}{c}{\# of Sequences} \\
					\cmidrule(lr){3-8}
					&  & Train & Dev & Test & Train & Dev & Test \\
					\midrule
					Synthetic $M=4$ & $16$ & $42000$ & $2100$ & $2100$ & $2000$ & $100$ & $100$ \\
					Synthetic $M=8$ & $32$ & $42000$ & $2100$ & $2100$ & $2000$ & $100$ & $100$ \\
					Synthetic $M=16$ & $64$ & $42000$ & $2100$ & $2100$ & $2000$ & $100$ & $100$ \\
					IPTV & $49000$ & $27355$ & $4409$ & $4838$ & $1$ & $1$ & $1$ \\
					RoboCup & $528$ & $2195$ & $817$ & $780$ & $2$ & $1$ & $1$ \\
					\bottomrule
				\end{tabularx}
			\end{sc}
		\end{small}
	\end{center}
	\caption{Statistics of each dataset.}
	\label{tab:data_stats}
\end{table*}

\subsection{Details of Synthetic Dataset and Models}\label{app:synthetic_details}

We synthesized data for \cref{sec:synthetic} by sampling event sequences from the structured NHP specified by our Datalog program in that section.
We chose $N=4$ and $M=4, 8, 16$, and thus end up with three different datasets. 

For each $M$, we set the sequence length $I=21$ and then used the thinning algorithm \citep{mei-17-neuralhawkes,mei-19-smoothing} to sample the first $I$ events over $[0,\infty)$. We set $\dur=t_I$, i.e., the time of the last generated event.
We generated $2000$, $100$ and $100$ sequences for each training, dev and test set respectively. 
We showed the learning curves for $M=8 \text{ and } 16$ in \cref{fig:learncurve_nhp} and left out the plot for $M=4$ because it is boringly similar.

For the unstructured NHP baseline, the program given in \cref{sec:synthetic} is not quite accurate. 
To exactly match the architecture of \citet{mei-17-neuralhawkes}, we have to use the notation of \cref{app:param_share} to ensure that each of the $M N$ event types uses its its own parameters for its embedding and probability:

\vspace{-6pt}
\begin{minipage}[t]{0.49\linewidth}
\begin{lstlisting}[name=nhp_app,firstnumber=auto]
@\code{\boo{is\_process}(1).}@ @\progvdots@
@\code{\boo{is\_process}($M$).}@
\end{lstlisting}
\end{minipage}
\hfill
\begin{minipage}[t]{0.49\linewidth}
\begin{lstlisting}[name=nhp_app,firstnumber=auto]
@\code{\boo{is\_type}(1).}\progvdots@
@\code{\boo{is\_type}($N$).}@
\end{lstlisting}
  \smallskip
\end{minipage}
\begin{lstlisting}[name=nhp_app,firstnumber=auto]
@\code{\dep }\code{ {\embed}(\blk{world}, 8).}@ 
@\code{\dep }\code{ {\embed}(\blk{is\_event}, 8).} \label{line:decl-embed-e}@
@\code{\dep }\code{ {\isevent}(\evto{e}, 0).} \label{line:decl-event-e}@ 
@\code{\blk{is\_event}(M,N)}\code{ \dep } \\ \code{ \boo{is\_process}(M), \boo{is\_type}(N)}\\\code{\use \prm{emb(M,N)}.} \label{line:embed-e}@ 
@\code{\evto{e}(M,N)}\code{ \dep }\\ \code{ \blk{world}, \boo{is\_process}(M), \boo{is\_type}(N)}\\\code{\use \prm{prob(M,N)}.}@
@\code{\blk{world}}\code{ \see }\code{ \exoo{init}.}@
@\code{\blk{world}}\code{ \see }\code{ \evto{e}(M,N), \blk{is\_event}(M,N), \blk{world}.} \label{line:update-world}@ 
\end{lstlisting}

As \cref{sec:synthetic} noted, an event's probability is carried by an \evto{e} fact, but its embedding is carried by an \evt{is\_event} fact.  This is because the NHP uses dynamic event probabilities (which depend on \blk{world}) but static event embeddings (which do not).  Otherwise, we could merge the two by using dimension 8 for \evto{e} in \cref{line:decl-event-e}, and removing \evt{is\_event} by deleting it from \cref{line:update-world} and deleting \cref{line:decl-embed-e,line:embed-e}.

\subsection{Details of IPTV Dataset and our NDTT Model}\label{app:iptv_details}

For the IPTV domain, the time unit is 1 minute.  Thus, in the graph for time prediction, an error of $1.5$ (for example) means an error of $1.5$ minutes.  The exogenous \exoo{release} events were not included in the dataset of \citet{xu-18-online}, but \citeauthor{xu-18-online} (p.c.\@) kindly provided them to us.\looseness=-1

For our experiments in \cref{sec:iptv}, we used the events of days 1--200, days 201--220, and days 221--240 as training, dev and test data respectively---so there is just one long sequence in each case.  (We saved the remaining days for future experiments.)

We evaluated the ability of the trained model to extrapolate from days 1--200 to future events.  That is, for dev and test, we evaluated the model's predictive power on the held-out dev and test events respectively.  However, when predicting each event, the model was still allowed to condition on the \emph{full} history of that event (starting from day 1).  This full history was needed to determine the facts in the database, their embeddings, and the event intensities.

Each observed event has one of the forms
\begin{lstlisting}[name=iptv,firstnumber=1]
@\code{\exoo{init}}@ 
@\code{\exoo{release}(\dvar{P})}@
@\code{\evt{watch}(\dvar{U},\dvar{P})}@
\end{lstlisting}
For example, \code{\evt{watch}(u4,p49)} occurs whenever
user \code{u4} \evt{watch}es television program \code{\dvar{p49}}. 

The dataset also provides time-invariant facts of the form
\begin{lstlisting}[name=iptv,firstnumber=auto]
@\code{has\_tag(\dvar{P},\dvar{T})}@
\end{lstlisting}
which tag programs with attributes.\footnote{Users could also have tags, to record their demographics or interests.  However, the IPTV dataset does not provide such tags.}  For example:
\begin{lstlisting}[name=iptv,firstnumber=auto]
@\code{\boo{has\_tag}(p1,comedy).} \progvdots@
@\code{\boo{has\_tag}(p49,romance).}@
\end{lstlisting}

We develop our NDTT program as follows.  
A television program is added to the database only when it is released:
\begin{lstlisting}[name=iptv,firstnumber=auto]
@\code{\blk{program}(\dvar{P}) \see \exoo{release}(\dvar{P}).} \label{line:release}@ 
\end{lstlisting}
Now that \dvar{P} is a program, it can be watched:
\begin{lstlisting}[name=iptv,firstnumber=auto]
@\code{\evt{watch}(\dvar{U},\dvar{P}) \dephighway \blk{user}(\dvar{U}), \blk{program}(\dvar{P}).} \label{line:watch}@
\end{lstlisting}
The probability of a watch event depends on the current embeddings of the user and the program:
\begin{lstlisting}[name=iptv,firstnumber=auto]
@\code{\embed(\blk{user}, 8).}@
@\code{\embed(\blk{program}, 8).}@
\end{lstlisting}

Of course, we have to declare that `watch' is an event:
\begin{lstlisting}[name=iptv,firstnumber=auto]
@\code{event(\evt{watch},8).}@
\end{lstlisting}
Notice that we equipped \evt{watch} with a 8-dimensional embedding as well as a probability.  The embedding encodes some details of the event (who watched what).  This detailed watch event then updates what we know about both the user and the program, in order to predict future watch events:
\begin{lstlisting}[name=iptv,firstnumber=auto]
@\code{\blk{user}(\dvar{U}) \ \ \ \see \evt{watch}(\dvar{U},\dvar{P}).} \label{line:updateu}@
@\code{\blk{program}(\dvar{P}) \see \evt{watch}(\dvar{U},\dvar{P}).} \label{line:updatep}@
\end{lstlisting}
The \dephighway connector in \cref{line:watch} requested highway connections around \evt{watch}  (\cref{app:highway}), so these update \cref{line:updateu,line:updatep} not only consider $\seme{watch}{\dvar{U},\dvar{P}}$ but also directly consider $\sema{user}{\dvar{U}}$ and $\sema{program}{\dvar{P}}$.  This is similar to a traditional LSTM update, and in our initial pilot experiments we found it to work better than simply using \dep in \cref{line:watch}.

Where do the \blk{user} facts come from?  \Cref{line:updateu} would automatically add \code{\blk{user}(\dvar{U})} to the database upon the first time they watched a program.  But such an event \code{\evt{watch}(\dvar{U},\dvar{P})} is not itself possible (\cref{line:watch}) until \code{\blk{user}(\dvar{U})} is already in the database.  To break this circularity, we must populate the database with users in advance.  

If we simply declared these users as
\begin{lstlisting}[name=iptv,firstnumber=auto]
@\code{\blk{user}(u1).}@
@\code{\blk{user}(u2).}@@\progvdots@
\end{lstlisting}
then the model would include separate parameters for each of these rules.  However, fitting user-specific parameters would be hard for users who have only a small amount of data.  Instead, we make all the user rules share parameters (see \cref{app:param_share}):
\begin{lstlisting}[name=iptv,firstnumber=auto]
@\code{\blk{user}(u1) \use \prm{user\_init}.}@
@\code{\blk{user}(u2) \use \prm{user\_init}.}@@\progvdots@
\end{lstlisting}
Thus, all users start out in the same place,\footnote{\label{fn:nouserinit}We suspect that it would have been adequate for that initial user embedding to be the \vec{0} vector, which we could have specified by writing \code{\use~\prm{0}} instead of \code{\use~\prm{user\_init}}.  That is how we treated programs in this model (\cref{line:p1} below), and how we treated both users and programs in \cref{app:iptv_details_baseline}.  We regret the discrepancy.}  and a user’s embedding only depends entirely on programs that they’ve watched so far.  An update to the user's embedding  (\cref{line:updateu}) could be either material or epistemic: that is, it may reflect \emph{actual} changes over time in the user's taste, or merely changes in our \emph{knowledge} of the user's taste.  Ultimately, the training procedure learns whatever updates help the model to better predict the user's future \evt{watch} events.

There is one more subtlety regarding user embeddings.  In the program above, \code{\blk{user}(u1)} is true at all times, but is ``launched'' (in the sense of \cref{sec:ndtt_math}) only by the first event of the form \code{\evt{watch}(u1,\dvar{P})}.  Thus, we learn nothing about the user from the fact that time has elapsed without their having yet watched any programs: they do not yet have a cell block that can drift to track the passage of time.  To fix this, we add the following rule so that all users are simultaneously launched at time 0 by the exogenous \exoo{init} event:
\begin{lstlisting}[name=iptv,firstnumber=auto]
@\code{\blk{user}(\dvar{U}) \see \exoo{init}, \blk{user}(\dvar{U}).}@
\end{lstlisting}
This ensures that the user has an LSTM cell block starting at time 0, which can drift to mark the passage of time even before the user has watched any programs.  This rule for users is analogous to \cref{line:release} for programs.

Where do the \blk{program} facts come from?  We declare them much as we declared the \blk{user} facts:\footnote{Actually, if \code{p1} has at least one tag, then we can omit \cref{line:p1} because \cref{line:provep} below will be enough to prove that \code{p1} is a program.  In the IPTV dataset, every program does have at least one tag, so we omit all rules like \ref{line:p1}, which do not affect the facts or their embeddings.}
\begin{lstlisting}[name=iptv,firstnumber=auto]
@\code{\blk{program}(p1) \use \prm{0}.} \label{line:p1}@
@\code{\blk{program}(p2) \use \prm{0}.}@@\progvdots@
\end{lstlisting}
However, a program's embedding should also be affected by its tags:\footnote{Recall that facts like \code{has\_tag(p1,comedy)} were declared in the initial database, have no embeddings, and never change.}
\begin{lstlisting}[name=iptv,firstnumber=auto]
@\code{\blk{program}(\dvar{P})  \dep  has\_tag(\dvar{P},\dvar{T}), \blk{tag}({\dvar{T}})}. \label{line:provep}@
\end{lstlisting}
where each tag is declared separately:
\begin{lstlisting}[name=iptv,firstnumber=auto]
@\code{\embed(tag, 8).}@
@\code{\blk{tag}(adventure).} \label{line:adventure}@
@\code{\blk{tag}(comedy).} \label{line:comedy}@@\progvdots@
\end{lstlisting}
Note that the rules like \ref{line:adventure} and \ref{line:comedy} introduce tag-specific parameters.  For example, the bias vector of \cref{line:adventure} provides an embedding of the \code{adventure} tag.  As each tag has a lot of data, these tag-specific parameters should be easier to learn than user-specific parameters.

The initial embedding of a tag is then affected by who watches programs with that tag, and when.  In other words, just as the \code{\evt{watch}} events update our understanding of individual users,
they also track how the meaning of each tag changes over time:
\begin{lstlisting}[name=iptv,firstnumber=auto]
@\code{\blk{tag}(\dvar{T}) \see \exoo{init}, \blk{tag}(\dvar{T}).}@
@\code{\blk{tag}(\dvar{T}) \see \evt{watch}(\dvar{U},\dvar{P}), has\_tag(\dvar{P},\dvar{T}), \blk{tag}(\dvar{T}).} \label{line:updatet}@
\end{lstlisting}
As before, these updates are rich because the \evt{watch} event has an embedding and also supplies highway connections.

We finish with a final improvement to the model.  Above, \code{\blk{program}(\dvar{P})} is affected both by \code{\dvar{P}}'s tags via the \code{\dep} \cref{line:provep} and by its history of \code{\evt{watch}} events via the  \see  \cref{line:updatep}.  The NDTT equations would simply add these influences via rule (7).  Instead, we edit the program to combine these influences nonlinearly.  This gives a deeper architecture:

\begin{lstlisting}[name=iptv,firstnumber=auto]
@\code{\blk{program}(\dvar{P}) \dep \\\blk{program\_profile}(\dvar{P}), \blk{program\_history}(\dvar{P}).} \label{line:iptv-deeper-start}@
@\code{\blk{program\_profile}(\dvar{P}) \dep has\_tag(\dvar{P},\dvar{T}), \blk{tag}(\dvar{T}).} \label{line:provep-new}@
@\code{\blk{program\_history}(\dvar{P}) \see \exoo{release}(\dvar{P}).} \label{line:release-new}@
@\code{\blk{program\_history}(\dvar{P}) \see \\\evt{watch}(\dvar{U},\dvar{P}), \blk{user}(\dvar{U}), \blk{program}(\dvar{P}).} \label{line:updatep-new}\label{line:iptv-deeper-end}@
\end{lstlisting}
where \cref{line:provep-new,line:release-new,line:updatep-new} replace rules \ref{line:provep}, \ref{line:release}, and \ref{line:updatep} respectively.

In principle, facts with different functors can be embedded in vector spaces of different dimensionality, as needed.  But in all of our experiments, we used the same dimensionality for all functors, so as to have only a single hyperparameter to tune.  If the hyperparameter were 8, for example, our Datalog program would have the declarations
\begin{lstlisting}[name=iptv,firstnumber=auto]
@\code{\dep }\code{ {\embed}(\blk{user}, 8).} @
@\code{\dep }\code{ {\embed}(\blk{program}, 8).} @
@\code{\dep }\code{ {\embed}(\blk{profile}, 8).} @
@\code{\dep }\code{ {\embed}(\exoo{released}, 0).} @
@\code{\dep }\code{ {\embed}(\blk{watchhistory}, 8).} @
@\code{\dep }\code{ {\embed}(\blk{tag}, 8).} @
@\code{{\dep}  {\isevent}(\evt{watch}, 8).}@
\end{lstlisting}
where \code{\evt{watch}} has an extra dimension for its intensity.
The hyperparameter tuning method and its results are described in \cref{app:training_details} below.  

\subsection{Baseline Programs on IPTV Dataset}\label{app:iptv_details_baseline}

We also implemented baseline models that were inspired by the Know-Evolve \cite{trivedi-17-know} and DyRep \cite{trivedi-19-dyrep} frameworks.  Our architectures are not identical: for example, our \cref{line:watch-ke-prob} below models each event probability using a feed-forward network in place of a bilinear function.  However, Trivedi (p.c.\@) agrees that the architectures are similar.  Note that these prior papers did not apply their frameworks specifically to the IPTV dataset (nor to RoboCup).

The Know-Evolve and DyRep programs specify the same \blk{user}, \blk{program}, and \boo{has\_tag} facts as in \cref{app:iptv_details}, except that the initial embedding \code{user\_init} is fixed to $\vec{0}$ (see \cref{fn:nouserinit}).

The Know-Evolve program continues as follows.
\begin{lstlisting}[name=iptv_baseline,firstnumber=auto]
\end{lstlisting}

Whereas a \evt{watch} fact in \cref{app:iptv_details} carried both a probability and an embedding, here we split off the embedding into a separate fact and compute it differently from the probability, to be more similar to \citet{trivedi-17-know}:
\begin{lstlisting}[name=iptv_baseline,firstnumber=auto]
@\code{{\dep}  {\isevent}(\evto{watch}, 0).}@
@\code{{\dep}  {\embed}(\blk{watch\_emb}, 8).}@
@\code{\evto{watch}(\dvar{U},\dvar{P})}\code{ \dep }\code{ \blk{user}(\dvar{U}), \blk{program}(\dvar{P}).} \label{line:watch-ke-prob}@
@\code{\blk{watch\_emb}(\dvar{U},\dvar{P}) \dep \\ \blk{user}(\dvar{U}) \usec \prm{pair}, \blk{program}(\dvar{P}) \usec \prm{pair}.} \label{line:watch-emb}@
\end{lstlisting}
Notice that \cref{line:watch-emb} in effect multiplies the sum  $\sema{user}{\dvar{U}}+\sema{program}{\dvar{P}}$ by the \prm{pair} matrix before applying $\tanh$.

The cell blocks are now launched and updated as follows:
\begin{lstlisting}[name=iptv_baseline,firstnumber=auto]
@\code{\blk{user}(\dvar{U}) \ \ \ \see \exoo{init}, \blk{user}(\dvar{U}).} @
@\code{\blk{program}(\dvar{P}) \see \exoo{init}, \blk{program}(\dvar{P}).} \label{line:initp-ke}@
@\code{\blk{user}(\dvar{U}) \ \ \ \see \evto{watch}(\dvar{U},\dvar{P}), \blk{watch\_emb}(\dvar{U},\dvar{P}).} \label{line:updateu-ke}@
@\code{\blk{program}(\dvar{P}) \see \evto{watch}(\dvar{U},\dvar{P}), \blk{watch\_emb}(\dvar{U},\dvar{P}).} \label{line:updatep-ke}@
\end{lstlisting}
Of course, when the embedding of \code{\blk{user}(\dvar{U})} or \code{\blk{program}(\dvar{P})}
is updated, the embedding of \code{\blk{watch\_emb}(\dvar{U},\dvar{P})} also changes to reflect this.

What are the differences from \cref{app:iptv_details}?  Since \citet{trivedi-17-know} did not support changes over time to the set of possible events, we omitted this feature from our Know-Evolve program above.  Specifically, the program does not use the \exoo{release} events in the dataset---it treats all programs as having been released by \exoo{init} at time 0.  The program also has no highway connections, nor the deeper architecture at \crefrange{line:iptv-deeper-start}{line:iptv-deeper-end} of \cref{app:iptv_details}, and it does not make use of the program tags.

Our DyRep version of the program makes a few changes to follow the principles of \cite{trivedi-19-dyrep}.  The main ideas of DyRep are as follows:
\begin{itemize}[noitemsep]
\item Entities are represented as nodes in a graph (here: programs, users, and tags).
\item Each node has an embedding.
\item The properties of an entity are represented by labeled edges that link it to other nodes (here: \code{\boo{has\_tag}(\dvar{P},\dvar{T})}).
\item The graph structure can change due to exogenous forces (see \cref{line:release-dyrep} below).
\item Any pair of entities can communicate at any time.  (These communications are the events in our temporal event sequences, such as \code{\evto{watch}(\dvar{U},\dvar{P})}.)
\item The probability of an event depends on the embeddings of the two nodes that communicate (here: \cref{line:watch-ke-prob}).
\item When an event occurs, it updates the embeddings of (only) the two nodes that communicate (see \cref{line:updateu-dyrep,line:updatep-dyrep} below).
\item An update to a node's embedding also considers the embeddings of its neighbors in the graph (see \cref{line:seetag-dyrep} below).  
\end{itemize}
Thus, we replace \crefrange{line:initp-ke}{line:updatep-ke} above with
\begin{lstlisting}[name=iptv_baseline,firstnumber=auto]
@\code{\blk{program}(\dvar{P}) \see \exoo{release}(\dvar{P}).} \label{line:release-dyrep}@  
@\code{\blk{user}(\dvar{U}) \ \ \ \see \evto{watch}(\dvar{U},\dvar{P}),  \blk{user}(\dvar{U}) \ \ \ \use \prm{event}.$\!$} \label{line:updateu-dyrep}@  
@\code{\blk{program}(\dvar{P})} \code{\see }\code{ \evto{watch}(\dvar{U},\dvar{P}), \blk{program}(\dvar{P}) \use \prm{event}.} \label{line:updatep-dyrep}@ 
\end{lstlisting}
Thus, DyRep now permits the set of watchable programs (nodes) to change over time, but the \blk{user} and \blk{program} updates are less well-informed than in Know-Evolve: the updates to the user embedding no longer look at the current program embedding, nor vice-versa.\footnote{To allow better-informed updates within the DyRep formalism, we could have included edges between all users and all programs.  But then every update would depend on all users and all programs---which is exactly the ``everything-affects-everything'' problem that our paper aims to cure (\cref{sec:intro})!}
Indeed, DyRep no longer uses \blk{watch\_emb} and can drop \cref{line:watch-emb}.

Where our Know-Evolve program did not use tags, our DyRep program can encode tags using \boo{has\_tag} edges.  Thus, when a program \dvar{P} is watched, 
the update to the program's embedding depends in part on its tags:
\begin{lstlisting}[name=iptv_baseline,firstnumber=auto]
@\code{\blk{program}(\dvar{P})} \code{\see }\\ \code{ \evto{watch}(\dvar{U},\dvar{P}), \blk{tag}(\dvar{T}), \boo{has\_tag}(\dvar{P},\dvar{T}).} \label{line:seetag-dyrep}@
\end{lstlisting}
The embedding $\sema{tag}{\dvar{T}}$ is defined as in our full model of \cref{app:iptv_details},
except that it is now static (except for drift).  It is no longer updated by watch events, because the \code{\evto{watch}(\dvar{U},\dvar{P})} event only updates \dvar{U} and \dvar{P}.  
In contrast, the Datalog \cref{line:updatet} in \cref{app:iptv_details} was able to draw \dvar{T} into the computation by joining \code{\evt{watch}(\dvar{U},\dvar{P})} to \code{\boo{has\_tag}(\dvar{P},\dvar{T})}.

\subsection{Details of RoboCup Dataset and our NDTT Model}\label{app:robocup_details}

For the RoboCup domain, the time unit is 1 second.  Thus thus in the graph for time prediction, an error of $1.5$ (for example) means an error of $1.5$ seconds.

For our experiments in \cref{sec:robocup}, we used Final 2001 and 2002, Final 2003, and Final 2004 as training, dev, and test data respectively.  Each sequence is a single game and each dataset contains multiple sequences.

Each observed event has one of the forms
\begin{lstlisting}[name=robo,firstnumber=1]
@\code{\evt{kickoff}(\dvar{P})}@
@\code{\evt{kick}(\dvar{P})}@
@\code{\evt{goal}(\dvar{P})}@
@\code{\evt{pass}(\dvar{P},\dvar{Q})}@
@\code{\evt{steal}(\dvar{Q},\dvar{P})}@
@\code{\exoo{init}}@
\end{lstlisting}
which we will describe shortly.  The database also contains facts about the teams.  There are 2 teams, each with 11 robot players.  Any pair of players \code{\dvar{P}} and \code{\dvar{Q}} are either teammates or opponents:
\begin{lstlisting}[name=robo,firstnumber=auto]
@\code{teammate(\dvar{P},\dvar{Q}) \dep \\ in\_team(\dvar{P},\dvar{T}), in\_team(\dvar{Q},\dvar{T}), not\_eq(\dvar{P},\dvar{Q}).} \label{line:robocup-facts-first}@
@\code{opponent(\dvar{P},\dvar{Q}) \dep \\ in\_team(\dvar{P},\dvar{T}), in\_team(\dvar{Q},\dvar{S}), not\_eq(\dvar{T},\dvar{S}).}@
\end{lstlisting}
These relations are induced using the database facts
\begin{lstlisting}[name=robo,firstnumber=auto]
@\code{in\_team(a1,a).} \progvdots@
@\code{in\_team(a11,a).}@
@\code{in\_team(b1,b).} \progvdots@
@\code{in\_team(b11,b).}@
\end{lstlisting}
together with an inequality relation on entities, \code{not\_eq}, which can be spelled out with a quadratic number of additional facts if the Datalog implementation does not already provide it as a built-in relation:
\begin{lstlisting}[name=robo,firstnumber=auto]
@\code{not\_eq(a1, a2).} \text{\ \ \ \ \% players}@
@\code{not\_eq(a1, a3).} \progvdots@
@\code{not\_eq(b11, b10).}@
@\code{not\_eq(a, b).} \text{\ \ \ \ \ \ \% teams}@
@\code{not\_eq(b, a).} \label{line:robocup-facts-last}@
\end{lstlisting}

We allow the ball to be in the possession of either a specific player, or a team as a whole.
A game starts with team \code{a} taking possession of the ball:\footnote{It is a convention in the IPTV dataset that team \code{a} is the one that takes possession first.  If the starting team were decided by a coin flip, then we would use the ``event groups'' extension in \cref{app:simult} to decide whether \exoo{init} causes \code{has\_ball(a)} or \code{has\_ball(b)}.  This would allow us to learn the weight of the coin (for example, on the IPTV dataset, we would learn that the coin \emph{always} chooses team \code{a}); or if we knew it was a fair coin, we could model that by declaring that certain parameters are 0.}
\begin{lstlisting}[name=robo,firstnumber=auto]
@\code{has\_ball(a) \see \exoo{init}.}@
\end{lstlisting}
A random player \code{\dvar{P}} in team \code{a} now assumes possession of the ball, taking it from the team as a whole.\footnote{Notice that in our program, the possible kickoff events all have equal intensity, leading to a uniform distribution over players $\code{a1}, \ldots, \code{a11}$.  We will learn that this intensity is high, since the kickoff happens at a time close to 0.}
  This is called a \evt{kickoff} event, although in RoboCup---unlike human soccer---\code{\dvar{P}} does not kick the ball off into the distance but retains it.
\begin{lstlisting}[name=robo,firstnumber=auto]
@\code{\evt{kickoff}(\dvar{P}) \dep in\_team(\dvar{P},\dvar{T}), has\_ball(\dvar{T}).}@
@\code{!has\_ball(\dvar{T}) \see \evt{kickoff}(\dvar{P}), in\_team(\dvar{P},\dvar{T}).}@
@\code{has\_ball(\dvar{P}) \ \see \evt{kickoff}(\dvar{P}).}@
\end{lstlisting}
Thereafter, the player who has possession of the ball can kick it to a nearby location while retaining possession (``dribbling''),
\begin{lstlisting}[name=robo,firstnumber=auto]
@\code{\evt{kick}(\dvar{P}) \dep has\_ball(\dvar{P}).} \label{line:kick}@ 
\end{lstlisting}
or can pass the ball to a teammate,
\begin{lstlisting}[name=robo,firstnumber=auto]
@\code{\evt{pass}(\dvar{P},\dvar{Q}) \dep has\_ball(\dvar{P}), teammate(\dvar{P},\dvar{Q}).} \label{line:pass}@
@\code{!has\_ball(\dvar{P}) \see \evt{pass}(\dvar{P},\dvar{Q}).}@
@\code{has\_ball(\dvar{Q}) \ \see \evt{pass}(\dvar{P},\dvar{Q}).}@
\end{lstlisting}
or can score a goal,
\begin{lstlisting}[name=robo,firstnumber=auto]
@\code{\evt{goal}(\dvar{P}) \dep has\_ball(\dvar{P}).}@
\end{lstlisting}
Scoring a goal instantly updates the database to transfer the ball to the other team,
\begin{lstlisting}[name=robo,firstnumber=auto]
@\code{!has\_ball(\dvar{P}) \see \evt{goal}(\dvar{P}).}@
@\code{has\_ball(\dvar{S}) \ \see \evt{goal}(\dvar{P}), in\_team(\dvar{P},\dvar{T}), not\_eq(\dvar{T},\dvar{S}).}@
\end{lstlisting}
after which someone in the other team can kick off the ball and continue the game. 
When a player \code{\dvar{P}} has the ball, a player \code{\dvar{Q}} in the other team can steal it:
\begin{lstlisting}[name=robo,firstnumber=auto]
@\code{\evt{steal}(\dvar{Q},\dvar{P}) \dep has\_ball(\dvar{P}), opponent(\dvar{P},\dvar{Q}).}@
@\code{!has\_ball(\dvar{P}) \see \evt{steal}(\dvar{Q},\dvar{P}).}@
@\code{has\_ball(\dvar{Q}) \ \see \evt{steal}(\dvar{Q},\dvar{P}).}@
\end{lstlisting}

In our experiments, we got the best results by declaring non-zero embeddings of both teams and players, such as
\begin{lstlisting}[name=robo,firstnumber=auto]
@\code{\dep }\code{ {\embed}(\blk{team}, 8).}@
@\code{\dep }\code{ {\embed}(\blk{player}, 8).} @
\end{lstlisting}
Since there are only two teams, the embeddings of the two teams jointly serve as a kind of global state---but one that may be smaller than the global state we would use for a simple NHP model.  In our actual experiments (\cref{sec:robocup}), hyperparameter search (\cref{app:training_details}) chose 32-dimensional NDTT embeddings, giving a total of 64 dimensions for the pair of teams.  In contrast, it chose a 128-dimensional global state for the simple NHP baseline model.

Ideally, we would like the embedding $\sema{player}{\dvar{P}}$ to track our probability distribution over the state of the robot player, such as its latent position on the field and its latent energy level.  We would also like the embedding of a team to track our probability distribution over the state of the team and the latent position of the ball.  We do not observe these latent properties in our dataset.  However, they certainly affect the progress of the game.  For example, if two players pass or steal, they must be near each other; so if we have \code{\evt{pass}(\dvar{P},\dvar{Q})} and \code{\evt{steal}(\dvar{R},\dvar{Q})} nearby in time, then by the triangle inequality, \code{\dvar{P}} and \code{\dvar{R}} must be close together, which raises the probability of \code{\evt{steal}(\dvar{P},\dvar{R})}.  Changes in the mean and variance of these probability distributions are then tracked by updates and drift of the embeddings, with the variance generally decreasing when an event occurs (because it gives information) and increasing between events (because uncertainty about the latent changes accumulates over time, as in a drunkard’s walk).

The team and player embeddings are launched at time 0 using the exogenous \exoo{init} event:
\begin{lstlisting}[name=robo,firstnumber=auto]
@\code{\blk{team}(\dvar{T}) \ \ \see \exoo{init}, in\_team(P,T).}@
@\code{\blk{player}(\dvar{P}) \see \exoo{init}, in\_team(P,T).}@
\end{lstlisting}

A player’s embedding is updated whenever that player participates in an event.  We elected to reduce the number of parameters by sharing parameters not only across players, but also across similar kinds of events (this was also done by the prior work DyRep).  
\begin{lstlisting}[name=robo,firstnumber=auto]
@\code{\blk{player}(\dvar{P}) \see \evt{kickoff}(\dvar{P}) \use \prm{individual}.}@
@\code{\blk{player}(\dvar{P}) \see \evt{kick}(\dvar{P}) \ \ \ \use \prm{individual}.} \label{line:kickupdate}@
@\code{\blk{player}(\dvar{P}) \see \evt{goal}(\dvar{P}) \ \ \ \use \prm{individual}.}@
@\code{\blk{player}(\dvar{P}) \see \evt{pass}(\dvar{P},\dvar{Q}) \ \use \prm{individual\_agent}.} \label{line:passsteal-start}@
@\code{\blk{player}(\dvar{Q}) \see \evt{pass}(\dvar{P},\dvar{Q}) \ \use \prm{individual\_patient}.}@
@\code{\blk{player}(\dvar{Q}) \see \evt{steal}(\dvar{Q},\dvar{P}) \use \prm{individual\_agent}.}@
@\code{\blk{player}(\dvar{P}) \see \evt{steal}(\dvar{Q},\dvar{P}) \use \prm{individual\_patient}.} \label{line:passsteal-end}@
\end{lstlisting}
The parameter sharing notation was explained in \cref{app:param_share}.  The above rules use the linguistic names ``agent'' and ``patient'' to refer to the player who acts and the player who is acted upon, respectively.

A team’s embedding is also updated when any player acts.  We could have done this by saying that the team’s embedding pools over all of its players, so it is updated when they are updated,
\begin{lstlisting}[name=robo,firstnumber=auto]
@\code{\blk{team}(\dvar{T}) \dep \blk{player}(\dvar{P}), in\_team(\dvar{P},\dvar{T})}@
\end{lstlisting}
but instead we directly updated the team embeddings using update rules parallel to the ones above.  For example, \cref{line:kickupdate} also has a variant that affects not the player \code{\dvar{P}} that kicked the ball, but that player's team \code{\dvar{T}}, as well as a second variant that affects the opposing team.
\begin{lstlisting}[name=robo,firstnumber=auto]
@\code{\blk{team}(\dvar{T}) \see\\ \evt{kick}(\dvar{P}), in\_team(\dvar{P},\dvar{T}) \use \prm{team}.} \label{line:kickteam}@
@\code{\blk{team}(\dvar{S}) \see\\ \evt{kick}(\dvar{P}), in\_team(\dvar{P},\dvar{T}), not\_eq(\dvar{T},\dvar{S}) \\ \use \prm{team\_other}.} \label{line:kickteam-other}@
\end{lstlisting}
We similarly have variants of \crefrange{line:passsteal-start}{line:passsteal-end}:
\begin{lstlisting}[name=robo,firstnumber=auto]
@\code{\blk{team}(\dvar{T}) \see\\ \evt{pass}(\dvar{P},\dvar{Q}), in\_team(\dvar{P},\dvar{T}) \\\use \prm{team\_agent}.}@
@\code{\blk{team}(\dvar{S}) \see\\ \evt{pass}(\dvar{P},\dvar{Q}), in\_team(\dvar{P},\dvar{T}), not\_eq(\dvar{T},\dvar{S}) \\\use \prm{team\_nonagent}.} \label{line:nonpatient}@
@\code{\blk{team}(\dvar{T}) \see\\ \evt{steal}(\dvar{P},\dvar{Q}), in\_team(\dvar{P},\dvar{T}) \\\use \prm{team\_agent}.}@
@\code{\blk{team}(\dvar{S}) \see\\ \evt{steal}(\dvar{P},\dvar{Q}), in\_team(\dvar{P},\dvar{T}), not\_eq(\dvar{T},\dvar{S}) \\\use \prm{team\_nonagent}.}@
\end{lstlisting}
Here ``non-agent'' refers to the team that does not contain the agent (in the case of \cref{line:nonpatient}, it does not contain the patient either).

Finally, we can improve the model by enriching the dependencies.  
Earlier, we embedded the \evt{kick} event using \cref{line:kick}, repeated here:
\begin{lstlisting}[name=robo,firstnumber=auto]
@\code{\evt{kick}(\dvar{P}) \dep has\_ball(\dvar{P}).}@
\end{lstlisting}
But then the probability that robot player \code{\dvar{P}} kicks at time $t$ (if it has the ball) would be constant with respect to both \dvar{P} and $t$.  We want to make this probability  sensitive to the states at time $t$ of the player \dvar{P}, the player’s team \dvar{T}, and the other team \dvar{S}.   So we modify the rule to add those facts as conditions (in blue):
\begin{lstlisting}[name=robo,firstnumber=auto]
@\code{\evt{kick}(\dvar{P}) \dephighway\\ has\_ball(\dvar{P}), \blk{player}(\dvar{P}), \blk{team}(\dvar{T}), \blk{team}(\dvar{S}), in\_team(\dvar{P},\dvar{T}), not\_eq(\dvar{T},\dvar{S}).} \label{line:kickhighway}@
\end{lstlisting}

Because this rule uses \dephighway to request highway connections, all three of these states will also be consulted directly when a \code{\evt{kick}(\dvar{P})} event updates the states of player \code{\dvar{P}} and both teams (via \cref{line:kick,line:kickteam,line:kickteam-other}).  To deepen the network, we further give the event \code{\evt{kick}(\dvar{P})} its own embedding, which is a nonlinear combination of all of these states, and which is also consulted when the event causes an update.
\begin{lstlisting}[name=robo,firstnumber=auto]
@\code{ \dep \isevent(\evt{kick},8).}@
\end{lstlisting}

We handle the other event types similarly to \evt{kick}.  In the case of an event that involves two players \dvar{P} and \dvar{Q}, we also add the state of player \dvar{Q} (the patient) as a fourth blue condition.  For example, we expand the old \cref{line:pass} to
\begin{lstlisting}[name=robo,firstnumber=auto]
@\code{\evt{pass}(\dvar{P},\dvar{Q}) \dep\\ has\_ball(\dvar{P}), teammate(\dvar{P},\dvar{Q}), \blk{player}(\dvar{P}), \blk{player}(\dvar{Q}), \blk{team}(\dvar{T}), \blk{team}(\dvar{S}), has\_ball(\dvar{P}), in\_team(\dvar{P},\dvar{T}), not\_eq(\dvar{T},\dvar{S}).}@
\end{lstlisting}

\begin{figure*}[!ht]
	\begin{center}

		\begin{subfigure}[t]{1.00\linewidth}
			\begin{center}
				\includegraphics[width=0.32\linewidth]{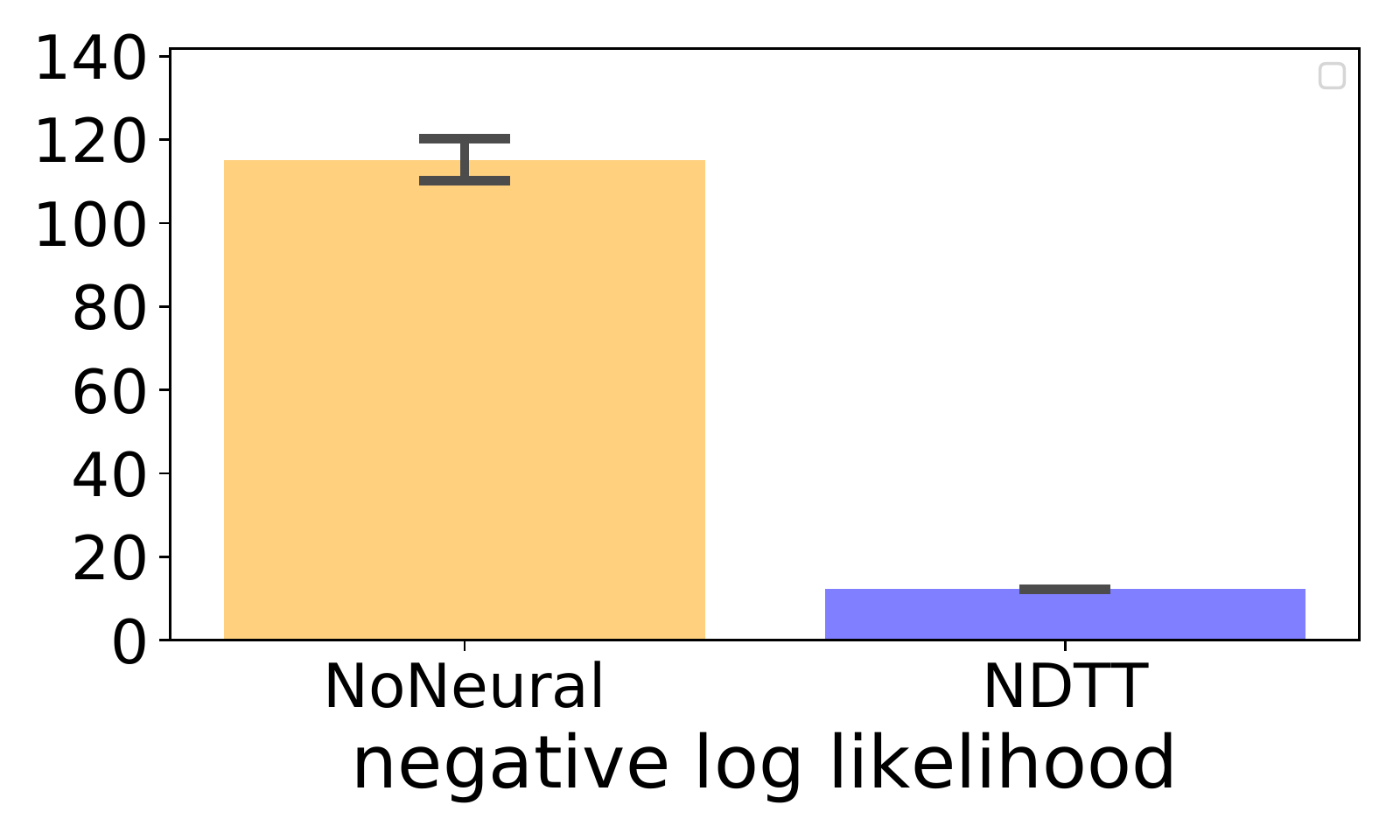}
				~
				\includegraphics[width=0.32\linewidth]{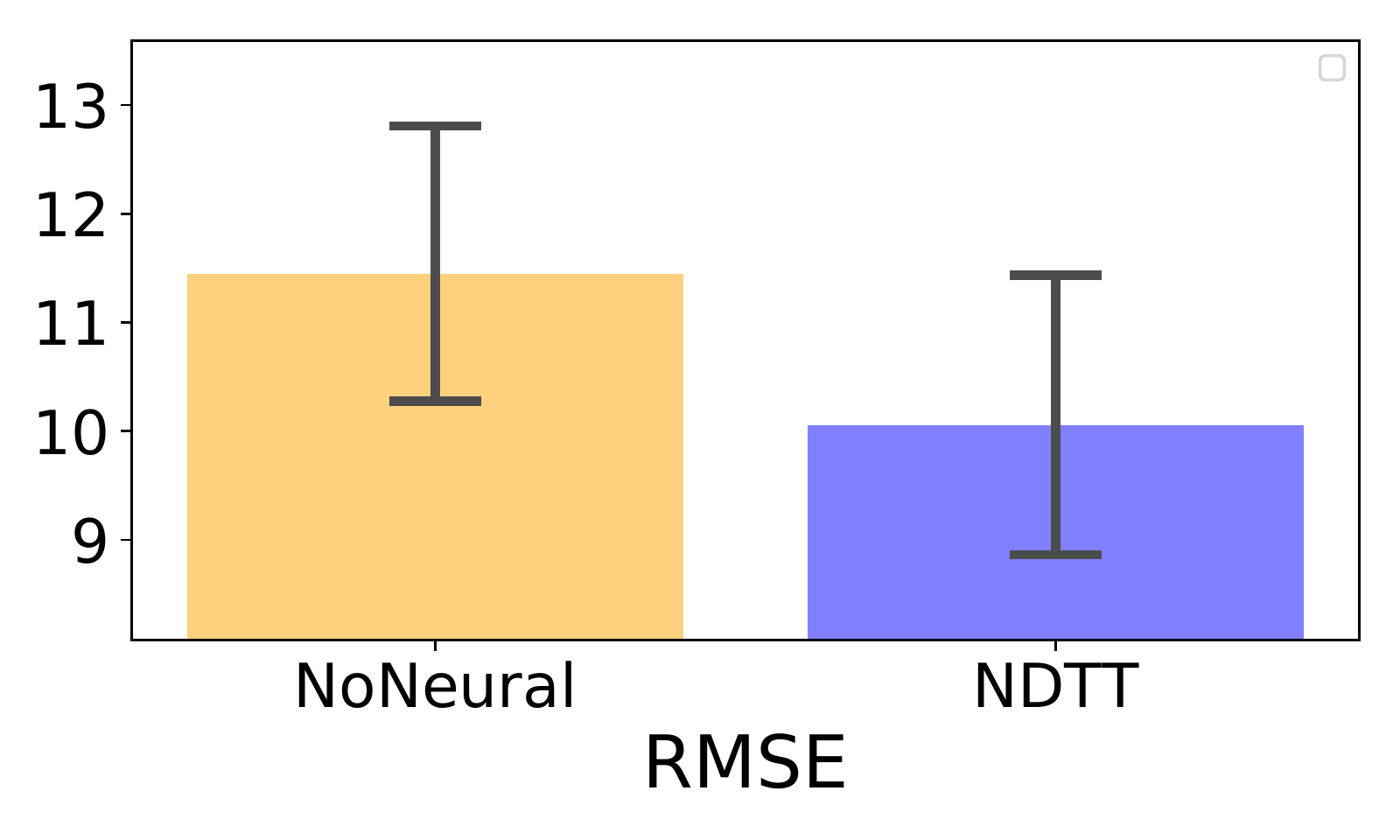}
				~
				\includegraphics[width=0.32\linewidth]{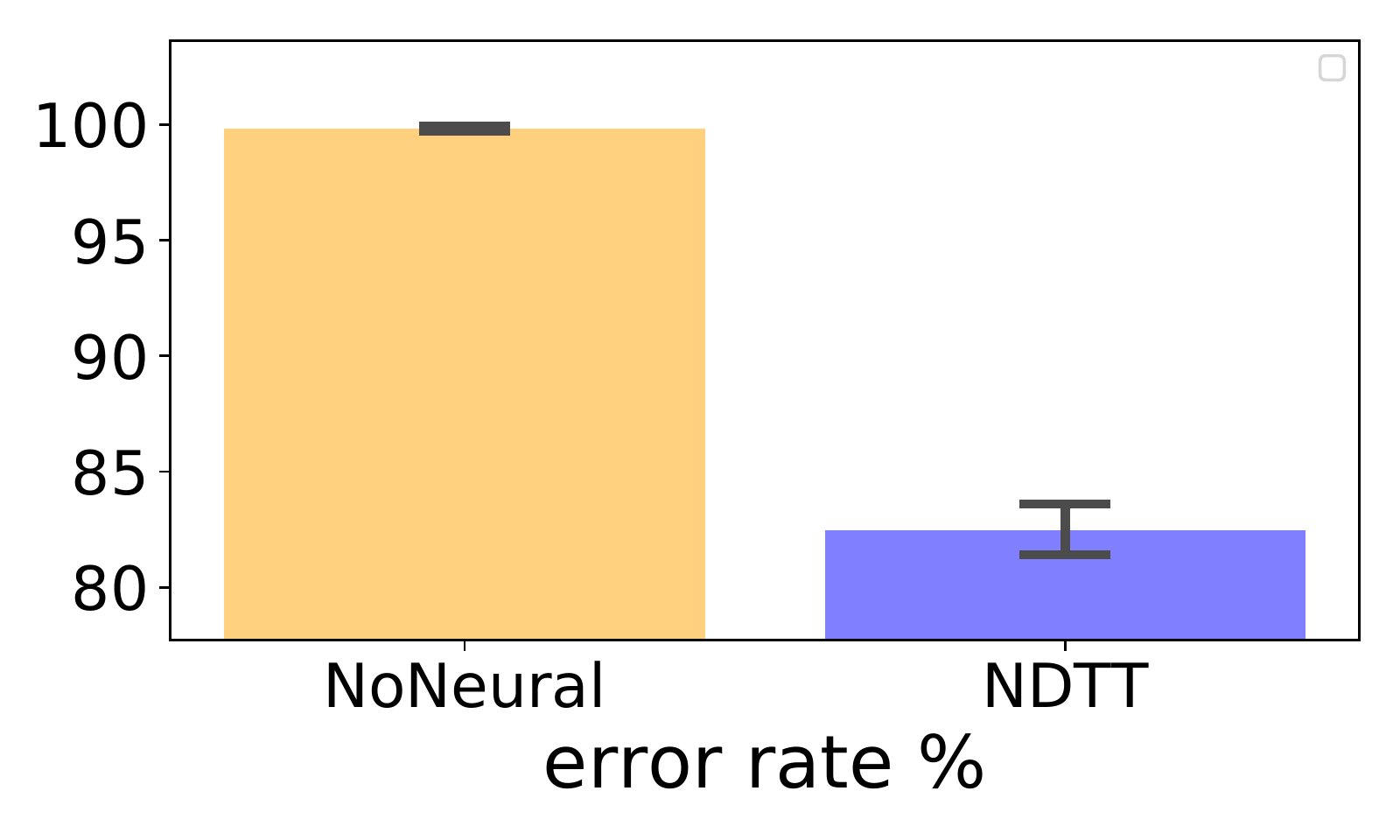}
				\vspace{-6pt}
				\caption{IPTV Dataset}\label{fig:pred_iptv_noneural}
			\end{center}
		\end{subfigure}

		\begin{subfigure}[t]{1.00\linewidth}
			\begin{center}
				\includegraphics[width=0.32\linewidth]{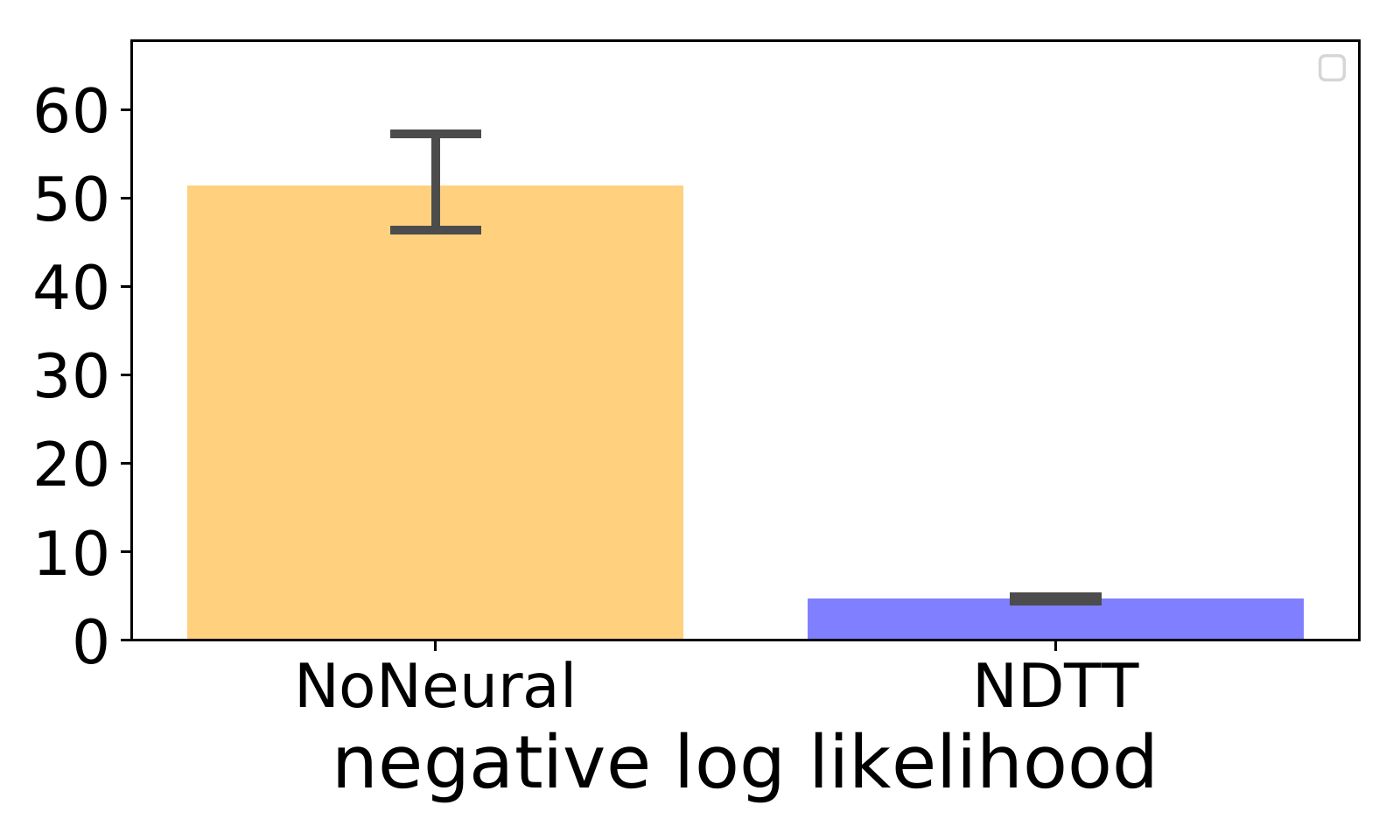}
				~
				\includegraphics[width=0.32\linewidth]{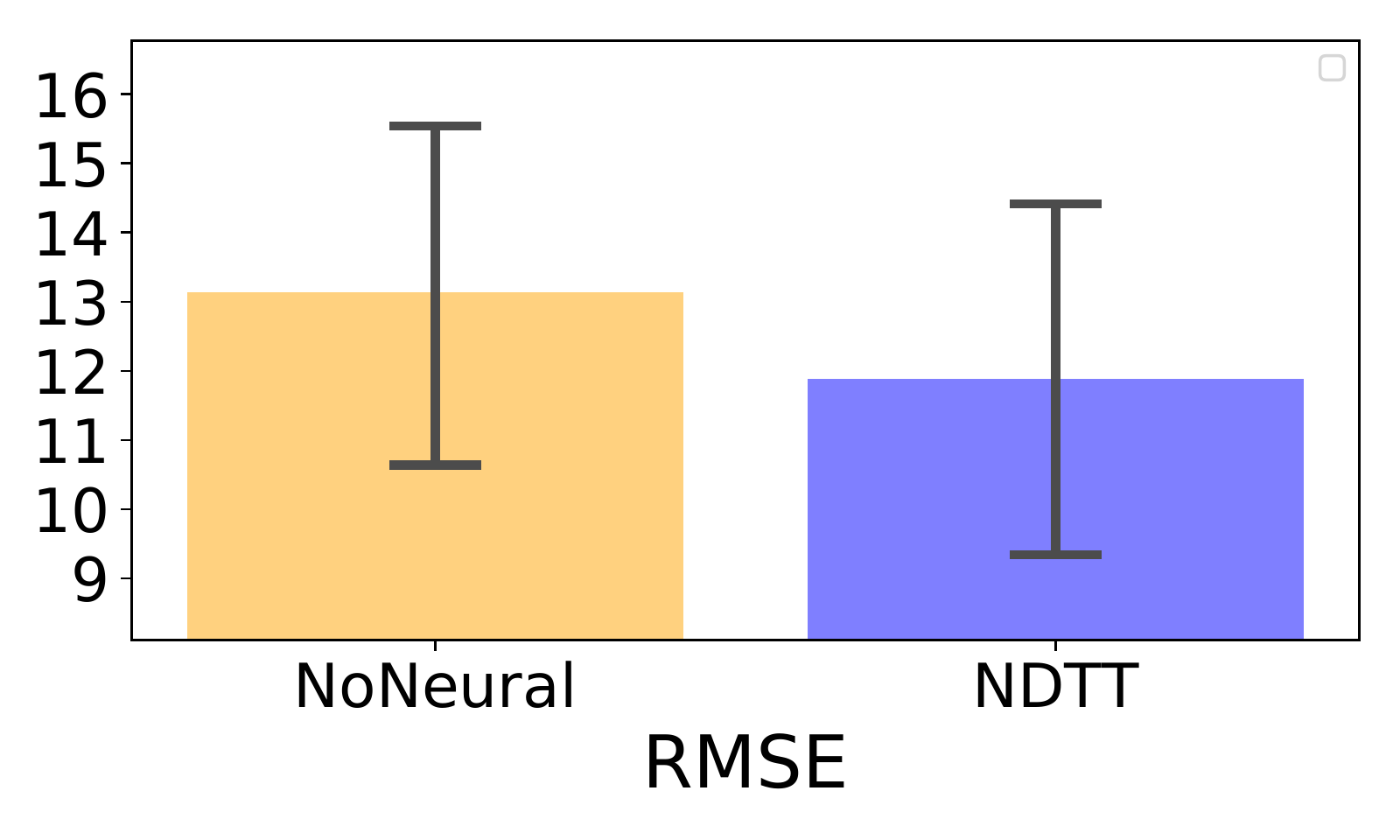}
				~
				\includegraphics[width=0.32\linewidth]{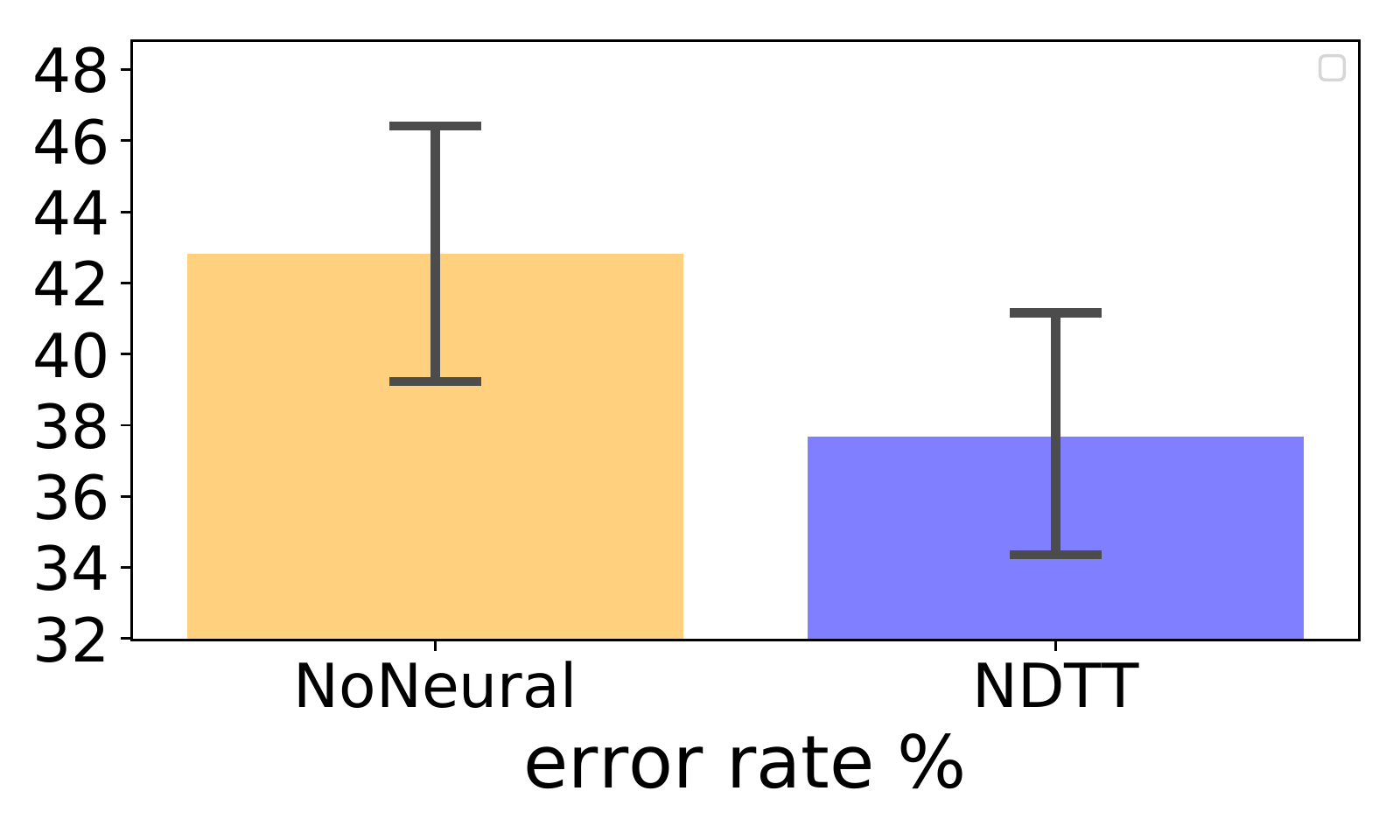}
				\vspace{-6pt}
				\caption{RoboCup Dataset}\label{fig:pred_robocup_noneural}
			\end{center}
		\end{subfigure}

		\caption{Ablation study of taking away neural networks from our Datalog programs in the real-world domains. The format of the graphs is the same as in \cref{fig:pred_results}.  The results imply that neural networks have been learning useful representations that are not explicitly specified in the Datalog programs.}
		\vspace{-12pt}
		\label{fig:pred_ablation_noneural}
	\end{center}
\end{figure*}

\subsection{Baseline Programs on RoboCup Dataset}\label{app:robocup_details_baseline}

As before, we also implemented baseline models that are inspired by the Know-Evolve and DyRep  frameworks (Trivedi, p.c.).  The non-embedded database facts about players and teams are specified just as in \cref{app:robocup_details} (\crefrange{line:robocup-facts-first}{line:robocup-facts-last}).

Like the Know-Evolve program for IPTV, the Know-Evolve program for RoboCup has no embeddings for its events:
\begin{lstlisting}[name=robocup_app,firstnumber=auto]
@\code{\dep }\code{ {\isevent}(\evto{kickoff}, 0).} @
@\code{\dep }\code{ {\isevent}(\evto{kick}, 0).} @
@\code{\dep }\code{ {\isevent}(\evto{goal}, 0).} @
@\code{\dep }\code{ {\isevent}(\evto{pass}, 0).} @
@\code{\dep }\code{ {\isevent}(\evto{steal}, 0).} @
\end{lstlisting}

As in IPTV, the embeddings are handled by separate facts.  Know-Evolve's embedding of an event does not depend on the event's type, but only on its set of participants. Thus, the \evto{kickoff}, \evto{kick}, and \evto{goal} events are simply represented by the embedding of the single player that participates in those events, which is defined exactly as in our full model of \cref{app:robocup_details}:
\begin{lstlisting}[name=robocup_app,firstnumber=auto]
@\code{\dep }\code{ {\embed}(\blk{player}, 8).} @
@\code{\blk{player}(\dvar{P}) \see \exoo{init}, in\_team(P,T).}@
\end{lstlisting}
For the \evto{pass} and \evto{steal} events, we also need an embedding for each \emph{unordered pair} of players (analogous to \blk{watch\_emb} in \cref{app:iptv_details_baseline} \cref{line:watch-emb}):
\begin{lstlisting}[name=robocup_app,firstnumber=auto]
@\code{\dep \embed(\blk{players}, 8).}@
@\code{\blk{players}(\dvar{P},\dvar{Q}) \dep \\
    \blk{player}(\dvar{P}) \usec \prm{pair}, \blk{player}(\dvar{Q}) \usec \prm{pair}.}@
\end{lstlisting}
All of these embeddings evolve over time.  Since teams do not participate directly in events, they do not have embeddings, in contrast to our full model in \cref{app:robocup_details}.

Each event's probability depends nonlinearly on the concatenated embeddings of its participants, e.g.,
\begin{lstlisting}[name=robocup_app,firstnumber=auto]
@\code{\evto{kick}(\dvar{P})}\ \ \code{ \dep }\code{ \blk{player}(\dvar{P}).} \label{line:kick-ke}@
@\code{\evto{pass}(\dvar{P},\dvar{Q})}\code{ \dep }\\ \code{ \blk{player}(\dvar{P}), \blk{player}(\dvar{Q}), \boo{teammate}(\dvar{P},\dvar{Q}).} \label{line:pass-ke}@
\end{lstlisting}
Note that because Know-Evolve does not allow changes over time in the set of possible events, it assigns a positive probability to the above events even at times when \dvar{P} does not have the ball.

Actually, \citet{trivedi-17-know,trivedi-19-dyrep} allow any event to take place at any time between any pair of entities.  Our Know-Evolve and DyRep programs take the liberty of going beyond this to impose some \emph{static} domain-specific restrictions on which events are possible.  For example, in RoboCup, \cref{line:pass-ke} only allows \evto{pass}ing between teammates, and \cref{line:kick-ke} only allows \evto{kick}ing from a player to itself (i.e., the ``pair'' of participants for \code{\evto{kick}(\dvar{P})} has only one unique participant).

An event updates the embeddings of its participants, e.g.,
\begin{lstlisting}[name=robocup_app,firstnumber=auto]
@\code{\blk{player}(\dvar{P}) \see \usec \prm{kick} \\ \evto{kick}(\dvar{P}), \blk{player}(\dvar{P}) \usec \prm{only}.} \label{line:robocup-ke-see-first}@
@\code{\blk{player}(\dvar{P}) \see \usec \prm{pass} \\ \evto{pass}(\dvar{P},\dvar{Q}), \blk{players}(\dvar{P},\dvar{Q}) \usec \prm{agent}.}@
@\code{\blk{player}(\dvar{Q}) \see \usec \prm{pass} \\ \evto{pass}(\dvar{P},\dvar{Q}), \blk{players}(\dvar{P},\dvar{Q}) \usec \prm{patient}.} \label{line:robocup-ke-see-last}@
\end{lstlisting}
where the bias vector is determined by the event type (e.g., \prm{kick} or \prm{pass}),
while the weight matrix is determined by the role played in the event of the participant being updated
(\prm{agent}, \prm{patient}, or \prm{only}---see \cref{app:robocup_details}).  Both types of parameters are shared across multiple rules.

For the DyRep program, the same events are possible as for Know-Evolve, and most of the rules are the same.  However, recall from \cref{app:iptv_details_baseline} that DyRep permits us to define a graph of entities.  Robot players are entities, of course.  We also consider the ball to be an entity, which is connected to player \dvar{P} by an edge when \dvar{P} possesses the ball.  This allows DyRep to update the embeddings of the participants in a \evto{pass} or \evto{steal} event to record the fact that the one who had the ball now lacks it, and vice-versa.  The model can therefore learn that \code{\evto{pass}(\dvar{P},\dvar{Q})} and \code{\evto{steal}(\dvar{Q},\dvar{P})} are much more probable when \dvar{P} has the ball.

DyRep requires the following new rules to handle the ball:
\begin{lstlisting}[name=robocup_app,firstnumber=auto]
@\code{\dep }\code{ {\embed}(\blk{ball}, 8).} @ 
@\code{\blk{ball}}\code{ \see }\code{ \exoo{init}.}@
\end{lstlisting}
as well as all of the rules from \cref{app:robocup_details} that update \boo{has\_ball}, which manage the edges of the evolving graph.  Note that $\sem{\blk{ball}}$ may drift over time but is never updated, since \blk{ball} is never one of the participants in an event.

Now we mechanically obtain the DyRep model by replacing Know-Evolve rules such as
\crefrange{line:robocup-ke-see-first}{line:robocup-ke-see-last} with DyRep-style
versions:
\begin{lstlisting}[name=robocup_app,firstnumber=auto]
@\code{\blk{player}(\dvar{P}) \see \usec \prm{kick} \\ \evto{kick}(\dvar{P}), \ \  \blk{player}(\dvar{P}) \use \prm{event}.}@
@\code{\blk{player}(\dvar{P}) \see \usec \prm{pass} \\ \evto{pass}(\dvar{P},\dvar{Q}), \blk{player}(\dvar{P}) \use \prm{event}.}@
@\code{\blk{player}(\dvar{Q}) \see \usec \prm{pass} \\ \evto{pass}(\dvar{P},\dvar{Q}), \blk{player}(\dvar{Q}) \use \prm{event}.}@
\end{lstlisting}
and then mechanically adding influences from the neighbors of \dvar{P} and \dvar{Q} (where the ball is the only possible neighbor):
\begin{lstlisting}[name=robocup_app,firstnumber=auto]
@\code{\blk{player}(\dvar{P}) \see  \\ \evto{kick}(\dvar{P}), \blk{ball} \usec \prm{ball}, \boo{has\_ball}(\dvar{P}).} \label{line:player-kick-ball-dyrep}@  
@\code{\blk{player}(\dvar{P}) \see  \\ \evto{pass}(\dvar{P},\dvar{Q}), \blk{ball} \usec \prm{ball}, \boo{has\_ball}(\dvar{P}).} \label{line:player-pass-ball-agent-dyrep}@
@\code{\blk{player}(\dvar{Q}) \see  \\ \evto{pass}(\dvar{P},\dvar{Q}), \blk{ball} \usec \prm{ball}, \boo{has\_ball}(\dvar{Q}).} \label{line:player-pass-ball-patient-dyrep}@
\end{lstlisting}

\paragraph{Remarks.} Recall that the DyRep model can unfortunately generate domain-impossible event sequences in which \dvar{P} kicks or passes the ball without actually having it.  However, such events never happen in \emph{observed} data.  As a result, the above rules can be simplified if we are only updating embeddings based on observed events (which is true in our experiments).  We can then remove the explicit \code{\boo{has\_ball}(\dvar{P})} condition from \cref{line:player-kick-ball-dyrep,line:player-pass-ball-agent-dyrep} because it is surely true when these rules are triggered by observed events.  And we can remove \cref{line:player-pass-ball-patient-dyrep} altogether, because its condition \code{\boo{has\_ball}(\dvar{Q})} is surely false when this rule is triggered by an observed event.  But then \boo{has\_ball} plays no role in the DyRep model anymore!  This shows that in effect, the model tracks the ball's possessor only by updating $\code{\blk{player}(\dvar{P})}$ whenever it observes an event with participant $\dvar{P}$ in which $\dvar{P}$ has the ball.  This type of tracking is imprecise (in particular, it does not immediately detect when $\dvar{P}$ \emph{acquires} the ball), which is why the DyRep model cannot learn from data to assign probability $\approx 0$ to domain-impossible events.

\subsection{Training Details}\label{app:training_details}

For every model in \cref{sec:exp}, including the baseline models, we had to choose the dimension $D$ that is specified in the $\embed$ and $\isevent$ declarations of its NDTT program.
  For simplicity, all declarations within a given program used the same
  dimension $D$, so that each program had a single hyperparameter to tune.  We
  tuned this hyperparameter separately for each combination of program, domain, and training size (e.g., each point in
  \cref{fig:learncurve_nhp} and each bar in
  \cref{fig:pred_results,fig:pred_robocup_ablation,fig:pred_ablation_noneural}), always choosing
  the $D$ that achieved the best performance on the dev set.  Our
  search space was \{4, 8, 16, 32, 64, 128\}.  In practice, the
  optimal $D$ for a model of a non-synthetic dataset (\cref{sec:iptv}) was usually 32 or 64.

To train the parameters for a given $D$, we used the Adam algorithm \citep{kingma-15} with its default settings and set the minibatch size to 1. 
We performed early stopping based on log-likelihood on the held-out dev set.

\subsection{Ablation Study II Details}\label{app:neural_away_details}

In the final experiment of \cref{sec:robocup}, all embeddings have dimension 0.  Each event type still has an extra dimension for its intensity (see \cref{sec:prob}).  The set of possible events at any time is unchanged.  However, the intensity of each possible event now depends only on \emph{which} rules proved or updated that possible event (through the bias terms of those rules); it no longer depends on the embeddings of the specific atoms on the right-hand-sides of those rules.  Two events may nonetheless have different intensities if they were proved by different \dep rules, or proved or updated by different sequences of \see rules (where the difference may be in the identity of the \see rules or in their timing).

Our experimental results in \cref{fig:pred_ablation_noneural} show that the neural networks have really been learning representations that are actually helpful for probabilistic modeling and prediction. 

\end{document}